\documentclass{article} 
\usepackage{iclr2024_conference,times}

\usepackage{amsmath,amsfonts,bm}

\def\eqref#1{equation~\ref{#1}}

\def\1{\bm{1}}

\DeclareMathAlphabet{\mathsfit}{\encodingdefault}{\sfdefault}{m}{sl}
\SetMathAlphabet{\mathsfit}{bold}{\encodingdefault}{\sfdefault}{bx}{n}

\usepackage[utf8]{inputenc} 
\usepackage[T1]{fontenc}    
\usepackage{hyperref}       
\usepackage{url}            
\usepackage{booktabs}       
\usepackage{amsfonts}       
\usepackage{nicefrac}       
\usepackage{microtype}      
\usepackage{caption}

\usepackage{multicol}

\usepackage{amssymb}
\usepackage{amsmath}
\usepackage{mathtools}
\usepackage{mathabx}
\usepackage{ mathrsfs }
\usepackage{amsthm}
\usepackage{tikz}
\usepackage{verbatim}
\usepackage{bbold}
\usepackage{wallpaper}
\usepackage{tikz-cd}
\usepackage{epigraph}
\usepackage{dsfont}

\usepackage{floatrow}

\usepackage{wrapfig,booktabs}

\usepackage{pict2e}
\usepackage{cleveref}
\usepackage{autonum}	
\usepackage{tikz-cd}
\usepackage{url}
\usepackage{tensor}
\usepackage{url} 
\usepackage{enumitem}
\usepackage{multirow}
\newtheoremstyle{definition_style}
{15pt}
{10pt}
{\itshape}
{}
{\bfseries}
{ }
{\newline}
{}
\theoremstyle{definition}
\newtheorem{Th}{Theorem}[section]

\newtheorem{Def}[Th]{Definition}
\newtheorem{Thm}[Th]{Theorem}

\newtheorem{Lem}[Th]{Lemma}

\newtheorem{Cor}[Th]{Corollary}

\title{ResolvNet: A Graph Convolutional Network with multi-scale Consistency}

\author{Christian Koke, Abhishek Saroha, Yuesong Shen, Marvin Eisenberger \&  Daniel Cremers 
	 \\
Technical University of Munich \\
}

\iclrfinalcopy 
\begin{document}

\maketitle
\vspace{-5mm}
\begin{abstract}
	It is by now a well known fact in the graph learning community that the presence of bottlenecks severely limits the ability of graph neural networks to propagate information over long distances. What so far has not been appreciated is that,  counter-intuitively, also the presence of strongly connected sub-graphs may severely restrict information flow in common architectures. Motivated by this observation, we introduce the concept of multi-scale consistency. At the node level this concept refers to the retention of a connected propagation graph even if connectivity varies over a given graph. At the graph-level, multi-scale consistency refers to the fact that distinct graphs describing the same object at different resolutions should be assigned similar feature vectors. As we show, both properties are not satisfied by popular graph neural network architectures. To remedy these shortcomings, we introduce ResolvNet, a flexible  graph neural network based on the mathematical concept of resolvents. We rigorously establish its  multi-scale consistency theoretically and verify it in extensive experiments on real world data: Here networks based on this ResolvNet architecture prove expressive; out-performing baselines significantly on many tasks; in- and outside the multi-scale setting.
\end{abstract}

\section{Introduction}\label{Int}

Learning on graphs has developed into a rich and complex field, providing spectacular results
on problems as varied as protein design \citep{ bronstein_nature}, traffic forecasting \citep{diffusion_traffic_forecasting}, particle physics \citep{Shlomi_2021}, recommender systems \citep{
	gao2023survey} and traditional tasks such as node- and graph classification \citep{Wu_2021, DBLP:journals/mva/XiaoWDG22}.

Despite their successes, 
graph neural networks (GNNs) are still plagued by  fundamental issues: 
Perhaps 
best known is the  phenomenon of oversmoothing, capturing the fact that node-features generated
by common GNN architectures become less 
informative as network depth increases \citep{oversmoothing18,oversmoothing20}. 
%
From the perspective of information flow however deeper networks would be preferable, as a $K$ layer message passing network \citep{mpnncm}, may only facilitate information exchange between  nodes that are at most $K$-edges apart -- a phenomenon commonly referred to as under-reaching \citep{AlonaTal, BronsteinInBottle}. 
%

However, even if information \textit{is} reachable within $K$ edges, the structure of the graph might not be conducive to communicating it between distant nodes:
If bottlenecks are present in the graph at hand, information from an exponentially growing receptive field needs to be squashed into fixed-size vectors 
to pass through the bottleneck. This oversquashing-phenomenon \citep{AlonaTal, BronsteinInBottle} prevents common architectures from  propagating messages between distant nodes without information loss in the presence of bottlenecks.

What has so far  not been appreciated within the graph learning community is that 
-- somewhat counter-intuitively -- also the presence of strongly connected subgraphs 
severely restricts
the information flow within popular graph neural network architectures; as we establish in this work.
Motivated by this observation, we consider the setting of multi-scale graphs and introduce, define and study the corresponding problem of multi-scale consistency for graph neural networks:

Multi-scale graphs are graphs whose edges 
are 
distributed on (at least) two scales: One large scale indicating strong connections within certain (connected) clusters, and one regular scale indicating a 
weaker, regular
connectivity outside these subgraphs. 
The lack of multi-scale consistency of common architectures then arises 
as two sides of the same coin: At the node level, prominent GNNs are unable to consistently integrate multiple connectivity scales into their propagation schemes: They essentially only propagate information along edges corresponding to the largest 
scale. At the graph level, current methods are not stable to variations in resolution scale: Two graphs describing the same underlying object at different resolutions are assigned vastly different feature vectors.  
\paragraph{Contributions:}
We introduce the concept of  multi-scale consistency for GNNs and study its two defining characteristics at the node- and graph levels. We establish that common GNN architectures suffer from a lack of multi-scale consistency and -- to remedy this shortcoming -- propose the \textbf{ResolvNet} architecture. This method is able to consistently integrate multiple connectivity scales occurring within graphs. At the node level, this manifests as ResolvNet -- in contrast to common architectures -- not being limited to propagating information via a severely disconnected effective propagation scheme, when multiple scales are present within a given graph. At the graph-level, this leads to ResolvNet provably and numerically verifiably assigning similar feature vectors to graphs describing the same underlying object at varying resolution scales; a property which -- to the best of our knowledge -- no other graph neural network has demonstrated.

\section{Multi-Scale Graphs and Multi-Scale Consistency}\label{scales}


\subsection{Multi-Scale Graphs}\label{twoscalegraphs}
We are interested in graphs
with
edges 
distributed on (at least) two scales: A large scale indicating strong connections within certain clusters, and a regular scale indicating 
a 
weaker, regular
connectivity outside these subgraphs. 
Before 
giving a precise definition,
%
we 
consider two instructive examples:

\paragraph{Example I. Large Weights:} A two-scale geometry as outlined above, 
might 
e.g. arise within weighted graphs discretizing underlying continuous spaces: Here, 
edge weights are typically determined by the inverse discretization length ($w_{ij}\sim 1/d_{ij}$), which might 
vary over the graph \citep{PostBook, graph_approx}. Strongly connected sub-graphs would then correspond to clusters of  nodes that are spatially closely co-located. 
Alternatively, such different scales can occur in social networks; e.g. if edge-weights are set to number of exchanged messages. Nodes representing (groups of) close friends would then typically be connected by stronger edges than nodes encoding mere acquaintances, which 
would 
typically have exchanged fewer messages.

\begin{figure}[H]
	(a)\includegraphics[scale=0.27]{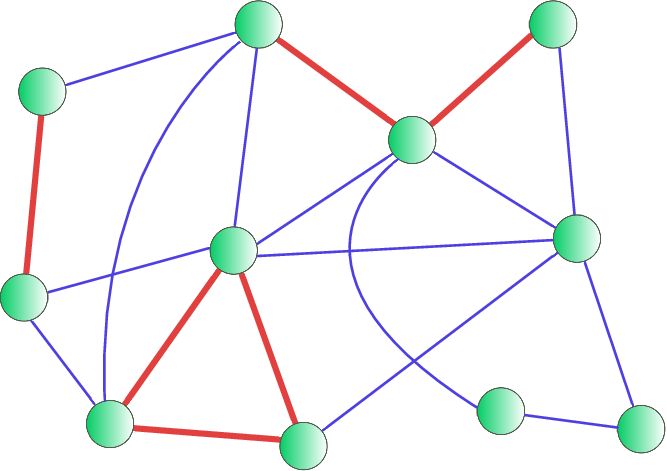}\hfill
	(b)\includegraphics[scale=0.27]{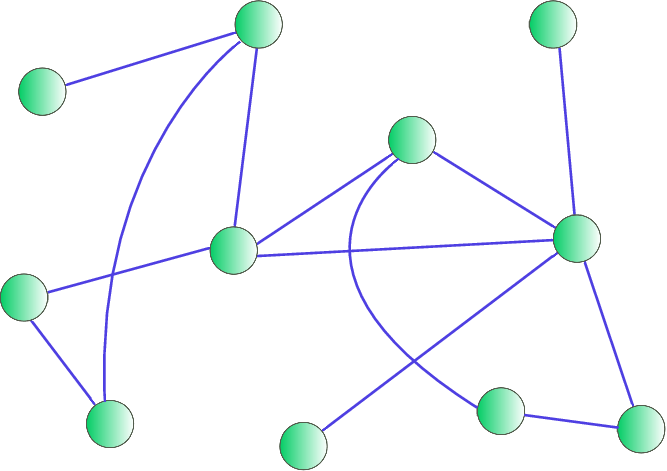}\hfill
	(c)\includegraphics[scale=0.27]{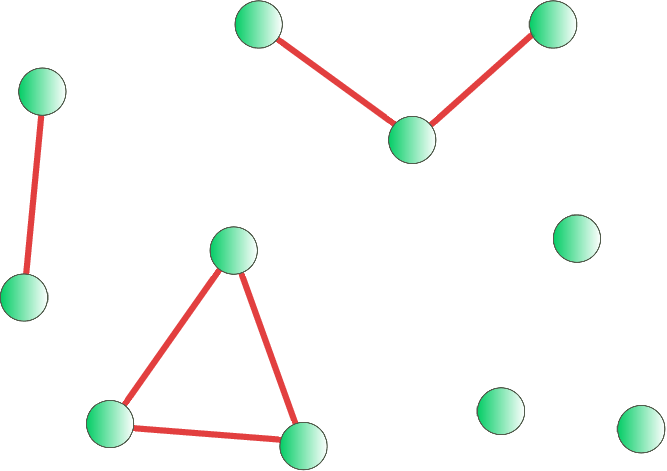}\hfill
	(d)\includegraphics[scale=0.27]{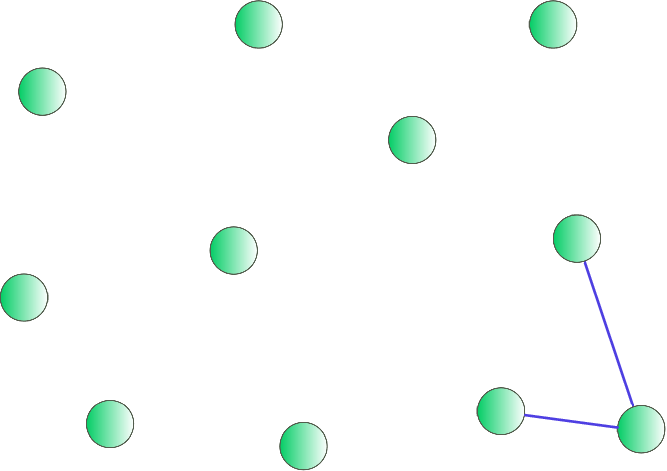}
	\captionof{figure}{(a) Graph $G$ with \textcolor{blue}{$\mathcal{E}_{\text{reg.}}$ (blue)} \& \textcolor{red}{$\mathcal{E}_{\text{high}}$ (red)};\ (b)  $G_{\text{reg.}}$; \ (c) $G_{\text{high}}$;\ (d) $G_{\text{excl.-reg. }}$    } 
	\label{graph_decomp}
\end{figure}

Given such a weighted graph, we partitions its weighted adjacency matrix $W = W_\text{reg.} + W_\text{high} $ into a disjoint sum over a part $W_\text{reg.} $ containing only regular edge-weights and part $ W_\text{high}$ containing only large edge-weights. This decomposition induces two graph structures  on the common node set $\mathcal{G}$: We set $G_{\text{reg.}} := (\mathcal{G},W_{\text{reg.}})$ and $G_{\text{high}}:=(\mathcal{G},W_{\text{high}})$ (c.f. also Fig. \ref{graph_decomp}).
%
%
%
%
\vspace{-0mm}
\vspace{-4mm}
In preparation for our discussion in Section \ref{previous_limits}, we also define the graph $G_{\text{excl.-reg.}}$ whose edges consists of those elements $(i,j) \in \mathcal{G} \times \mathcal{G}$ that do not have a neighbouring edge in $G_{\text{high}}$; i.e.  those edges $(i,j) \in \mathcal{E} \subsetneq \mathcal{G} \times \mathcal{G}$ so that for any $k \in \mathcal{G}$ we have $(W_\text{high})_{ik}, (W_\text{high})_{kj} = 0$ (c.f. Fig. \ref{graph_decomp} (d)).
%

\paragraph{Example 2. Many Connections:} Beyond weighted edges, disparate connectivities may also arise in unweightd graphs with binary adjacency matrices: In a social network where edge weights encode a binary friendship status for example, there might still exist closely knit communities within which every user is friends with every other,  while connections between such friend-groups may be sparser.

\begin{figure}[H]
	(a)\includegraphics[scale=0.27]{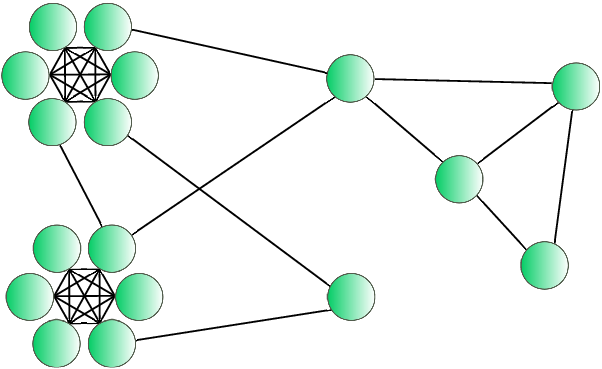}\hfill
	(b)\includegraphics[scale=0.27]{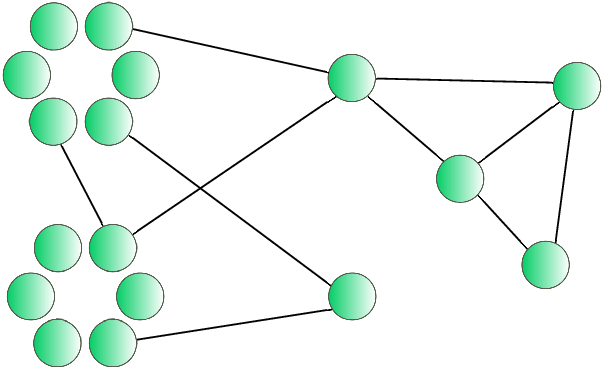}\hfill
	(c)\includegraphics[scale=0.27]{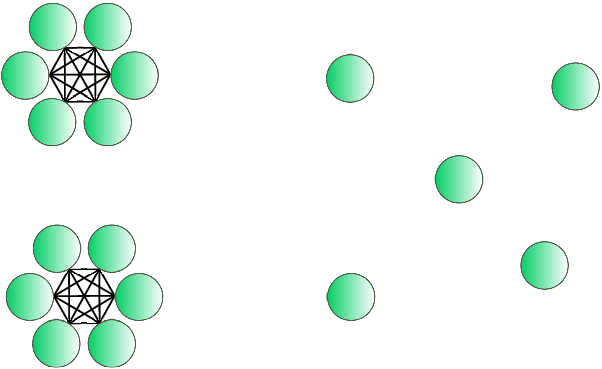}\hfill
	(d)\includegraphics[scale=0.27]{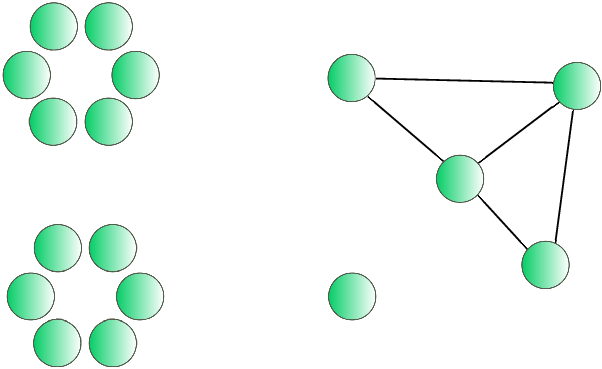}
	\captionof{figure}{(a) Graph $G$;
		\ (b)  $G_{\text{reg.}}$; \ (c) $G_{\text{high}}$;\ (d) $G_{\text{excl.-reg.}}$    } 
	\label{graph_decomp_unweighted}
\end{figure}

Here we may again split the adjacency matrix $W = W_\text{reg.} + W_\text{high} $ into a disjoint sum over a part $W_\text{reg.} $ encoding regular connectivity outside of tight friend groups and a summand $ W_\text{high}$ encoding closely knit communities into dense matrix blocks. Fig. \ref{graph_decomp_unweighted} depicts the corresponding graph structures. 


%

\vspace{-5mm}

\paragraph{Exact Definition:} To  unify both examples above into a common framework, we make use of tools from spectral graph theory; namely the spectral properties of the \textbf{Graph Laplacian}:

Given a graph $G$ on $N$ nodes, with 
weighted adjacency matrix $W$, diagonal degree matrix $D$ and node weights $\{\mu_i\}_{i =1}^N$ collected into the (diagonal) node-weight matrix $M = \text{diag}\left(\{\mu_i\}\right)$, the   (un-normalized) graph Laplacian $\Delta$ associated to the graph $G$ is defined as 
\begin{equation}
\Delta = M^{-1}(D - W).
\end{equation}
It is a well known fact in spectral graph theory, that much information about the connectivity of the graph $G$ is encoded into the first (i.e. smallest) non-zero eigenvalue $\lambda_1(\Delta)$ of this graph Laplacian $\Delta$ \citep{brouwer12,Chung:1997}. For an unweighted graph $G$ on $N$ nodes, this eigenvalue $\lambda_1(\Delta)$ is for example maximised if every node is connected to all other nodes (i.e. $G$ is an $N$-clique); in which case
we have 
$\lambda_1(\Delta) = N$. 
For weighted graphs, it is 
clear
that scaling all weights by a (large) constant $c$ exactly  also scales this eigenvalue  as $\lambda_1(\Delta) \mapsto c \cdot \lambda_1(\Delta)$. Thus the eigenvalue $\lambda_1(\Delta)$ is indeed a good proxy for measuring the strength of communities within a given graph $G$.

In order to formalize the concept of multi-scale graphs containing strongly connected subgraphs, we thus make the following definition:
\begin{Def}
	A Graph is called multi-scale if its weight-matrix $W$ admits a  \textit{disjoint} decomposition  
	\begin{equation}\label{scale_deco}
	W = W_{\text{reg.}} + W_{\text{high}} \ \ \ \ \text{with} \ \ \ \  \lambda_1(\Delta_{\text{high}}) > \lambda_{\max}(\Delta_{\text{reg.}}).
	\end{equation}
\end{Def}
Note that this decomposition  of $W$ also implies $\Delta = \Delta_{\text{reg.}} + \Delta_{\text{high}}$ for the respective Laplacians.
Note also that the graph-structure determined by   $G_{\text{high}}$ need not be completely connected for $\lambda_1(\Delta_{\text{high}})$ to be large (c.f. Fig.s \ref{graph_decomp} and \ref{graph_decomp_unweighted} (c)): If there are multiple disconnected communities,  $\lambda_1(\Delta_{\text{high}})$ is given as the minimal \textit{non-zero} eigenvalue of $\Delta_{\text{high}}$ restricted to these individual components of $G_{\text{high}}$. 
The largest eigenvalue $\lambda_{\max}(\Delta_{\text{reg.}})$ of $\Delta_{\text{reg.}}$ can be interpreted as measuring the "maximal connectivity" within the graph structure $G_{\text{reg.}}$: By  means of Gershgorin's circle theorem \citep{circle_thm}, we may bound it as $\lambda_{\max}(\Delta_{\text{reg.}}) \leq 2\cdot d_{\text{reg.},\max}$, with $d_{\text{reg.},\max}$ the maximal node-degree occurring in the graph $G_{\text{reg.}}$. Hence $\lambda_{\max}(\Delta_{\text{reg.}}) $ is small, if the connectivity within  $G_{\text{reg.}}$  is sparse.

\subsection{Multi-Scale consistency}\label{previous_limits}

We are now especially interested in the setting where the scales occurring in a given graph $G$ are well separated (i.e. $\lambda_1(\Delta_{\text{high}}) \gg \lambda_{\max}(\Delta_{\text{reg.}})$). Below, we describe how graph neural networks should ideally consistently incorporate such differing scales and detail how current architectures fail to do so. As the influence of multiple scales within graphs manifests differently depending on whether node-level- or graph-level tasks are considered, we will discuss these settings separately.
%
%
%
%

\subsubsection{Node Level Consistency: Retention of connected 
	propagation Graphs}\label{multiscale_consistency_node_level}

The fundamental purpose of graph neural networks is that of generating node embeddings not only dependent on local node-features, but also those of surrounding nodes. Even in the presence of multiple scales in a graph $G$, it is thus very much desirable that information is propagated  between all nodes connected via the edges of $G$ -- and not, say, only along the dominant scale (i.e. via $G_{\text{high}}$). 

This is however not the case for popular graph neural network architectures: Consider for example the graph convolutional network GCN \citep{Kipf}: Here, feature matrices $X$ are updated via the update rule $X \mapsto \hat{A}\cdot X$, with the off-diagonal elements of $\hat A$ given as $\hat{A}_{ij} = W_{ij}/{\sqrt{\hat{d}_i\cdot\hat{d}_j}}$.
Hence the relative importance $\hat{A}_{ij}$ of a message between a node $i$ of large (renormalised) degree $\hat{d}_i\gg 1$ and a node $j$ that is less strongly connected (e.g. $\hat{d}_j = \mathcal{O}(1)$)  is severely discounted.

In the presence of multiple scales as in Section \ref{twoscalegraphs}, this thus 
leads to messages essentially only being propagated over a disconnected effective propagation graph that is determined by the effective behaviour of $\hat{A}$ in the presence of multiple scales. As we show in Appendix \ref{limitprop} using
\begin{minipage}{0.53\textwidth}
	the decompositions $W = W_{\text{reg.}} + W_{\text{high}}$, the matrix $\hat{A}$ can in this setting effectively be approximated as:
	\begin{equation}\label{effective_renorm_adj}
	\hat{A} \approx \left( D^{-\frac12}_{\text{high}} W_{\text{high}} D^{-\frac12}_{\text{high}} + D^{-\frac12}_{\text{reg.}} \tilde{W}_{\text{excl.-reg.}} D^{-\frac12}_{\text{reg.}}\right)
	\end{equation}
	Thus information is essentially only propagated within the connected components of $G_{\text{high}}$ and via edges in $G_{\text{excl.-reg.}}$ (detached from edges in $G_{\text{high}}$).
	
\end{minipage}\hfill
\begin{minipage}{0.43\textwidth}
	\vspace{-3mm}
	\begin{figure}[H]
		(a)\includegraphics[scale=0.24]{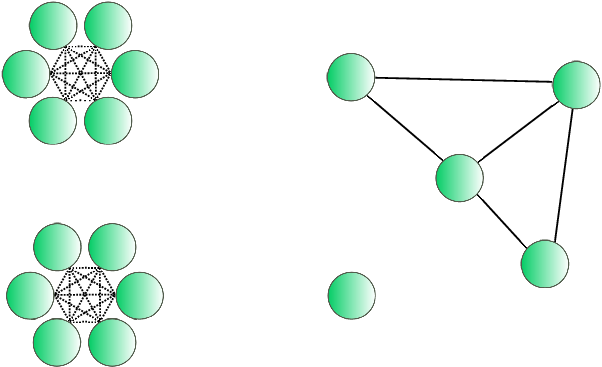}\hfill
		(b)\includegraphics[scale=0.21]{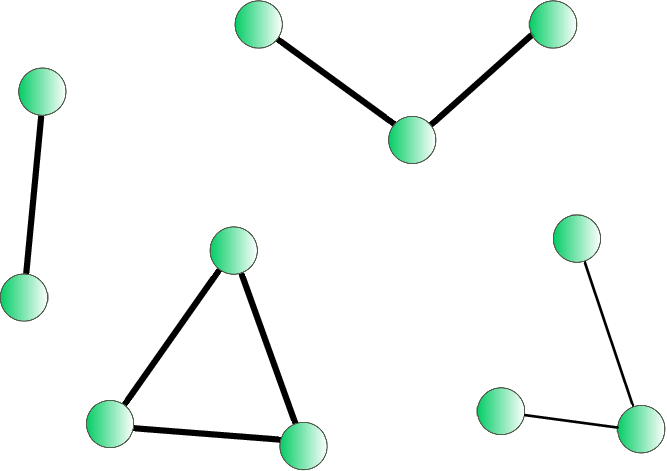}
		
		\captionof{figure}{Effective
			propagation graphs 
			for 
			original
			graphs in
			Fig. \ref{graph_decomp_unweighted} (a) and Fig. \ref{graph_decomp} (a)
		} 
		\label{previous_limit_graphs}
	\end{figure}
\end{minipage}

Appendix \ref{limitprop} further details that this reduction to propagating information only along a disconnected effective graph in the presence of multiple scales generically persists for popular methods (such as e.g. attention based methods  \citep{GAT} or spectral methods  \citep{Bresson}).

Propagating only over severely disconnected  effective graphs as in Fig. \ref{previous_limit_graphs} is clearly detrimental:
\begin{minipage}{0.6\textwidth}
	\begin{figure}[H]
		(a)\includegraphics[width=0.35\linewidth, trim= 0 -45 0 0]{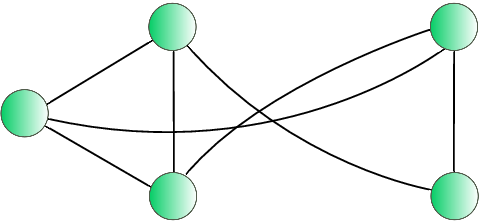}\hfill
		(b)\includegraphics[width=0.5\linewidth]{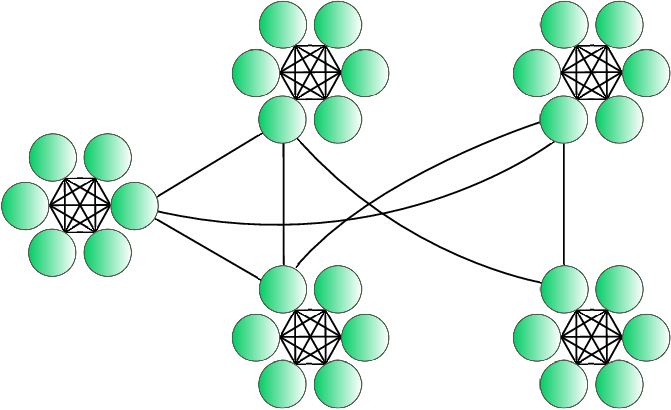}
		
		\captionof{figure}{Individual nodes (a) replaced by $6$-cliques (b)
		} 
		\label{tease_exp_gs}
	\end{figure}
\end{minipage}\hfill
\begin{minipage}{0.39\textwidth}
	\begin{figure}[H]
		\includegraphics[width=0.7\linewidth]{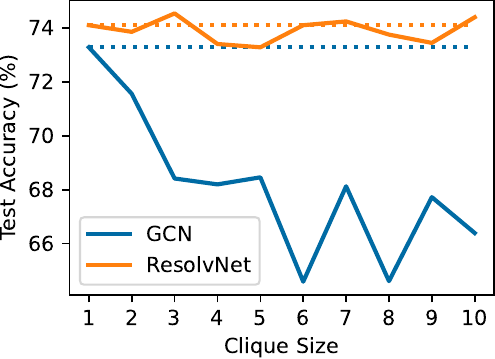}
		\captionof{figure}{Classification Accuracy} 
		\label{teaser}
	\end{figure}
\end{minipage}

As is evident from GCN's performance in  Fig.\ref{teaser}, duplicating individual nodes of a popular graph dataset into fully connected $k$-cliques   as in Fig. \ref{tease_exp_gs}  leads to a significant decrease in node-classification accuracy, as propagation between cliques becomes increasingly difficult with growing clique-size $k$. Details are provided in the  Experimental-Section \ref{experiments}. In principle however, duplicating nodes does not increase the complexity of the classification task at hand: Nodes and corresponding labels are only duplicated in the train-, val.- and test-sets.  What \emph{is} changing however, is the geometry underlying the problem; turning from a one-scale- into a two-scale setting with increasingly separated  scales.


In Section \ref{resolv_arch} below, we introduce ResolvNet, which is able to consistently integrate multiple scales within a given graph into its propagation scheme. As a result (c.f. Fig. \ref{teaser}) its classification accuracy is not affected by an increasing clique-size $k$ (i.e. an increasing imbalance in the underlying geometry).

\subsubsection{Graph Level  Consistency: Transferability between different Resolutions}\label{graph_level_desire} 
At the graph level, we desire that graph-level feature vectors $\Psi(G)$ generated by a network $\Psi$ for graphs $G$ are stable to changes in resolution scales: More precisely, if two graphs $G$ and $\underline{G}$ describe the same underlying object, space or phenomenon at different resolution scales, the generated feature vectors should be close, as they encode \emph{the same} object in the latent space. Ideally, we would have a Lipschitz continuity relation that allows to bound the difference in generated feature vectors $\|\Phi(G) - \Phi(\underline{G})\|$ in terms of a judiciously chosen distance $d(G,\underline{G})$ between the graphs as
\begin{equation}\label{desired_ineq}
\|\Psi(G) - \Psi(\underline{G})\|  \lesssim d(G,\underline{G}).
\end{equation}
Note that a relation such as (\ref{desired_ineq}) also allows to make statements about \emph{different} graphs $G, \widetilde{G}$ describing an underlying object at \emph{the same} resolution scale: If both such graphs are close to \emph{the same} coarse grained description $\underline{G}$, 
the triangle inequality yields $\|\Psi(G) - \Psi(\widetilde{G})\| \lesssim (d(G,\underline{G}) + d(\widetilde{G},\underline{G})) \ll 1$.

To make precise what we mean by the coarse grained description $\underline{G}$, we revisit the example of graphs discretizing an underlying continuous space, with edge weights  corresponding to inverse discretization length ($w_{ij}\sim 1/d_{ij}$). Coarse-graining -- or equivalently lowering the resolution scale -- then corresponds to merging multiple spatially co-located nodes in the original graph $G$ into single aggregate nodes in $\underline{G}$. As distance scales inversely with edge-weight, this means that we are precisely collapsing the strongly connected clusters within $G_{\text{high}}$ into single nodes. Mathematically, we then make this definition of the (lower resolution) coarse-grained graph $\underline{G}$ exact as follows:

%
%
%
%
%
%

\begin{minipage}{0.5\textwidth}
	\begin{Def}\label{limit_def}
		Denote by 	$\underline{\mathcal{G}}$ the set of connected components in $G_{\text{high}}$. We give this set a graph structure $\underline{G}$ as follows: Let $R$ and $P$ be elements of $\underline{\mathcal{G}}$ (i.e. connected components in $G_{\text{high}}$). We define the real number $\underline{W}_{RP}$ as
		$	\underline{W}_{RP} = \sum_{r\in R}\sum_{p\in P} W_{rp}$,
		with $r$ and $p$ nodes in the original graph $G$.
		We define the set of edges $\underline{\mathcal{E}}$ on $\underline{G}$ as  
		$	\underline{\mathcal{E}} = \{(R,P)\in\underline{\mathcal{G}}\times\underline{\mathcal{G}}: \underline{W}_{RP} >0 \}$
		and assign $\underline{W}_{RP}$ as weight to such edges.
		Node weights of nodes in $\underline{G}$ are defined similarly by aggregating weights of all nodes $r$ contained in the connected component $R$ of $G_{\text{high}}$ as
		$
		\underline{\mu}_R = \sum_{r \in R} \mu_r.
		$
	\end{Def}
	
\end{minipage}
\hfill
\begin{minipage}{0.48\textwidth}
	\vspace{-2mm}
	\begin{figure}[H]
		(a)\includegraphics[scale=0.26]{graph}\hfill
		\includegraphics[scale=0.26]{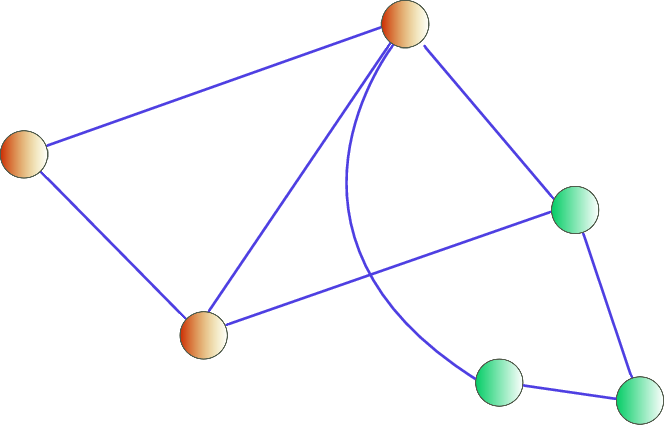}(b)
		(c)\includegraphics[scale=0.26]{graph_unweighted}\hfill
		\includegraphics[scale=0.26]{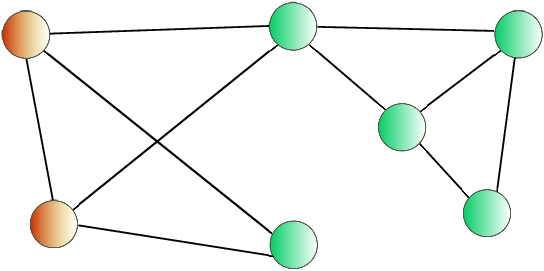}(d)
		
		\captionof{figure}{  Original $G$ (a,c)   and coarsified $\underline{G}$ (b,d)} 
		\label{really_true_limit}
	\end{figure}
\end{minipage}

This definition is of course also applicable 
to
Example 2 of Section \ref{twoscalegraphs}. Collapsing corresponding strongly connected component in a social network might then e.g. be interpreted as moving from interactions between individual users to considering interactions between (tightly-knit) communities.

While there have been theoretical investigations into this issue of \textbf{transferability} of graph neural networks between \emph{distinct graphs} describing the \emph{same system} \citep{levie, DBLP:conf/nips/RuizCR20, DBLP:journals/corr/abs-2109-10096, limitless}, the construction of an actual network with such properties -- especially outside the asymptotic realm of very large graphs -- has -- to the best of our knowledge -- so far not been successful. In Theorem \ref{graph_level_top_stab} and Section \ref{experiments} below, we show however that the ResolvNet architecture introduced in Section \ref{resolv_arch} below indeed provably and numerically verifiably 
satisfies 
(\ref{desired_ineq}), and is thus robust to variations in fine-print articulations of graphs describing the same  object.

\section{ResolvNet}\label{resolv_arch}
We now design a network -- termed ResolvNet -- that can consistently incorporate multiple scales within a given graph into its propagation scheme. At the node level, we clearly want to avoid disconnected effective propagation schemes as discussed in Section \ref{multiscale_consistency_node_level} in settings with well-separated connectivity scales. At the graph level -- following the discussion of Section \ref{graph_level_desire} -- we want to ensure that graphs $G$ containing strongly connected clusters and graphs $\underline{G}$ where these clusters are collapsed into single nodes are assigned similar feature vectors.

We can ensure both properties at the same time, if we manage to design a network whose propagation scheme when deployed on a multi-scale graph $G$ is effectively described by propagating over a coarse grained version $\underline{G}$ 
if the connectivity within the strongly connected clusters $G_{\text{high}}$ of $G$ is very large:

\begin{itemize}
	\item At the node level, this avoids effectively propagating over disconnected limit graphs as in Section \ref{multiscale_consistency_node_level}. Instead, information within strongly connected clusters is approximately homogenized and message passing is then performed on a (much better connected) coarse-grained version $\underline{G}$ of the original graph $G$ (c.f. Fig. \ref{really_true_limit}).
	\item At the graph level, this means that the stronger the connectivity within the strongly connected clusters is, the more the employed propagation on $G$ is like that on its coarse grained version $\underline{G}$.
	As we will see below, this can then be used to ensure the continuity property (\ref{desired_ineq}).
\end{itemize}

\subsection{The Resolvent Operator}\label{resolvent-operator}
As we have seen in Section \ref{multiscale_consistency_node_level} (and as is further discussed in Appendix \ref{limitprop}), standard message passing schemes are unable to generate networks having our desired multi-scale consistency properties.

A convenient multi-scale description of graphs is instead provided by the graph Laplacian $\Delta$ (c.f. Section \ref{twoscalegraphs}), as this operator encodes information about coarse geometry of a graph $G$ into small eigenvalues, while fine-print articulations of graphs correspond to large eigenvalues.
\citep{brouwer12,Chung:1997}. We are thus motivated to make use of this operator in our propagation scheme for ResolvNet.

In the setting of Example I of Section  \ref{twoscalegraphs}, letting the weights within $G_{\text{high}}$ go to infinity (i.e. increasing the connectivity within the strongly connected clusters) however implies $\|\Delta\|\rightarrow\infty$ for the norm of the Laplacian on $G$. Hence we \emph{can not} implement propagation simply as $X \mapsto \Delta\cdot X$: This would not reproduce the corresponding propagation scheme on $\underline{G}$ as we increase the connectivity within $G_{\text{high}}$: The Laplacian on $G$ does not converge to the Laplacian on $\underline{G}$ in the usual sense (it instead diverges $\|\Delta\|\rightarrow\infty$).

In order to capture convergence between operators  with such (potentially) diverging norms, mathematicians have developed other -- more refined -- concepts: 
Instead of distances between original operators, one considers distances between 
\textbf{resolvents} of such operators \citep{Teschl} :
\begin{Def}
	The resolvent of an operator $\Delta$ is defined as 
	$R_z(\Delta) :=  \left(\Delta - z\cdot Id  \right)^{-1}$,
	with $Id$ the identity mapping. Such resolvents are defined whenever $z$ is not an eigenvalue of $\Delta$.
\end{Def}
%

For Laplacians, taking $z<0$ hence ensures $R_z(\Delta)$ is defined.
Using this concept, we now rigorously establish convergence (in the resolvent sense) of the Laplacian  $\Delta$ on $G$ to the  (coarse grained) Laplacian $\underline{\Delta}$ on $\underline{G}$ as the connectivity within $G_{\text{high}}$ is increased.
To rigorously do so, we need to be able to translate signals between the original graph $G$ and its coarse-grained version $\underline{G}$:
\begin{Def}\label{proj_ops} Let $x$ be a scalar graph signal.
	Denote by $\mathds{1}_R$ the vector that has $1$ as entries on nodes $r$ belonging to the connected (in $G_{\text{high}}$) component $R$ and has entry zero for all nodes not in $R$. We define the down-projection operator $J^\downarrow$ component-wise via evaluating at node $R$ in $\underline{\mathcal{G}}$ as $(J^\downarrow x)_R = \langle \mathds{1}_R, x\rangle/\underline{\mu}_R$. This is then extended to feature \textit{matrices} $\{X\}$ via linearity. 
	The interpolation operator $J^\uparrow$ 
	is defined as $J^\uparrow u = \sum_R u_R \cdot\mathds{1}_R $; where $u_R$ is a scalar value (the component entry of $u$ at $R\in \underline{\mathcal{G}}$) and the sum is taken over all connected components of $G_{\text{high}}$.
\end{Def}
%
%
%
%
%
%
%
%
%
With these preparations, we can rigorously establish that the \textit{resolvent} of $\Delta$ 
approaches that of $\underline{\Delta}$:





\begin{Thm}\label{main_resolvent_theorem}
	We have $R_z(\Delta) \rightarrow  J^\uparrow R_z(\underline{\Delta}) J^\downarrow$ as 
	connectivity within $G_{\text{high}}$ 
	increases. Explicitly:
	\begin{equation}\label{resolvent_closeness}
	\left\|R_z(\Delta) -  J^\uparrow R_z(\underline{\Delta}) J^\downarrow\right\| =\mathcal{O}\left(  \frac{\lambda_{\max}(\Delta_{\text{reg.}})}{\lambda_1(\Delta_{\text{high}})}\right)
	\end{equation}
\end{Thm}
The fairly involved proof of Theorem \ref{main_resolvent_theorem}
is contained in Appendix \ref{main_thm_proof} and builds on previous work: We extend preliminary results in \cite{limitless} 
by establishing \textit{omni-directional} transferability (c.f. Theorem \ref{varying_spaces}
below)
and go beyond the toy-example of expanding a \textit{single} node into a fixed and connected
sub-graph with pre-defined edge-weights.
%
%

The basic idea behind ResolvNet is then to (essentially) implement message passing as $X \mapsto R_z(\Delta) \cdot X$. Building on Theorem \ref{main_resolvent_theorem}, Section \ref{stability} below then makes precise how this rigorously enforces multiscale-consistency as introduced in Section \ref{previous_limits} in the corresponding ResolvNet architecture.

\subsection{The ResolvNet Architecture}
Building on Section \ref{resolvent-operator}, we now design filters for which  feature propagation essentially proceeds along the coarsified graph of Definition \ref{limit_def} as opposed to the disconnected effective graphs of Section \ref{multiscale_consistency_node_level}, if multiple -- well separated -- edge-weight scales are present.

To this end, we note that 
Theorem \ref{main_resolvent_theorem}  states for $\lambda_1(\Delta_{\text{high}}) \gg \lambda_{\max}(\Delta_{\text{reg.}})$
(i.e.  well separated scales),
that applying  
$R_z(\Delta)$ 
to a signal $x$ 
is essentially 
the same as first projecting $x$ to 
$\underline{G}$ via $J^\downarrow$, then applying
$R_z(\underline{\Delta})$ there and finally lifting back to 
$G$ with $J^\uparrow$.
Theorem \ref{polynom_cor} In Appendix \ref{main_thm_proof} establishes that this behaviour also persists for powers of resolvents; i.e. we also have  $R^k_z(\Delta) \approx  J^\uparrow R^k_z(\underline{\Delta}) J^\downarrow$.

\paragraph{Resolvent filters:}

This motivates us to choose our learnable filters as polynomials in resolvents
\begin{equation}\label{res_poly_filter}
f_{z,\theta}(\Delta):= \sum\limits_{k=a}^K\theta_i \left[(\Delta - z Id)^{-1}\right]^k
\end{equation}
with learnable parameters $\{\theta_k\}_{k=a}^K$. Thus our method can be interpreted as a spectral method \citep{Bresson}, with learned functions $f_{z,\theta}(\lambda) = \sum_{k=a}^K \theta_k (\lambda - z)^{-k}$
applied to the operator $\Delta$ determining our convolutional filters.
The parameter $a$, which determines the starting index of the sum in (\ref{res_poly_filter}), may either be set to $a = 0$ (Type-$0$) or $a= 1$ (Type-I). As we show in
Theorem \ref{varying_spaces} below,
this choice will determine transferability properties of our models based on such filters.

Irrespectively,
both Type-$0$ and Type-I filters are able to learn a wide array of functions, as the following theorem (proved in Appendix \ref{approx_thm_proof}) shows:

\begin{Thm}\label{approx_theorem}	Fix $\epsilon > 0$ and $z<0$. For 
	arbitrary 
	functions $g,h : [0,\infty] \rightarrow \mathds{R}$ with $\lim_{\lambda\rightarrow\infty}g(\lambda)=\text{const.}$ and $\lim_{\lambda\rightarrow\infty}h(\lambda)=0$, there are filters $f_{z,\theta}^0, f_{z,\theta}^I$ of Type-$0$ and Type-I  respectively such that $\|f_{z,\theta}^0-g\|_{\infty}, \|f_{z,\theta}^I-h\|_{\infty} < \epsilon$.
\end{Thm}

\paragraph{The ResolvNet Layer:}   Collecting resolvent filters into a convolutional architecture, the layer wise update rule is then given as follows: Given a feature matrix $X^\ell \in \mathds{R}^{N \times F_\ell}$ in layer $\ell$, with column vectors $\{X^\ell_j\}_{j=1}^{F_\ell}$, the feature vector $X^{\ell+1}_i$  in layer $\ell+1$ is then calculated as 
$
X^{\ell+1}_i = \text{ReLu}\left( \sum_{j=1}^{F_{\ell}}  f_{z,\theta^{\ell+1}_{ij}}(\Delta) \cdot X^{\ell}_j + b_i^{\ell+1} \right)
$
with a learnable bias vector $b^{\ell+1}_i$. Collecting biases into a matrix $B^{\ell+1} \in \mathds{R}^{N\times F_{\ell+1}}$, we can efficiently implement this using matrix-multiplications as
\begin{equation}\label{matriximplementation}
X^{\ell+1} = \text{ReLu}\left( \sum\limits_{k = a}^{K}  (T - \omega Id)^{-k} \cdot X^{\ell}  \cdot W^{\ell+1}_{k}  + B^{\ell+1} \right)
\end{equation}
with weight matrices $\{W_k^{\ell+1}\}$ in $\mathds{R}^{F_{\ell}\times F_{\ell+1}}$. 
Biases are implemented as $b_i = \beta_i\cdot\mathds{1}_G$, with $\mathds{1}_G$ the vector of all ones on $G$ and $\beta_i \in \mathds{R}$ learnable. This is done to ensure that the effective propagation on $\underline{G}$ (if well seperated scales are present in $G$)
is not disturbed by non-transferable bias terms on the level of entire networks. This can be traced back to the fact that 
$J^\downarrow \mathds{1}_{G} = \mathds{1}_{\underline{G}}$ and $J^\uparrow \mathds{1}_{\underline{G}} = \mathds{1}_{G}$. 
A precise discussion of this matter is contained in Appendix \ref{stab_theo}.


\paragraph{Graph level feature aggregation:} As we will also consider the prediction of \textit{graph-level} properties in our experimental Section \ref{experiments} below, we need to sensibly aggregate node-level features into graph-level features on \textit{node-weighted} graphs: As opposed to standard 
aggregation schemes (c.f. e.g. \citet{GIN}), we define an aggregation scheme $\Psi$ that takes into account node weights and maps a feature matrix $X \in \mathds{R}^{N \times F}$ to a graph-level feature vector  $\Psi(X) \in \mathds{R}^{F}$  via $\Psi(X)_j = \sum_{i = 1}^N |X_{ij}|\cdot\mu_i$.

\section{Multi-Scale consistency and Stability Guarantees}\label{stability}

\paragraph{Node Level:}
We now establish rigorously that instead of propagating along disconnected effective graphs (c.f. Fig. \ref{previous_limit_graphs}),
ResolvNet instead propagates node features along the coarse-grained graphs of Fig. \ref{really_true_limit} if multiple separated scales are present:

%
\begin{Thm}\label{varying_spaces}
	Let $\Phi$ and $\underline{\Phi}$	be the maps associated to ResolvNets with the same learned weight matrices and biases but
	deployed on graphs $G$ and $\underline{G}$ as defined in Section \ref{resolv_arch}. We have	
	\begin{equation}\label{stab_eq_0}
	\|\Phi(J^\uparrow \underline{X}) -  J^\uparrow\underline{\Phi}(\underline{X})\|_2 \leq \left(C_1(\mathscr{W})\cdot\|\underline{X} \|_2+C_2(\mathscr{W},\mathscr{B})\right)  \cdot  	\left\|R_z(\Delta) -  J^\uparrow R_z(\underline{\Delta}) J^\downarrow\right\|
	\end{equation}
	if the network is based on Type-$0$ resolvent filters (c.f. Section \ref{resolv_arch}). Additionally, we have
	\begin{equation}\label{stab_eq_I}
	\|\Phi(X) - J^\uparrow\underline{\Phi}(J^\downarrow X)\|_2 \leq \left(C_1(\mathscr{W})\cdot\|X \|_2+C_2(\mathscr{W},\mathscr{B})\right)  \cdot  	\left\|R_z(\Delta) -  J^\uparrow R_z(\underline{\Delta}) J^\downarrow\right\|
	\end{equation}
	if only Type-I filters are used in the network. Here $C_1(\mathscr{W})$ and $C_2(\mathscr{W},\mathscr{B})$ are constants that depend polynomially on singular values of learned weight matrices $\mathscr{W}$ and biases $\mathscr{B}$.
\end{Thm}
The 
proof -- as well as additional results --  may be found in Appendix \ref{top_stab_theo}. Note that Theorem \ref{main_resolvent_theorem} implies that 
both equations tends to zero for increasing scale separation 
$\lambda_1(\Delta_{\text{high}}) \gg \lambda_{\max}(\Delta_{\text{reg.}})$.


The difference between utilizing Type-$0$ and Type-I resolvent filters, already alluded to in the preceding Section \ref{resolv_arch}, now can be understood as follows: Networks based on Type-$0$ filters effectively propagate signals \textit{lifted} from the coarse grained graph $\underline{G}$ to the original graph $G$ along 
$\underline{G}$ if $\lambda_1(\Delta_{\text{high}}) \gg \lambda_{\max}(\Delta_{\text{reg.}})$. In contrast -- in the same setting -- networks based on Type-I resolvent filters effectively first \textit{project} any input signal on $G$ to $\underline{G}$, propagate there and then lift
back to $G$.

\paragraph{Graph Level:}
Beyond a single 
graph, we  also establish graph-level multi-scale consistency:
As discussed in Section \ref{graph_level_desire}, if two graphs describe  the same underlying object (at different resolution scales) 
corresponding feature vectors should be similar.
This is captured by our next result:
\begin{Thm}\label{graph_level_top_stab}
	Denote by $\Psi$ the aggregation method introduced in Section \ref{resolv_arch}. With $\mu(G) = \sum_{i =1}^N \mu_i$ the total weight of the graph $G$, we have in the setting of  Theorem \ref{varying_spaces} with Type-I filters, that
	\begin{equation}
	\|\Psi\left(\Phi(X)\right) - \Psi\left(\underline{\Phi}(J^\downarrow X)\right)\|_2 \leq \sqrt{\mu(G)} \left(C_1(\mathscr{W})\|X \|_2+C_2(\mathscr{W},\mathscr{B})\right)    	\left\|R_z(\Delta) -  J^\uparrow R_z(\underline{\Delta}) J^\downarrow\right\|.
	\end{equation}
\end{Thm}
%

This result thus indeed establishes the desired continuity relation (\ref{desired_ineq}), with the distance metric $d(G,\underline{G})$  provided by the similarity $\left\|R_z(\Delta) -  J^\uparrow R_z(\underline{\Delta}) J^\downarrow\right\|$ of the resolvents of the two graphs. 

\section{Experiments}\label{experiments}
\paragraph{Node Classification:}
To establish that the proposed \textbf{ResolvNet} architecture \textbf{not only performs well} in multi-scale settings,  
we conduct node classification experiments on multiple  \textit{un-weighted}
real world datasets,
ranging in edge-homophily $h$ from $h=0.11$ (very heterophilic), to $h=0.81$ (very homophilic).
Baselines constitute an ample set of established and recent methods:  
Spectral approaches, are represented by ChebNet \citep{Bresson}, 
GCN \citep{Kipf}, BernNet \citep{bernnet},
ARMA \citep{ARMA}
and MagNet \citep{magnet}.
Spatial methods are given by
GAT \citep{GAT}, SAGE \citep{SAGE} and  GIN \citep{GIN}. 
We also consider 
PPNP \citep{PredictThenPropagate} 
and NSD \citep{Sheaf}.
Details on datasets, experimental setup and hyperparameters are provided in Appendix \ref{exp_det}.

\begin{table*}[h!]
	\caption{Average Accuracies [$\%$] with uncertainties encoding the 95 \% confidence Level. Top three models are coloured-coded as \textcolor{red}{\textbf{First}}, \textcolor{blue}{\textbf{Second}}, \textcolor{violet}{\textbf{Third}}.  }
	\begin{footnotesize}
		\begingroup
		\vspace{-3mm}
		\setlength{\tabcolsep}{0.8pt} 
		\renewcommand{\arraystretch}{1} 
		\begin{tabular}{lcccc|ccccc|}
			\toprule
			\ & \textbf{MS. Acad.} &\textbf{Cora} & \textbf{Pubmed} & \textbf{Citeseer}& \textbf{Cornell} &\textbf{Actor} &\textbf{Squirrel}  & \textbf{Texas} \\
			$h$ & 0.81 & 0.81 & 0.80& 0.74 & 0.30& 0.22 &0.22  & 0.11\\

			\midrule
			
			SAGE   & \textcolor{blue}{$\mathbf{91.75}$}\tiny{$\pm 0.09$}   &  $80.68$\tiny{$\pm 0.30$} &  $74.42$\tiny{$\pm 0.42$}  & $72.68$\tiny{$\pm0.32$} & $86.01$\tiny{$\pm 0.72$} & $28.88 $\tiny{$\pm0.32 $}  & $25.99$\tiny{$\pm 0.28$}    & $88.92$\tiny{$\pm 0.73$} \\
			
			GIN    & $72.93$\tiny{$\pm1.94$}  & $74.12$\tiny{$\pm1.21$} &   $74.59$\tiny{$\pm0.45$} & $68.11$\tiny{$\pm0.69$}  & $65.58$\tiny{$\pm1.23$}  & $ 23.69$\tiny{$\pm 0.28$}  & $24.91$\tiny{$\pm 0.58$}  
			&$72.64$\tiny{$\pm1.19$} \\
			GAT & $89.49$\tiny{$\pm0.15$}  & $80.12$\tiny{$\pm0.33$} & $77.12$\tiny{$\pm0.41$}  & \textcolor{violet}{$\mathbf{73.20}$}\tiny{$\pm0.37$} & $74.39$\tiny{$\pm0.93$}
			& $24.55$\tiny{$\pm0.28 $}  & \textcolor{blue}{$\mathbf{27.22}$}\tiny{$\pm 0.31$}  
			& $75.31$\tiny{$\pm1.09$} \\

			NSD 	 & $90.78$\tiny{$\pm 0.13$}  & $70.34$\tiny{$\pm 0.47$} &  $69.74$\tiny{$\pm 0.50$} & $64.39$\tiny{$\pm 0.50$}  & \textcolor{red}{$\mathbf{87.78}$}\tiny{$\pm 0.65$}  
			& $27.62$\tiny{$\pm 0.39$} & $24.96$\tiny{$\pm 0.27$} &\textcolor{red}{$\mathbf{91.64}$}\tiny{$\pm 0.62$} \\
			
			PPNP      & $91.22$\tiny{$\pm0.13$}  & \textcolor{blue}{$\mathbf{83.77}$}\tiny{$\pm0.27$} & \textcolor{blue}{$\mathbf{78.42}$}\tiny{$\pm0.31$}  & \textcolor{blue}{$\mathbf{73.25}$}\tiny{$\pm0.37$} & $71.93$\tiny{$\pm0.84$} 
			& $ 25.93$\tiny{$\pm 0.35$}  &$23.69$\tiny{$\pm0.43$}  
			& $70.73$\tiny{$\pm1.27$} \\

			ChebNet & \textcolor{violet}{$\mathbf{91.62}$}\tiny{$\pm 0.10$}& $78.70$\tiny{$\pm 0.37$}&$73.63$\tiny{$\pm 0.43$}& $72.10$\tiny{$\pm 0.43$}  &  $85.99$\tiny{$\pm 0.10$} &
			\textcolor{blue}{$\mathbf{29.51}$}\tiny{$\pm0.31$} & $25.68$\tiny{$\pm0.28$}  & \textcolor{blue}{$\mathbf{91.01}$}\tiny{$\pm 0.59$}\\

			GCN & $90.81$\tiny{$\pm 0.10$}  & $81.49$\tiny{$\pm 0.36$} & $76.60$\tiny{$\pm 0.44$}  & $71.34$\tiny{$\pm 0.45$} &$73.35$\tiny{$\pm 0.88$} 
			& $24.60 $\tiny{$\pm 0.28$} & \textcolor{red}{$\mathbf{30.40}$}\tiny{$\pm 0.40$}   
			& $76.16$\tiny{$\pm 1.12$} \\
			
			
			MagNet    & $87.23$\tiny{$\pm 0.16$}   &  $76.50$\tiny{$\pm 0.42$} &  $68.23$\tiny{$\pm 0.44$} & $70.92$\tiny{$\pm 0.49$} & \textcolor{blue}{$\mathbf{87.15}$}\tiny{$\pm 0.66$} 
			& \textcolor{red}{$\mathbf{30.50} $}\tiny{$\pm0.32 $}  &$23.54 $\tiny{$\pm0.32 $}  
			& \textcolor{violet}{$\mathbf{90.84}$}\tiny{$\pm 0.54$}\\	
			
			ARMA & $88.97$\tiny{$\pm 0.18$}  & $81.24$\tiny{$\pm 0.24$} &  $76.28$\tiny{$\pm 0.35$} & $70.64$\tiny{$\pm 0.45$} & $83.68$\tiny{$\pm 0.80$} 
			& $24.40 $\tiny{$\pm 0.45$}  & $22.72 $\tiny{$\pm 0.42$}  
			& $87.41$\tiny{$\pm 0.73$} \\

			BernNet   & $91.37$\tiny{$\pm 0.14$}  & \textcolor{violet}{$\mathbf{83.26}$}\tiny{$\pm 0.24$}  & \textcolor{violet}{$\mathbf{77.24}$}\tiny{$\pm 0.37$} & $73.11$\tiny{$\pm 0.34$} & \textcolor{violet}{$\mathbf{87.14}$}\tiny{$\pm 0.57$}  
			& $ 28.90$\tiny{$\pm 0.45$}  & $ 22.86$\tiny{$\pm0.32 $} 
			&$89.81$ \tiny{$\pm 0.68$} \\

			\midrule

			ResolvNet &  \textcolor{red}{$\mathbf{92.73}$}\tiny{$\pm0.08$}  &  \textcolor{red}{$\mathbf{84.16}$}\tiny{$\pm 0.26$}   &  \textcolor{red}{$\mathbf{79.29}$}\tiny{$\pm 0.36$} &   \textcolor{red}{$\mathbf{75.03}$}\tiny{$\pm 0.29$}  & $84.92$\tiny{$\pm 1.43$}
			& \textcolor{violet}{$\mathbf{29.06}$}\tiny{$\pm 0.32$}  & \textcolor{violet}{$ \mathbf{26.51}$}\tiny{$\pm 0.23 $}  
			& $87.73$\tiny{$\pm 0.89$}   \\

			\bottomrule
		\end{tabular}
		\endgroup
	\end{footnotesize}
	\vskip -0.1in
	\label{node_result_table}
\end{table*}
As is evident from Table \ref{node_result_table}, \textbf{ResolvNet out-performs all baselines in the homophilic setting}.
This can be traced back to the inductive bias ResolvNet is equipped with by design: It might be summarized as "Nodes that are strongly connected should be assigned similar feature vectors" (c.f. Theorem \ref{varying_spaces}) . This inductive bias -- necessary to achieve a consistent incorporation of multiple scales -- is
of course counterproductive in the presence of heterophily; here nodes that are connected by edges generically have \textit{differing} labels and should thus be assigned different feature vectors. 
However the ResolvNet architecture also performs well on most heterophilic graphs: It e.g. out-performs NSD -- a recent state of the art method specifically developed for heterophily -- on two such 
graphs.
%
%
%
%

%
%
%
\textbf{Node Classification for increasingly separated scales:} To test ResolvNet's ability to consistently incorporate multiple scales in the unweighted setting against a representative baseline, we duplicated individual nodes on the Citeseer dataset  \citep{Citeseer} $k$-times to form (fully connected) $k$-cliques; keeping the train-val-test partition constant.
GCN and ResolvNet were then trained on the same ($k$-fold expanded) train-set and asked to classify nodes on the ($k$-fold expanded) test-partition. As discussed  in Section \ref{desired_ineq} (c.f. Fig.\ref{teaser}) GCN's performance decreased significantly, while ResolvNet's accuracy stayed essentially constant; showcasing its ability to consistently incorporate multiple scales.

%
%
%


%
%
\paragraph{Regression on real-world multi-scale graphs:} In order to showcase the properties of our newly developed method on real world data admitting a two-scale behaviour, we evaluate on the task of molecular property prediction.
While   ResolvNet is not designed for 
this setting,
this task still allows to fairly compare its expressivity and stability properties against other non-specialized graph neural networks \citep{OGBN}.
Our  dataset (QM$7$; \citet{rupp}) contains descriptions of $7165$ organic molecules; each containing hydrogen and up to seven types of heavy atoms. A molecule is represented by its Coulomb matrix, whose off-diagonal elements 	$	C_{ij}	 = Z_iZ_j/|\vec{x}_i-\vec{x}_j|$ correspond to the Coulomb repulsion between atoms $i$ and $j$. We treat $C$ as a weighted adjacency matrix.  Prediction target is the molecular atomization energy, which -- crucially -- depends on long range interaction within molecules \citep{Zhang_2022}. However, with edge-weights $C_{ij}$ scaling as inverse distance, long range propagation of information is scale-suppressed in the graph determined by $C$, when compared to the much larger weights between closer atoms. We choose Type-I filters in ResolvNet, set node weights as atomic charge ($\mu_i = Z_i$) 	and use  one-hot encodings of atomic charges $Z_i$  as node-wise input features.

\begin{minipage}{0.69\textwidth}
	
	As is evident from Table \ref{qm7_result_table}, our method produces significantly lower mean-absolute-errors (MAEs) than  baselines of Table \ref{node_result_table}    deployable on weighted graphs.  We attribute this to the fact that our model allows for long range information propagation within each molecule, as propagation along corresponding edges is suppressed for baselines but not for our model (c.f. Section \ref{multiscale_consistency_node_level}).  Appendix contains additional experiments on   QM$9$ \citep{QM9}; finding similar performance for (long-range dependent) energy targets.
\end{minipage}\hfill
\begin{minipage}{0.29\textwidth}
	\vspace{-4mm}
	\begin{table}[H]
		\small
		\centering
		\caption{QM$7$-MAE }
		\begingroup
		
		\setlength{\tabcolsep}{0.5pt} 
		\renewcommand{\arraystretch}{1} 
		\vspace{-3mm}
		\begin{tabular}{lr}
			\toprule
			\textbf{QM$7$} & MAE $[kcal/mol]$   \\
			\midrule
			BernNet   & $113.57$\tiny{$\pm  62.90$}  \\
			GCN & $61.32 $\tiny{$\pm 1.62$}   \\
			ChebNet & $59.57$\tiny{$\pm 1.58$}\\	
			ARMA & $59.39$\tiny{$\pm 1.79$}   \\
			\midrule	
			ResolvNet &  $\mathbf{16.52}$\tiny{$\pm 0.67$}    \\
		\end{tabular}
		\endgroup
		\label{qm7_result_table}
	\end{table}
\end{minipage}

\paragraph{Stability to varying the resolution-scale:}\
To numerically verify the Stability-Theorem \ref{graph_level_top_stab} -- which guarantees similar graph-level feature vectors for graphs describing the same underlying object at different resolution scales -- we conduct additional experiments:
We modify (all) molecular graphs  of QM$7$ 
by deflecting hydrogen atoms (H) out of their equilibrium positions towards the respective nearest heavy atom. This introduces a two-scale setting precisely as discussed in section \ref{scales}: Edge weights between heavy atoms remain the same, while  Coulomb repulsions between H-atoms and respective nearest heavy atom increasingly diverge.
Given an original molecular graph $G$ with node weights $\mu_i=Z_i$, the corresponding coarse-grained graph $\underline{G}$ corresponds to a description where heavy atoms and surrounding H-atoms are aggregated into single super-nodes.
Node-features of aggregated nodes are now no longer one-hot encoded charges, but normalized bag-of-word vectors whose individual entries  encode how much of the total charge of a given super-node is contributed by individual atom-types.
Appendix \ref{exp_det} provides additional details and examples.

In this setting, we now compare features generated for coarsified graphs $\{\underline{G}\}$, with feature
\begin{minipage}{0.56\textwidth}
	generated for graphs $\{G\}$ where 
	hydrogen atoms have 
	been deflected 
	but have not yet
	completely arrived  at the positions of nearest  heavy atoms.
	Feature vectors are generated with the previously trained networks
	of
	Table \ref{qm7_result_table}. A corresponding plot is presented in Figure  \ref{collapse_graph}. Features generated by ResolvNet converge as the larger scale  increases (i.e. the distance between hydrogen and heavy atoms decreases). This result numerically verifies the scale-invariance Theorem \ref{graph_level_top_stab}. As reference, we also plot the norm differences corresponding to 
	baselines, which  do not decay.
	We might thus conclude that these models -- as opposed to ResolvNet -- are scale- and resolution sensitive when generating graph level features.
	For BernNet we 
	observe a divergence behaviour, which we attribute to numerical instabilities.
\end{minipage}\ \
\begin{minipage}{0.4\textwidth}
	\vspace{-3mm}
	\begin{figure}[H]
		\caption{Feature-vector-difference for collapsed ($\underline{F}$) and deformed ($F$) graphs. }
		\includegraphics[scale=0.40]{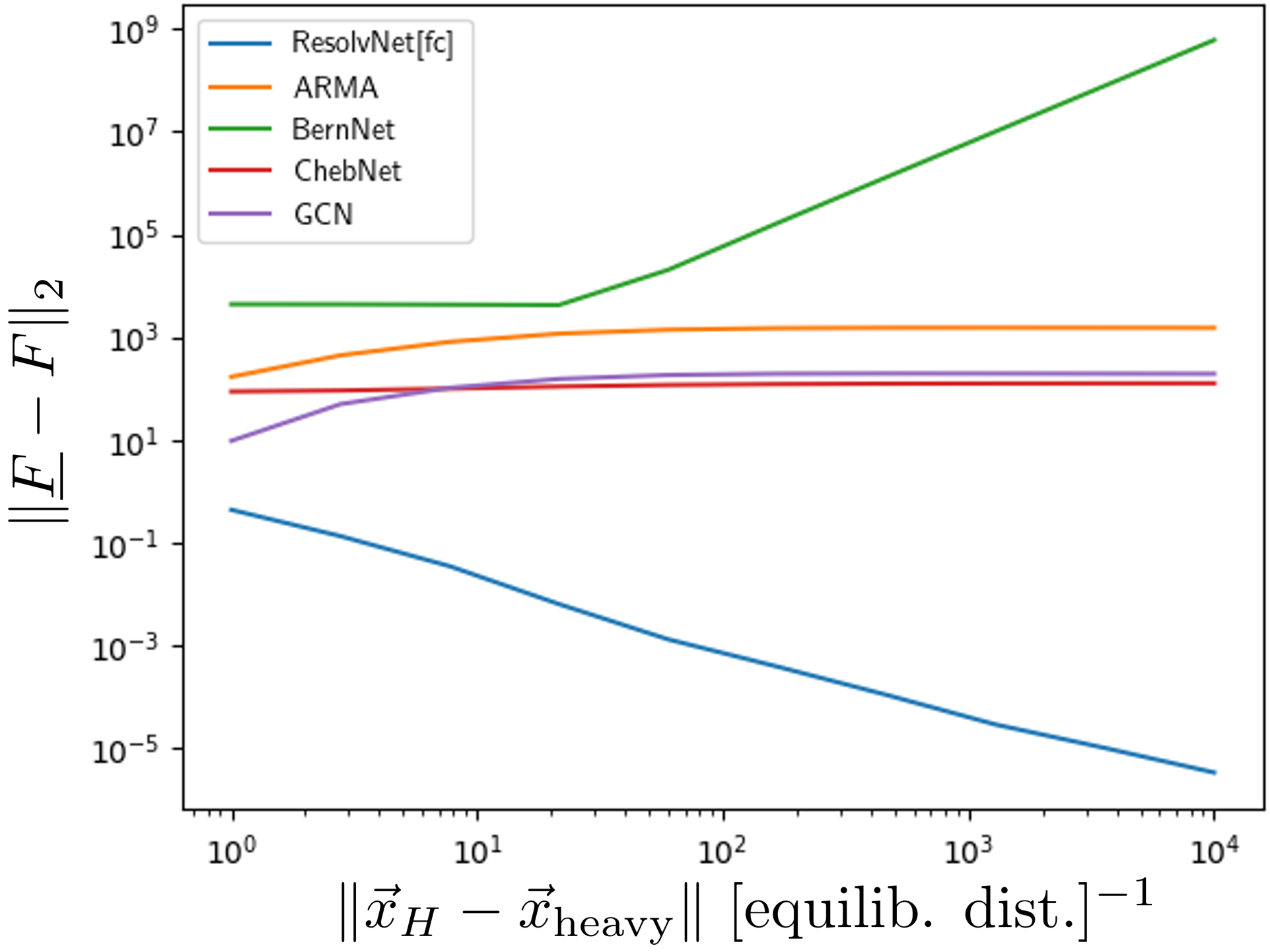}
		\vspace{-6mm}
		\noindent
		\label{collapse_graph}
	\end{figure}
\end{minipage}\noindent

As a final experiment, we treat the coarse-grained molecular graphs $\{\underline{G}\}$ 
as a model for data obtained from a resolution-limited observation process, 
that is unable to resolve positions of hydrogen
\begin{minipage}{0.69\textwidth}
	and only provides information about how many H-atoms are bound to a given heavy atom.
	Given models trained on higher resolution data,
	atomization energies for such 
	observed molecules 
	are now to be predicted.
	Table \ref{qm7_collapsed_result_table} contains 
	corresponding results. While the performance of baselines decreases significantly  
	if the resolution scale is varied during inference, the prediction accuracy of ResolvNet remains high; even slightly increasing.
	While ResolvNet out-performed baselines by a factor of three on same-resolution-scale data (c.f.Table \ref{qm7_result_table}), its
	lead increases to a factor of $10$ and higher in the multi-scale setting. 
	
\end{minipage}\hfill
\begin{minipage}{0.29\textwidth}
	\vspace{-4mm}
	
	\begin{table}[H]
		\small
		\centering
		\caption{QM$7_{\text{coarse}}$-MAE}
		\begingroup
		\vspace{-3mm}
		\setlength{\tabcolsep}{0.5pt} 
		\renewcommand{\arraystretch}{1} 
		\begin{tabular}{lr}
			\toprule
			\textbf{QM$7$} & MAE $[kcal/mol]$  \\
			\midrule
			BernNet   & $580.67$\tiny{$\pm  99.27$}  \\
			GCN & $124.53 $\tiny{$\pm 34.58$}   \\
			ChebNet & $645.14$\tiny{$\pm 34.59$}\\	
			ARMA & $248.96$\tiny{$\pm 15.56$}   \\
			\midrule	
			ResolvNet &  $\mathbf{16.23}$\tiny{$\pm 2.74$}    \\
		\end{tabular}
		\endgroup
		\label{qm7_collapsed_result_table}
	\end{table}
\end{minipage}

\section{Conclusion}
This work introduced the concept of multi-scale consistency: At the node level this  refers to the retention of a 
propagation scheme not solely determined by the largest given connectivity scale.
At the graph-level it mandates 
that distinct graphs describing the same object at different resolutions should be assigned similar feature vectors. Common GNN architectures were shown to not be multi-scale consistent, while  the newly introduced ResolvNet architecture was theoretically and experimentally established 
to have this property. Deployed on real world data, ResolvNet proved expressive and stable; out-performing baselines 
significantly on many tasks in- and outside the multi-scale setting. 

%
%
%
%

\bibliography{Bibliography}

\begin{thebibliography}{47}
\providecommand{\natexlab}[1]{#1}
\providecommand{\url}[1]{\texttt{#1}}
\expandafter\ifx\csname urlstyle\endcsname\relax
  \providecommand{\doi}[1]{doi: #1}\else
  \providecommand{\doi}{doi: \begingroup \urlstyle{rm}\Url}\fi

\bibitem[Alon \& Yahav(2021)Alon and Yahav]{AlonaTal}
Uri Alon and Eran Yahav.
\newblock On the bottleneck of graph neural networks and its practical
  implications.
\newblock In \emph{International Conference on Learning Representations}, 2021.
\newblock URL \url{https://openreview.net/forum?id=i80OPhOCVH2}.

\bibitem[B{\'a}r{\'a}ny \& Solymosi(2017)B{\'a}r{\'a}ny and
  Solymosi]{circle_thm}
Imre B{\'a}r{\'a}ny and J{\'o}zsef Solymosi.
\newblock \emph{Gershgorin Disks for Multiple Eigenvalues of Non-negative
  Matrices}, pp.\  123--133.
\newblock Springer International Publishing, Cham, 2017.
\newblock \doi{10.1007/978-3-319-44479-6_6}.
\newblock URL \url{https://doi.org/10.1007/978-3-319-44479-6_6}.

\bibitem[Bianchi et~al.(2019)Bianchi, Grattarola, Livi, and Alippi]{ARMA}
Filippo~Maria Bianchi, Daniele Grattarola, Lorenzo~Francesco Livi, and Cesare
  Alippi.
\newblock Graph neural networks with convolutional arma filters.
\newblock \emph{IEEE Transactions on Pattern Analysis and Machine
  Intelligence}, 44:\penalty0 3496--3507, 2019.

\bibitem[Blum \& Reymond(2009)Blum and Reymond]{blum}
L.~C. Blum and J.-L. Reymond.
\newblock 970 million druglike small molecules for virtual screening in the
  chemical universe database {GDB-13}.
\newblock \emph{J. Am. Chem. Soc.}, 131:\penalty0 8732, 2009.

\bibitem[Bodnar et~al.(2022)Bodnar, Giovanni, Chamberlain, Li{\`{o}}, and
  Bronstein]{Sheaf}
Cristian Bodnar, Francesco~Di Giovanni, Benjamin~Paul Chamberlain, Pietro
  Li{\`{o}}, and Michael~M. Bronstein.
\newblock Neural sheaf diffusion: {A} topological perspective on heterophily
  and oversmoothing in gnns.
\newblock \emph{CoRR}, abs/2202.04579, 2022.
\newblock URL \url{https://arxiv.org/abs/2202.04579}.

\bibitem[Brouwer \& Haemers(2012)Brouwer and Haemers]{brouwer12}
Andries~E. Brouwer and Willem~H. Haemers.
\newblock \emph{Spectra of Graphs}.
\newblock New York, NY, 2012.
\newblock \doi{10.1007/978-1-4614-1939-6}.

\bibitem[Chung(1997)]{Chung:1997}
F.~R.~K. Chung.
\newblock \emph{Spectral Graph Theory}.
\newblock American Mathematical Society, 1997.

\bibitem[Defferrard et~al.(2016)Defferrard, Bresson, and
  Vandergheynst]{Bresson}
Micha{\"e}l Defferrard, Xavier Bresson, and Pierre Vandergheynst.
\newblock Convolutional neural networks on graphs with fast localized spectral
  filtering.
\newblock \emph{Advances in neural information processing systems}, 29, 2016.

\bibitem[Fey \& Lenssen(2019)Fey and Lenssen]{pyg}
Matthias Fey and Jan~E. Lenssen.
\newblock Fast graph representation learning with {PyTorch Geometric}.
\newblock In \emph{ICLR Workshop on Representation Learning on Graphs and
  Manifolds}, 2019.

\bibitem[Gao et~al.(2023)Gao, Zheng, Li, Li, Qin, Piao, Quan, Chang, Jin, He,
  and Li]{gao2023survey}
Chen Gao, Yu~Zheng, Nian Li, Yinfeng Li, Yingrong Qin, Jinghua Piao, Yuhan
  Quan, Jianxin Chang, Depeng Jin, Xiangnan He, and Yong Li.
\newblock A survey of graph neural networks for recommender systems:
  Challenges, methods, and directions, 2023.

\bibitem[Gasteiger et~al.(2019{\natexlab{a}})Gasteiger, Bojchevski, and
  G{\"{u}}nnemann]{PPNP}
Johannes Gasteiger, Aleksandar Bojchevski, and Stephan G{\"{u}}nnemann.
\newblock Predict then propagate: Graph neural networks meet personalized
  pagerank.
\newblock In \emph{7th International Conference on Learning Representations,
  {ICLR} 2019, New Orleans, LA, USA, May 6-9, 2019}. OpenReview.net,
  2019{\natexlab{a}}.
\newblock URL \url{https://openreview.net/forum?id=H1gL-2A9Ym}.

\bibitem[Gasteiger et~al.(2019{\natexlab{b}})Gasteiger, Bojchevski, and
  G{\"{u}}nnemann]{PredictThenPropagate}
Johannes Gasteiger, Aleksandar Bojchevski, and Stephan G{\"{u}}nnemann.
\newblock Predict then propagate: Graph neural networks meet personalized
  pagerank.
\newblock In \emph{7th International Conference on Learning Representations,
  {ICLR} 2019, New Orleans, LA, USA, May 6-9, 2019}. OpenReview.net,
  2019{\natexlab{b}}.
\newblock URL \url{https://openreview.net/forum?id=H1gL-2A9Ym}.

\bibitem[Gilmer et~al.(2017)Gilmer, Schoenholz, Riley, Vinyals, and
  Dahl]{mpnncm}
Justin Gilmer, Samuel~S Schoenholz, Patrick~F Riley, Oriol Vinyals, and
  George~E Dahl.
\newblock Neural message passing for quantum chemistry.
\newblock In \emph{International conference on machine learning}, pp.\
  1263--1272. PMLR, 2017.

\bibitem[Hamilton et~al.(2017)Hamilton, Ying, and Leskovec]{SAGE}
William~L. Hamilton, Zhitao Ying, and Jure Leskovec.
\newblock Inductive representation learning on large graphs.
\newblock In Isabelle Guyon, Ulrike von Luxburg, Samy Bengio, Hanna~M. Wallach,
  Rob Fergus, S.~V.~N. Vishwanathan, and Roman Garnett (eds.), \emph{Advances
  in Neural Information Processing Systems 30: Annual Conference on Neural
  Information Processing Systems 2017, December 4-9, 2017, Long Beach, CA,
  {USA}}, pp.\  1024--1034, 2017.
\newblock URL
  \url{https://proceedings.neurips.cc/paper/2017/hash/5dd9db5e033da9c6fb5ba83c7a7ebea9-Abstract.html}.

\bibitem[He et~al.(2021)He, Wei, Huang, and Xu]{bernnet}
Mingguo He, Zhewei Wei, Zengfeng Huang, and Hongteng Xu.
\newblock Bernnet: Learning arbitrary graph spectral filters via bernstein
  approximation.
\newblock In Marc'Aurelio Ranzato, Alina Beygelzimer, Yann~N. Dauphin, Percy
  Liang, and Jennifer~Wortman Vaughan (eds.), \emph{Advances in Neural
  Information Processing Systems 34: Annual Conference on Neural Information
  Processing Systems 2021, NeurIPS 2021, December 6-14, 2021, virtual}, pp.\
  14239--14251, 2021.
\newblock URL
  \url{https://proceedings.neurips.cc/paper/2021/hash/76f1cfd7754a6e4fc3281bcccb3d0902-Abstract.html}.

\bibitem[Horn \& Johnson(2012)Horn and Johnson]{horn}
Roger~A Horn and Charles~R Johnson.
\newblock \emph{Matrix analysis}.
\newblock Cambridge university press, 2012.

\bibitem[Hu et~al.(2020)Hu, Fey, Zitnik, Dong, Ren, Liu, Catasta, and
  Leskovec]{OGBN}
Weihua Hu, Matthias Fey, Marinka Zitnik, Yuxiao Dong, Hongyu Ren, Bowen Liu,
  Michele Catasta, and Jure Leskovec.
\newblock Open graph benchmark: Datasets for machine learning on graphs.
\newblock In Hugo Larochelle, Marc'Aurelio Ranzato, Raia Hadsell,
  Maria{-}Florina Balcan, and Hsuan{-}Tien Lin (eds.), \emph{Advances in Neural
  Information Processing Systems 33: Annual Conference on Neural Information
  Processing Systems 2020, NeurIPS 2020, December 6-12, 2020, virtual}, 2020.
\newblock URL
  \url{https://proceedings.neurips.cc/paper/2020/hash/fb60d411a5c5b72b2e7d3527cfc84fd0-Abstract.html}.

\bibitem[Kato(1976)]{Kato}
Tosio Kato.
\newblock \emph{{Perturbation theory for linear operators; 2nd ed.}}
\newblock Grundlehren der mathematischen Wissenschaften : a series of
  comprehensive studies in mathematics. Springer, Berlin, 1976.
\newblock URL \url{https://cds.cern.ch/record/101545}.

\bibitem[Kipf \& Welling(2017)Kipf and Welling]{Kipf}
Thomas~N. Kipf and Max Welling.
\newblock Semi-supervised classification with graph convolutional networks.
\newblock In \emph{5th International Conference on Learning Representations,
  {ICLR} 2017, Toulon, France, April 24-26, 2017, Conference Track
  Proceedings}. OpenReview.net, 2017.
\newblock URL \url{https://openreview.net/forum?id=SJU4ayYgl}.

\bibitem[Koke(2023)]{limitless}
Christian Koke.
\newblock Limitless stability for graph convolutional networks.
\newblock In \emph{11th International Conference on Learning Representations,
  {ICLR} 2023, Kigali, Rwanda, May 1-5, 2023}. OpenReview.net, 2023.
\newblock URL \url{https://openreview.net/forum?id=XqcQhVUr2h0}.

\bibitem[Levie et~al.(2019)Levie, Bronstein, and Kutyniok]{levie}
Ron Levie, Michael~M. Bronstein, and Gitta Kutyniok.
\newblock Transferability of spectral graph convolutional neural networks.
\newblock \emph{CoRR}, abs/1907.12972, 2019.
\newblock URL \url{http://arxiv.org/abs/1907.12972}.

\bibitem[Li et~al.(2018{\natexlab{a}})Li, Han, and Wu]{oversmoothing18}
Qimai Li, Zhichao Han, and Xiao{-}Ming Wu.
\newblock Deeper insights into graph convolutional networks for semi-supervised
  learning.
\newblock In Sheila~A. McIlraith and Kilian~Q. Weinberger (eds.),
  \emph{Proceedings of the Thirty-Second {AAAI} Conference on Artificial
  Intelligence, (AAAI-18), the 30th innovative Applications of Artificial
  Intelligence (IAAI-18), and the 8th {AAAI} Symposium on Educational Advances
  in Artificial Intelligence (EAAI-18), New Orleans, Louisiana, USA, February
  2-7, 2018}, pp.\  3538--3545. {AAAI} Press, 2018{\natexlab{a}}.
\newblock \doi{10.1609/aaai.v32i1.11604}.
\newblock URL \url{https://doi.org/10.1609/aaai.v32i1.11604}.

\bibitem[Li et~al.(2018{\natexlab{b}})Li, Yu, Shahabi, and
  Liu]{diffusion_traffic_forecasting}
Yaguang Li, Rose Yu, Cyrus Shahabi, and Yan Liu.
\newblock Diffusion convolutional recurrent neural network: Data-driven traffic
  forecasting.
\newblock In \emph{6th International Conference on Learning Representations,
  {ICLR} 2018, Vancouver, BC, Canada, April 30 - May 3, 2018, Conference Track
  Proceedings}. OpenReview.net, 2018{\natexlab{b}}.
\newblock URL \url{https://openreview.net/forum?id=SJiHXGWAZ}.

\bibitem[Maskey et~al.(2021)Maskey, Levie, and
  Kutyniok]{DBLP:journals/corr/abs-2109-10096}
Sohir Maskey, Ron Levie, and Gitta Kutyniok.
\newblock Transferability of graph neural networks: an extended graphon
  approach.
\newblock \emph{CoRR}, abs/2109.10096, 2021.
\newblock URL \url{https://arxiv.org/abs/2109.10096}.

\bibitem[McCallum et~al.(2000)McCallum, Nigam, Rennie, and Seymore]{Cora}
Andrew McCallum, Kamal Nigam, Jason Rennie, and Kristie Seymore.
\newblock Automating the construction of internet portals with machine
  learning.
\newblock \emph{Inf. Retr.}, 3\penalty0 (2):\penalty0 127--163, 2000.
\newblock \doi{10.1023/A:1009953814988}.
\newblock URL \url{https://doi.org/10.1023/A:1009953814988}.

\bibitem[Namata et~al.(2012)Namata, London, Getoor, and Huang]{Pubmed}
Galileo Namata, Ben London, Lise Getoor, and Bert Huang.
\newblock Query-driven active surveying for collective classification.
\newblock 2012.

\bibitem[Oono \& Suzuki(2020)Oono and Suzuki]{oversmoothing20}
Kenta Oono and Taiji Suzuki.
\newblock Graph neural networks exponentially lose expressive power for node
  classification.
\newblock In \emph{8th International Conference on Learning Representations,
  {ICLR} 2020, Addis Ababa, Ethiopia, April 26-30, 2020}. OpenReview.net, 2020.
\newblock URL \url{https://openreview.net/forum?id=S1ldO2EFPr}.

\bibitem[Pablo~Gainza(2023)]{bronstein_nature}
et.~al. Pablo~Gainza.
\newblock Deciphering interaction fingerprints from protein molecular surfaces
  using geometric deep learning.
\newblock \emph{Nature}, 2023.

\bibitem[Pei et~al.(2020)Pei, Wei, Chang, Lei, and Yang]{geom}
Hongbin Pei, Bingzhe Wei, Kevin Chen-Chuan Chang, Yu~Lei, and Bo~Yang.
\newblock Geom-gcn: Geometric graph convolutional networks, 2020.
\newblock URL \url{https://arxiv.org/abs/2002.05287}.

\bibitem[Post(2012)]{PostBook}
Olaf. Post.
\newblock \emph{Spectral Analysis on Graph-like Spaces / by Olaf Post.}
\newblock Lecture Notes in Mathematics, 2039. Springer Berlin Heidelberg,
  Berlin, Heidelberg, 1st ed. 2012. edition, 2012.
\newblock ISBN 3-642-23840-8.

\bibitem[Post \& Simmer(2021)Post and Simmer]{graph_approx}
Olaf Post and Jan Simmer.
\newblock Graph-like spaces approximated by discrete graphs and applications.
\newblock \emph{Mathematische Nachrichten}, 294\penalty0 (11):\penalty0
  2237--2278, 2021.
\newblock \doi{https://doi.org/10.1002/mana.201900108}.

\bibitem[Ramakrishnan et~al.(2014)Ramakrishnan, Dral, Rupp, and von
  Lilienfeld]{QM9}
Raghunathan Ramakrishnan, Pavlo~O Dral, Matthias Rupp, and O~Anatole von
  Lilienfeld.
\newblock Quantum chemistry structures and properties of 134 kilo molecules.
\newblock \emph{Scientific Data}, 1, 2014.

\bibitem[Rozemberczki et~al.(2021)Rozemberczki, Allen, and Sarkar]{Squirrel}
Benedek Rozemberczki, Carl Allen, and Rik Sarkar.
\newblock Multi-scale attributed node embedding.
\newblock \emph{J. Complex Networks}, 9\penalty0 (2), 2021.
\newblock \doi{10.1093/comnet/cnab014}.
\newblock URL \url{https://doi.org/10.1093/comnet/cnab014}.

\bibitem[Ruiz et~al.(2020)Ruiz, Chamon, and Ribeiro]{DBLP:conf/nips/RuizCR20}
Luana Ruiz, Luiz F.~O. Chamon, and Alejandro Ribeiro.
\newblock Graphon neural networks and the transferability of graph neural
  networks.
\newblock In Hugo Larochelle, Marc'Aurelio Ranzato, Raia Hadsell,
  Maria{-}Florina Balcan, and Hsuan{-}Tien Lin (eds.), \emph{Advances in Neural
  Information Processing Systems 33: Annual Conference on Neural Information
  Processing Systems 2020, NeurIPS 2020, December 6-12, 2020, virtual}, 2020.
\newblock URL
  \url{https://proceedings.neurips.cc/paper/2020/hash/12bcd658ef0a540cabc36cdf2b1046fd-Abstract.html}.

\bibitem[Rupp et~al.(2012)Rupp, Tkatchenko, M\"uller, and von Lilienfeld]{rupp}
M.~Rupp, A.~Tkatchenko, K.-R. M\"uller, and O.~A. von Lilienfeld.
\newblock Fast and accurate modeling of molecular atomization energies with
  machine learning.
\newblock \emph{Physical Review Letters}, 108:\penalty0 058301, 2012.

\bibitem[Sen et~al.(2008)Sen, Namata, Bilgic, Getoor, Galligher, and
  Eliassi-Rad]{Citeseer}
Prithviraj Sen, Galileo Namata, Mustafa Bilgic, Lise Getoor, Brian Galligher,
  and Tina Eliassi-Rad.
\newblock Collective classification in network data.
\newblock \emph{AI Magazine}, 29\penalty0 (3):\penalty0 93, Sep. 2008.
\newblock \doi{10.1609/aimag.v29i3.2157}.
\newblock URL
  \url{https://ojs.aaai.org/index.php/aimagazine/article/view/2157}.

\bibitem[Shchur et~al.(2018)Shchur, Mumme, Bojchevski, and
  G{\"{u}}nnemann]{MSAcademic}
Oleksandr Shchur, Maximilian Mumme, Aleksandar Bojchevski, and Stephan
  G{\"{u}}nnemann.
\newblock Pitfalls of graph neural network evaluation.
\newblock \emph{CoRR}, abs/1811.05868, 2018.
\newblock URL \url{http://arxiv.org/abs/1811.05868}.

\bibitem[Shlomi et~al.(2021)Shlomi, Battaglia, and Vlimant]{Shlomi_2021}
Jonathan Shlomi, Peter Battaglia, and Jean-Roch Vlimant.
\newblock Graph neural networks in particle physics.
\newblock \emph{Machine Learning: Science and Technology}, 2\penalty0
  (2):\penalty0 021001, jan 2021.
\newblock \doi{10.1088/2632-2153/abbf9a}.
\newblock URL \url{https://doi.org/10.1088%2F2632-2153%2Fabbf9a}.

\bibitem[Tang et~al.(2009)Tang, Sun, Wang, and Yang]{actor}
Jie Tang, Jimeng Sun, Chi Wang, and Zi~Yang.
\newblock Social influence analysis in large-scale networks.
\newblock In John F.~Elder IV, Fran{\c{c}}oise Fogelman{-}Souli{\'{e}},
  Peter~A. Flach, and Mohammed~Javeed Zaki (eds.), \emph{Proceedings of the
  15th {ACM} {SIGKDD} International Conference on Knowledge Discovery and Data
  Mining, Paris, France, June 28 - July 1, 2009}, pp.\  807--816. {ACM}, 2009.
\newblock \doi{10.1145/1557019.1557108}.
\newblock URL \url{https://doi.org/10.1145/1557019.1557108}.

\bibitem[Teschl(2014)]{Teschl}
Gerald Teschl.
\newblock \emph{Mathematical Methods in Quantum Mechanics}.
\newblock American Mathematical Society, 2014.

\bibitem[Topping et~al.(2021)Topping, Di~Giovanni, Chamberlain, Dong, and
  Bronstein]{BronsteinInBottle}
Jake Topping, Francesco Di~Giovanni, Benjamin~Paul Chamberlain, Xiaowen Dong,
  and Michael~M. Bronstein.
\newblock Understanding over-squashing and bottlenecks on graphs via curvature,
  2021.
\newblock URL \url{https://arxiv.org/abs/2111.14522}.

\bibitem[Velickovic et~al.(2018)Velickovic, Cucurull, Casanova, Romero,
  Li{\`{o}}, and Bengio]{GAT}
Petar Velickovic, Guillem Cucurull, Arantxa Casanova, Adriana Romero, Pietro
  Li{\`{o}}, and Yoshua Bengio.
\newblock Graph attention networks.
\newblock In \emph{6th International Conference on Learning Representations,
  {ICLR} 2018, Vancouver, BC, Canada, April 30 - May 3, 2018, Conference Track
  Proceedings}. OpenReview.net, 2018.
\newblock URL \url{https://openreview.net/forum?id=rJXMpikCZ}.

\bibitem[Wu et~al.(2021)Wu, Pan, Chen, Long, Zhang, and Yu]{Wu_2021}
Zonghan Wu, Shirui Pan, Fengwen Chen, Guodong Long, Chengqi Zhang, and
  Philip~S. Yu.
\newblock A comprehensive survey on graph neural networks.
\newblock \emph{{IEEE} Transactions on Neural Networks and Learning Systems},
  32\penalty0 (1):\penalty0 4--24, jan 2021.
\newblock \doi{10.1109/tnnls.2020.2978386}.
\newblock URL \url{https://doi.org/10.1109%2Ftnnls.2020.2978386}.

\bibitem[Xiao et~al.(2022)Xiao, Wang, Dai, and
  Guo]{DBLP:journals/mva/XiaoWDG22}
Shunxin Xiao, Shiping Wang, Yuanfei Dai, and Wenzhong Guo.
\newblock Graph neural networks in node classification: survey and evaluation.
\newblock \emph{Mach. Vis. Appl.}, 33\penalty0 (1):\penalty0 4, 2022.
\newblock \doi{10.1007/s00138-021-01251-0}.
\newblock URL \url{https://doi.org/10.1007/s00138-021-01251-0}.

\bibitem[Xu et~al.(2019)Xu, Hu, Leskovec, and Jegelka]{GIN}
Keyulu Xu, Weihua Hu, Jure Leskovec, and Stefanie Jegelka.
\newblock How powerful are graph neural networks?
\newblock In \emph{7th International Conference on Learning Representations,
  {ICLR} 2019, New Orleans, LA, USA, May 6-9, 2019}. OpenReview.net, 2019.
\newblock URL \url{https://openreview.net/forum?id=ryGs6iA5Km}.

\bibitem[Zhang et~al.(2022)Zhang, Wang, Muniz, Panagiotopoulos, Car, and
  E]{Zhang_2022}
Linfeng Zhang, Han Wang, Maria~Carolina Muniz, Athanassios~Z. Panagiotopoulos,
  Roberto Car, and Weinan E.
\newblock A deep potential model with long-range electrostatic interactions.
\newblock \emph{The Journal of Chemical Physics}, 156\penalty0 (12), mar 2022.
\newblock \doi{10.1063/5.0083669}.
\newblock URL \url{https://doi.org/10.1063%2F5.0083669}.

\bibitem[Zhang et~al.(2021)Zhang, He, Brugnone, Perlmutter, and Hirn]{magnet}
Xitong Zhang, Yixuan He, Nathan Brugnone, Michael Perlmutter, and Matthew~J.
  Hirn.
\newblock Magnet: {A} neural network for directed graphs.
\newblock In Marc'Aurelio Ranzato, Alina Beygelzimer, Yann~N. Dauphin, Percy
  Liang, and Jennifer~Wortman Vaughan (eds.), \emph{Advances in Neural
  Information Processing Systems 34: Annual Conference on Neural Information
  Processing Systems 2021, NeurIPS 2021, December 6-14, 2021, virtual}, pp.\
  27003--27015, 2021.
\newblock URL
  \url{https://proceedings.neurips.cc/paper/2021/hash/e32084632d369461572832e6582aac36-Abstract.html}.

\end{thebibliography}


\begin{thebibliography}{3}
\providecommand{\natexlab}[1]{#1}
\providecommand{\url}[1]{\texttt{#1}}
\expandafter\ifx\csname urlstyle\endcsname\relax
  \providecommand{\doi}[1]{doi: #1}\else
  \providecommand{\doi}{doi: \begingroup \urlstyle{rm}\Url}\fi

\bibitem[Bengio \& LeCun(2007)Bengio and LeCun]{Bengio+chapter2007}
Yoshua Bengio and Yann LeCun.
\newblock Scaling learning algorithms towards {AI}.
\newblock In \emph{Large Scale Kernel Machines}. MIT Press, 2007.

\bibitem[Goodfellow et~al.(2016)Goodfellow, Bengio, Courville, and
  Bengio]{goodfellow2016deep}
Ian Goodfellow, Yoshua Bengio, Aaron Courville, and Yoshua Bengio.
\newblock \emph{Deep learning}, volume~1.
\newblock MIT Press, 2016.

\bibitem[Hinton et~al.(2006)Hinton, Osindero, and Teh]{Hinton06}
Geoffrey~E. Hinton, Simon Osindero, and Yee~Whye Teh.
\newblock A fast learning algorithm for deep belief nets.
\newblock \emph{Neural Computation}, 18:\penalty0 1527--1554, 2006.

\end{thebibliography}
\bibliographystyle{iclr2024_conference}

\appendix

\newpage
\section{Effective Propagation Schemes}\label{limitprop}

For definiteness, we here discuss limit-propagation schemes in the setting where \textbf{edge-weights} are large. The discussion for high-connectivity in the Sense of Example II of Section \ref{twoscalegraphs} proceeds in complete analogy.

\ \\

In this section, we then take up again the setting of Section \ref{scales}. We reformulate this setting here in a slightly modified language, that is more adapted to discussing effective propagation schemes of standard architectures:\\

We partition edges on  a weighted graph $G$, into two disjoint sets $\mathcal{E} = \mathcal{E}_{\text{reg.}} \dot{\cup} \mathcal{E}_{\text{high}} $, where the set of edges with large weights is given by: 
\begin{equation}
\mathcal{E}_{\text{high}} := \{(i,j)\in \mathcal{E}:   w_{ij} \geq S_{\text{high}}\}
\end{equation}
and the set with small weights is given by:
\begin{equation}
\mathcal{E}_{\text{reg.}} := \{(i,j)\in \mathcal{E}: w_{ij} \leq S_{\text{reg.}}\}
\end{equation}
for weight scales $S_\text{high} > S_\text{reg.} > 0$. Without loss of generality, assume $S_\text{reg.}$ to be as low as possible (i.e. $S_\text{reg.} = \max_{(i,j)\in \mathcal{E}_\text{reg.}} w_{ij}$) and $S_\text{high}$ to be as high as possible (i.e. $S_\text{large} = \min_{(i,j)\in \mathcal{E}_\text{high}}$) and no weights in between the scales.
\begin{figure}[H]
	(a)\includegraphics[scale=0.27]{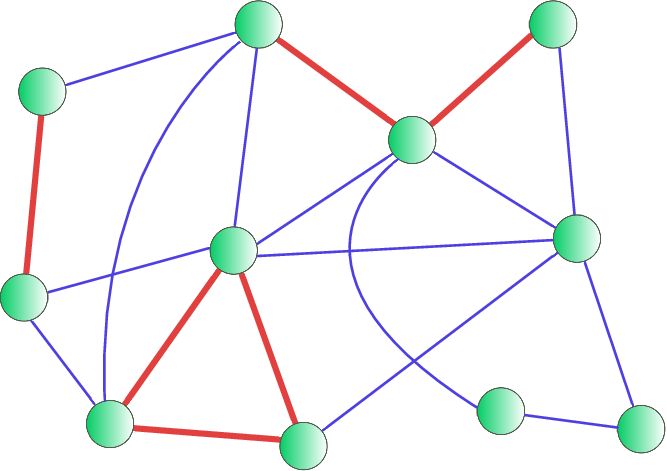}\hfill
	(b)\includegraphics[scale=0.27]{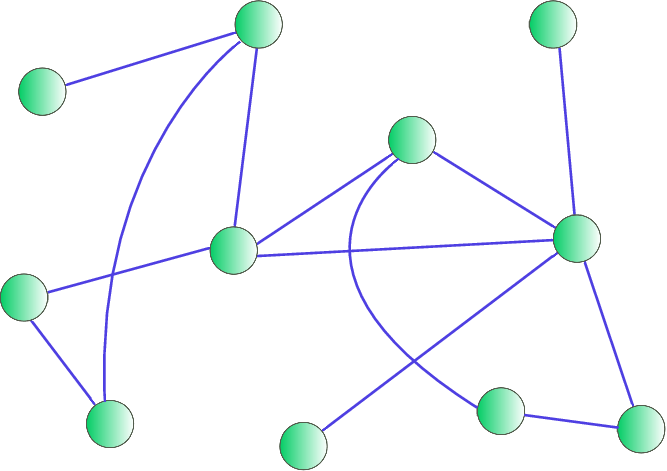}\hfill
	(c)\includegraphics[scale=0.27]{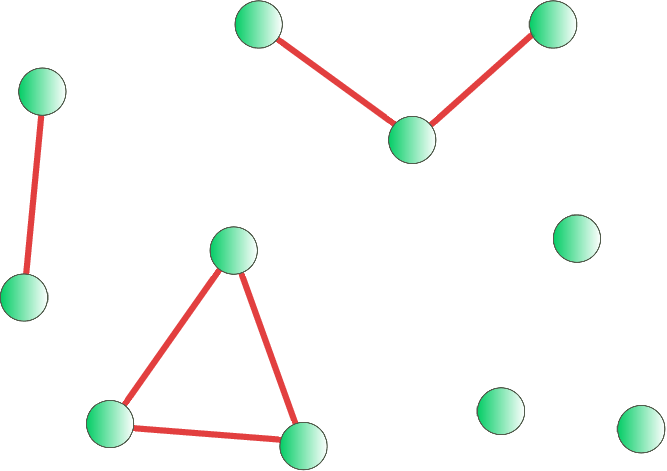}\hfill
	(d)\includegraphics[scale=0.27]{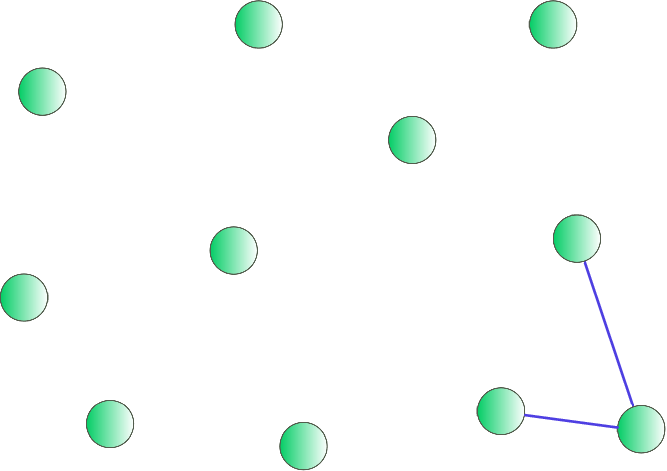}
	\captionof{figure}{(a) Graph $G$ with \textcolor{blue}{$\mathcal{E}_{\text{reg.}}$ (blue)} \& \textcolor{red}{$\mathcal{E}_{\text{high}}$ (red)};\ (b)  $G_{\text{reg.}}$; \ (c) $G_{\text{high}}$;\ (d) $G_{\text{reg., exclusive}}$    } 
	\label{graph_decomp_app}
\end{figure}
This decomposition induces two graph structures corresponding to the disjoint edge sets on the node set $\mathcal{G}$: We set $G_{\text{reg.}} := (\mathcal{G},\mathcal{E}_{\text{reg.}})$ and $G_{\text{high}}:=(\mathcal{G},\mathcal{E}_{\text{high}})$ c.f. Fig.  \ref{graph_decomp_app}).\\
We also introduce the set of edges 
$\mathcal{E}_{\text{reg., exclusive}} := \{(i,j)\in \mathcal{E}_{\text{reg.}}| \ \forall k \in \mathcal{G}:\ (i,k) \notin \mathcal{E}_{\text{high}}\ \&\ (k,j) \notin \mathcal{E}_{\text{high}} \}$
connecting nodes that do not have an incident edge in $\mathcal{E}_{\text{high}}$. A corresponding example-graph $G_{\text{reg., exclusive}}$ is depicted in Fig. \ref{graph_decomp_app} (d).\\
\ \\
We are now interested in the behaviour of graph convolution schemes if the scales are well separated:
\begin{equation}
S_{\text{high}} \gg S_{\text{reg.}}
\end{equation}

\subsection{Spectral Convolutional Filters}
We first discuss resulting limit-propagation schemes for spectral convolutional networks. Such networks implement convolutional filters as a mapping
\begin{equation}
x \longmapsto g_\theta(T)x
\end{equation}
for a node feature $x$, a learnable function $g_\theta$ and a graph shift operator $T$.

\subsubsection{Need for Normalization}\label{spectral_need_f_normalization}
The graph shift operator $T$ facilitating the graph convolutions needs to be normalized for established spectral graph convolutional architectures:

For \cite{ARMA}, this e.g. arises as a necessity for convergence of the proposed implementation scheme for the rational filters introduced there (c.f. eq. (10) in \cite{ARMA}).

The work \cite{Bresson} needs its graph shift operator to be normalized, as it approximates generic filters via a Chebyshev expansion. As argued in \cite{Bresson}, such Chebyshev polynomials form an orthogonal basis for the space $L^2([-1,1], dx/\sqrt{1-x^2})$. Hence, the spectrum of the operator $T$ to which the (approximated and learned) function $g_\theta$ is applied needs to be contained in the interval $[-1,1]$.

In \cite{Kipf}, it has been noted that for the architecture proposed there, choosing $T$ to have eigenvalues in the range $[0,2]$ (as opposed to the normalized ranges $[0,1]$ or $[-1,1]$) has the potential to lead to vanishing- or exploding gradients as well as numerical instabilities. To alleviate this, \cite{Kipf} introduces a "renormalization trick" (c.f. Section 2.2. of \cite{Kipf} to produce a normalized graph shift operator on which the network is then based.

We can understand the relationship between normalization of graph shift operator  $T$ and the stability of corresponding convolutional filters explicitly: Assume that we have 
\begin{equation}
\|T\| \gg 1.
\end{equation}
This might e.g. happen when basing networks on the un-normalized graph Laplacian $\Delta$ or the weight-matrix $W$ if edge weights are potentially large (such as in the setting $S_{\text{high}} \gg S_{\text{reg.}}$ that we are considering).

By the spectral mapping theorem (see e.g. \cite{Teschl}), we have
\begin{equation}\label{specctral_mapping}
\sigma\left( g_\theta(T)\right) = \left\{g_\theta(\lambda) : \lambda \in \sigma(T)\right\},
\end{equation}
with $\sigma(T)$ denoting the spectrum (i.e. the set of eigenvalues) of $T$. For the largest (in absolute value) eigenvalue $\lambda_{\max}$ of $T$, we have
\begin{equation}\label{sp_norm_a_l_ev}
|\lambda_{\max}| = \|T\|.
\end{equation}
Since learned functions are either implemented directly as a polynomial (as e.g. in \cite{Bresson, bernnet}) or approximated as a Neumann type power iteration (as e.g. in \cite{ARMA,PPNP}) which can be thought of as a polynomial, we have 
\begin{equation}
\lim\limits_{\lambda \rightarrow \pm \infty}  |g_\theta(\lambda)| = \infty.
\end{equation}
Thus in view of (\ref{specctral_mapping}) and (\ref{sp_norm_a_l_ev})  we have for $\|T\|$ sufficiently large, that
\begin{equation}
\|g_\theta(T)\| =  |g_\theta(\pm\|T\|)|   
\end{equation}
with the sign $\pm$ determined by $\lambda_{\max}\gtrless 0$. Since non-constant polynomials behave at least linearly for large inputs, there is a constant $C>0$ such that 
\begin{equation}
C\cdot\|T\| \leq\|g_\theta(T)\|
\end{equation}
for all sufficiently large $\|T\|$.	We thus have the estimate
\begin{equation}
\|x\|\cdot C\cdot \|T\| \leq  \|g_\theta(T)x\|
\end{equation}
for at least one input signal $x$ (more precisely all $x$ in the eigen-space corresponding to the largest (in absolute value) eigenvalue $\lambda_{\max}$). Thus if $T$ is not normalized (i.e. $\|T\|$ is not sufficiently bounded), the norm of 
(hidden) features might increase drastically when moving from one (hidden) layer to the next. This behaviour persists for all input signals $x$  have components in eigenspaces corresponding to large (in absolute value) eigenvalues of $T$.


\subsubsection{Spectral Normalizations}\label{spectral_n_edges}
\begin{minipage}{0.6\textwidth}
	As discussed in the previous Section \ref{spectral_need_f_normalization}, instabilities arising from non-normalized graph shift operators can be traced back to the problem of such operators having large eigenvalues. It was thus -- among other considerations -- suggested in \cite{Bresson} to  base convolutional filters on the spectrally normalized graph shift operator
	\begin{equation}
	T = \frac{1}{\lambda_{\max}(\Delta)}\Delta,
	\end{equation}
	
\end{minipage}\hfill
\begin{minipage}{0.36\textwidth}
	\begin{figure}[H]
		\includegraphics[scale=0.3]{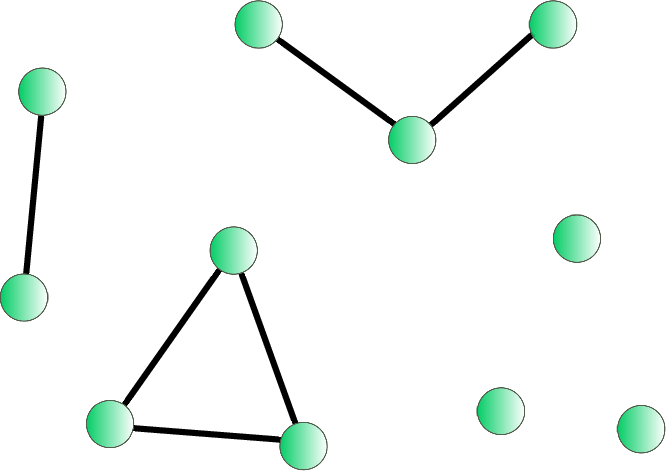}\hfill
		
		\captionof{figure}{Limit graph corresponding to Fig \ref{graph_decomp_app} for spectral  normalization 
		} 
		\label{spectral_norm_limit_app}
	\end{figure}
\end{minipage}

with $\Delta$ the un-normalized graph Laplacian. In the setting $S_{\text{high}} \gg S_{\text{reg.}}$ we are considering, this leads to an effective feature propagation along $G_{\text{high}}$ (c.f. also Fig. \ref{spectral_norm_limit_app}) only, as Theorem \ref{spectr_norm_limit_prop} below establishes:
\ \\

\begin{Thm}\label{spectr_norm_limit_prop}
	With 
	\begin{equation}
	T = \frac{1}{\lambda_{\max}(\Delta)}\Delta,
	\end{equation}
	and the scale decomposition as introduced in Section \ref{scales}, 	we have that
	\begin{equation}\label{sp_norm_to_establish}
	\left\|T -  \frac{1}{\lambda_{\max}(\Delta_{\text{high}})}\Delta_{\text{high}} \right\| = \mathcal{O}\left(\frac{S_{\text{reg.}}}{S_{\text{high}}} \right)
	\end{equation}
	for  $S_{\text{high}} \gg S_{\text{reg.}}$.
\end{Thm}
\begin{proof}
	For convenience in notation, let us write 
	\begin{equation}
	T_{\text{high}} = \frac{1}{\lambda_{\max}(\Delta_{\text{high}})}\Delta_{\text{high}}
	\end{equation}	
	and similarly 
	\begin{equation}
	T_{\text{reg.}} = \frac{1}{\lambda_{\max}(\Delta_{\text{reg.}})}\Delta_{\text{reg.}}.
	\end{equation}	
	In section \ref{scales}, we already noted that 
	\begin{equation}
	\Delta = \Delta_{\text{high}} + \Delta_{\text{reg.}},
	\end{equation}
	which we may rewrite as
	\begin{equation}\label{too_many_sp_norm_eqs}
	\Delta = \lambda_{\max}(\Delta_{\text{high}})\cdot \left(T_{\text{high}} + \frac{\lambda_{\max}(\Delta_{\text{reg.}})}{\lambda_{\max}(\Delta_{\text{high}})}\cdot T_{\text{reg.}} \right).
	\end{equation}
	Let us consider the equivalent expression 
	\begin{equation}\label{to_apply_Kato}
	\frac{1}{\lambda_{\max}(\Delta_{\text{high}})}\cdot \Delta = T_{\text{high}} + \frac{\lambda_{\max}(\Delta_{\text{reg.}})}{\lambda_{\max}(\Delta_{\text{high}})}\cdot T_{\text{reg.}}.
	\end{equation}
	We next note that
	\begin{equation}\label{sp_n_ev_division}
	\lambda_{\max}\left(  \frac{1}{\lambda_{\max}(\Delta_{\text{high}})}\cdot \Delta \right) = \frac{\lambda_{\max}(\Delta)}{\lambda_{\max}(\Delta_{\text{high}})}.
	\end{equation} 
	and
	\begin{equation}
	\lambda_{\max}\left(T_{\text{high}}\right) = 1
	\end{equation}
	since the operation of taking eigenvalues of operators is multiplicative in the sense of
	\begin{equation}
	\lambda_{\max}(|a|\cdot T) = |a|\cdot \lambda_{\max}(T)
	\end{equation}
	for non-negative $|a|\geq 0$.
	
	Since the right-hand-side of (\ref{to_apply_Kato}) constitutes an analytic  perturbation of $T_{\text{high}}$, we may apply analytic perturbation theory (c.f. e.g. \cite{Kato} for an extensive discussion) to this problem.  With this (together with $\|T_{\text{high}}\| = 1$) we find
	\begin{equation}\label{kato_sp_norm}
	\lambda_{\max}\left( \frac{1}{\lambda_{\max}(\Delta_{\text{high}})}\cdot\Delta \right) = 1 + \mathcal{O}\left( \frac{\lambda_{\max}(\Delta_{\text{reg.}})}{\lambda_{\max}(\Delta_{\text{high}})}\right).
	\end{equation}
	Using (\ref{sp_n_ev_division}) and the fact that 
	\begin{equation}\label{spec_norm_ev_rel}
	\frac{\lambda_{\max}(\Delta_{\text{reg.}})}{\lambda_{\max}(\Delta_{\text{high}})} \propto  \frac{S_{\text{reg.}}}{S_{\text{high}}},
	\end{equation}
	we thus have
	\begin{equation}
	\frac{\lambda_{\max}\left(\Delta \right) }{\lambda_{\max}(\Delta_{\text{high}})}  =1+ \mathcal{O}\left( \frac{S_{\text{reg.}}}{S_{\text{high}}}\right).
	\end{equation} 
	Since for small $\epsilon$, we also have
	\begin{equation}
	\frac{1}{1+\epsilon} = 1 + \mathcal{O}(\epsilon),
	\end{equation}
	the relation (\ref{spec_norm_ev_rel}) also implies
	\begin{equation}
	\frac{\lambda_{\max}(\Delta_{\text{high}})}{\lambda_{\max}\left(\Delta \right) }  =1+ \mathcal{O}\left( \frac{S_{\text{reg.}}}{S_{\text{high}}}\right).
	\end{equation} 
	Multiplying (\ref{too_many_sp_norm_eqs}) with $1/\lambda_{\max}(\Delta)$ yields
	\begin{align}\label{sp_norm_another_one}
	T = \frac{\lambda_{\max}(\Delta_{\text{high}})}{\lambda_{\max}(\Delta)}\cdot \left(T_{\text{high}} + \frac{\lambda_{\max}(\Delta_{\text{reg.}})}{\lambda_{\max}(\Delta_{\text{high}})}\cdot T_{\text{reg.}} \right).
	\end{align}
	Since $\|T_{\text{high}}\|,\|T_{\text{reg.}}\|=1$ and 
	\begin{equation}
	\frac{\lambda_{\max}(\Delta_{\text{reg.}})}{\lambda_{\max}(\Delta_{\text{high}})} \propto \frac{S_{\text{reg.}}}{S_{\text{high}}} < 1
	\end{equation}
	for sufficiently large $S_{\text{high}}$, relation (\ref{sp_norm_another_one}) implies 
	\begin{equation}
	\left\|T -  \frac{1}{\lambda_{\max}(\Delta_{\text{high}})}\Delta_{\text{high}} \right\| = \mathcal{O}\left(\frac{S_{\text{reg.}}}{S_{\text{high}}} \right)
	\end{equation}
	as desired.
	
	Note that we might in principle also make use of Lemma \ref{evs_lipschitz} below,  to provide quantitative bounds:
	Lemma \ref{evs_lipschitz}  states that
	\begin{equation}
	|\lambda_k(A) - \lambda_k(B)| \leq \|A-B\|
	\end{equation}
	for self-adjoint operators $A$ and $B$ and their respective $k^{\text{th}}$ eigenvalues ordered by magnitude. On a graph with $N$ nodes, we clearly have $\lambda_{\max} = \lambda_N$ for eigenvalues of (rescaled) graph Laplacians, since all such eigenvalues are non-negative.
	This implies for the difference $|1 - \lambda_{\max}(\Delta)/\lambda_{\max}(\Delta_{\text{high}})|$ arising in (\ref{kato_sp_norm}) that explicitly
	\begin{align}
	\left|1 - \frac{\lambda_{\max}(\Delta)}{\lambda_{\max}(\Delta_{\text{high}})}\right| & \leq \frac{\lambda_{\max}(\Delta_{\text{reg.}})}{\lambda_{\max}(\Delta_{\text{high}})}.
	\end{align}
	This in turn can then be used to provide a quantitative bound in (\ref{sp_norm_to_establish}). Since we are only interested in the qualitative behaviour for $S_{\text{high}} \gg S_{\text{reg.}}$, we shall however not pursue this further.


\end{proof}

It remains to state and establish Lemma \ref{evs_lipschitz} referenced at the end of the proof of Theorem \ref{spectr_norm_limit_prop}:\\

\begin{Lem}\label{evs_lipschitz}
	Let $A$ and $B$ be two hermitian $n \times n$ dimensional matrices. Denote by
	$\{\lambda_k(M)\}_{k=1}^n$ the eigenvalues of a hermitian matrix in increasing order.\\
	With this we have:
	
	\begin{equation}
	|\lambda_k(A) - \lambda_k(B)| \leq ||A-B||.
	\end{equation}
\end{Lem}
\begin{proof}
	After the redefinition $B\mapsto(-B)$, what we need to prove is
	\begin{equation}
	|\lambda_i(A+B) - \lambda_i(A)| \leq ||B||
	\end{equation}	
	for Hermitian $A,B$. 
	Since we have
	\begin{equation}
	\lambda_i(A) - \lambda_i (A+B) = \lambda_i((A+B) +(-B)) - \lambda_i(A+B)
	\end{equation}
	and $||-B|| = ||B||$ it follows that it suffices to prove
	\begin{equation}
	\lambda_i(A + B) - \lambda_i(A) \leq ||B||
	\end{equation}
	for arbitrary hermitian $A,B$.

	We note that the Courant-Fischer	 $\min-\max$ theorem tells us that if $A$ is an $n \times n$ Hermitian matrix, we have
	\begin{equation}
	\lambda_i(M) = \sup\limits_{\dim(V)=i}\inf\limits_{v\in V, ||v|| =1} v^*Mv.
	\end{equation}
	
	With this we find
	\begin{align}
	\lambda_i(A + B) - \lambda_i(A) &=  \sup\limits_{\dim(V)=i}\inf\limits_{v\in V, ||v|| =1} v^*(A+B)v - \sup\limits_{\dim(V)=i}\inf\limits_{v\in V, ||v|| =1} v^*Av \\
	&\leq  \sup\limits_{\dim(V)=i}\inf\limits_{v\in V, ||v|| =1} v^*Av + \sup\limits_{\dim(V)=i}\inf\limits_{v\in V, ||v|| =1} v^*Bv\\
	& - \sup\limits_{\dim(V)=i}\inf\limits_{v\in V, ||v|| =1} v^*Av \\
	&=\sup\limits_{\dim(V)=i}\inf\limits_{v\in V, ||v|| =1} v^*Bv\\
	&= \sup\limits_{\dim(V)=i}\inf\limits_{v\in V, ||v|| =1} v^*Bv\\
	&\leq \max\limits_{1\leq k \leq n}\{|\lambda_k(B)|\}\\
	&= ||B||.
	\end{align}
\end{proof}

\subsubsection{Symmetric Normalizations}\label{sym_n_edges}
\begin{minipage}{0.6\textwidth}
	Most common spectral graph convolutional networks (such as e.g. \cite{bernnet, ARMA, Bresson}) base the learnable filters that they propose on the symmetrically normalized graph Laplacian
	\begin{equation}
	\mathscr{L} = Id - D^{-\frac12}WD^{-\frac12}.
	\end{equation}
	In the setting $S_{\text{high}} \gg S_{\text{reg.}}$ we are considering, this leads to an effective feature propagation along edges in $\mathcal{E}_{\text{high}}$ and $\mathcal{E}_{\text{low, exclusive}}$ (c.f. also Fig. \ref{sym_norm_limit_app}) only, as Theorem \ref{sym_norm_limit_prop} below establishes:

\end{minipage}\hfill
\begin{minipage}{0.36\textwidth}
	\begin{figure}[H]
		\includegraphics[scale=0.3]{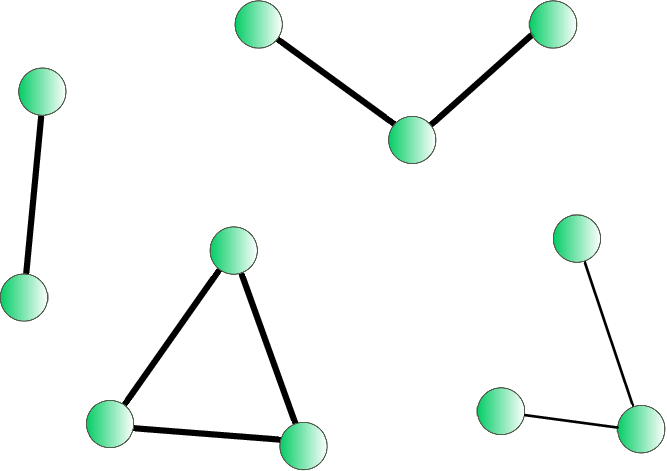}\hfill
		
		\captionof{figure}{Limit graph corresponding to Fig \ref{graph_decomp_app} for symmetric  normalization 
		} 
		\label{sym_norm_limit_app}
	\end{figure}
\end{minipage}

\begin{Thm}\label{sym_norm_limit_prop}
	With 
	\begin{equation}
	T = Id -  D^{-\frac12}WD^{-\frac12},
	\end{equation}
	and the scale decomposition as introduced in Section \ref{scales}, 	we have that
	\begin{equation}\label{sym_norm_to_establish}
	\left\|T - \left(Id - D^{-\frac12}_{\text{high}} W_{\text{high}} D^{-\frac12}_{\text{high}} - D^{-\frac12}_{\text{reg.}} W_{\text{low, exclusive}} D^{-\frac12}_{\text{reg.}}\right)\right\| =
	\mathcal{O}\left(\sqrt{\frac{S_{\text{reg.}}}{S_{\text{high}}}}\right)
	\end{equation}
	for  $S_{\text{high}} \gg S_{\text{reg.}}$.
\end{Thm}
\begin{proof}
	We first note that instead of (\ref{sym_norm_to_establish}), we may equivalently establish 
	\begin{equation}\label{sym_norm_to_establish_actual}
	\left\|D^{-\frac12} W D^{-\frac12}- \left(  D^{-\frac12}_{\text{high}} W_{\text{high}} D^{-\frac12}_{\text{high}} + D^{-\frac12}_{\text{reg.}} W_{\text{low, exclusive}} D^{-\frac12}_{\text{reg.}}\right)\right\| = \mathcal{O}\left(\sqrt{\frac{S_{\text{reg.}}}{S_{\text{high}}}}\right).
	\end{equation}
	In Section \ref{scales}, we already noted that 
	\begin{equation}
	W = W_{\text{high}} + W_{\text{reg.}}.
	\end{equation}		
	With this, we may write
	\begin{equation}\label{sym_norm_deco_second}
	D^{-\frac12} W D^{-\frac12} = D^{-\frac12} W_{\text{high}} D^{-\frac12} + D^{-\frac12} W_{\text{reg.}} D^{-\frac12}.
	\end{equation}
	Let us first examine the term $D^{-\frac12} W_{\text{high}} D^{-\frac12}$. We note for the corresponding matrix entries that
	\begin{equation}
	\left(D^{-\frac12} W_{\text{high}} D^{-\frac12}\right)_{ij} = \frac{1}{\sqrt{d_i}}\cdot (W_{\text{high}})_{ij}\cdot \frac{1}{\sqrt{d_j}}
	\end{equation}
	Let us use the notation
	\begin{equation}
	d_i^\text{high} = \sum\limits_{j=1}^N  (W_{\text{high}})_{ij}, \ \ \ \ d_i^\text{reg.} = \sum\limits_{j=1}^N (W_{\text{reg.}})_{ij}\ \ \text{and} \ \ d_i^\text{low,exclusive} = \sum\limits_{j=1}^N (W_{\text{low,exclusive}})_{ij}.
	\end{equation}
	We then find
	\begin{equation}
	\frac{1}{\sqrt{d_i}} = \frac{1}{\sqrt{d^\text{high}_i}}\cdot \frac{1}{\sqrt{1+\frac{d^\text{reg.}_i}{d^\text{high}_i}}}
	\end{equation}
	Using the Taylor expansion
	\begin{equation}
	\frac{1}{\sqrt{1+\epsilon}} = 1 - \frac{1}{2}\epsilon + \mathcal{O}(\epsilon^2),
	\end{equation}
	we thus have
	\begin{equation}
	\left(D^{-\frac12} W_{\text{high}} D^{-\frac12}\right)_{ij} = \frac{1}{\sqrt{d^{\text{high}}_i}}\cdot (W_{\text{high}})_{ij}\cdot \frac{1}{\sqrt{d^{\text{high}}_j}} + \mathcal{O}\left(\frac{d^\text{reg.}_i}{d^\text{high}_i}\right).
	\end{equation}
	Since we have
	\begin{equation}
	\frac{d^\text{reg.}_i}{d^\text{high}_i} \propto \frac{S_\text{reg.}}{S_\text{high}},
	\end{equation}
	this yields 
	\begin{equation}
	D^{-\frac12} W_{\text{high}} D^{-\frac12} =  D^{-\frac12}_{\text{high}} W_{\text{high}} D^{-\frac12}_{\text{high}} + \mathcal{O}\left(\frac{S_\text{reg.}}{S_\text{high}}\right).
	\end{equation}
	Thus let us turn towards the second summand on the right-hand-side of (\ref{sym_norm_deco_second}).
	We have
	\begin{equation}
	\left( D^{-\frac12} W_{\text{reg.}} D^{-\frac12}\right)_{ij} = \frac{1}{\sqrt{d_i}}\cdot (W_{\text{reg.}})_{ij}. \frac{1}{\sqrt{d_j}}.
	\end{equation}
	Suppose that either $i$ or $j$ is not in $G_{\text{low, exclusive}}$. Without loss of generality (since the matrix under consideration is symmetric), assume $i \notin G_{\text{low, exclusive}}$, but $(W_{\text{reg.}})_{ij} \neq 0$. We may again write
	\begin{equation}
	\frac{1}{\sqrt{ d_j}} = \frac{1}{\sqrt{ d^{\text{high}}_j}} \cdot \frac{1}{\sqrt{1+\frac{d^\text{reg.}_i}{d^\text{high}_i}}}.
	\end{equation}
	Since 
	\begin{equation}
	\frac{1}{\sqrt{1+\frac{d^\text{reg.}_i}{d^\text{high}_i}}} \leq 1,
	\end{equation}
	we have
	\begin{equation}
	\left|\left( D^{-\frac12} W_{\text{reg.}} D^{-\frac12}\right)_{ij}\right| \leq \left|\frac{1}{\sqrt{d_i}}\cdot (W_{\text{reg.}})_{ij}\right|\cdot \frac{1}{\sqrt{d_j^{\text{high}}}} = \mathcal{O}\left(\sqrt{\frac{S_{\text{reg.}}}{S_{\text{high}}}}\right).
	\end{equation}
	If instead we have $i,j \in G_{\text{low, exclusive}}$, then clearly
	\begin{equation}
	\left( D^{-\frac12} W_{\text{reg.}} D^{-\frac12}\right)_{ij} 
	= \left( D^{-\frac12}_{\text{reg.}} W_{\text{low,exclusive}} D^{-\frac12}_{\text{reg.}}\right)_{ij}.
	\end{equation}
	Thus in total we have established
	\begin{equation}
	D^{-\frac12} W D^{-\frac12}  = \left(  D^{-\frac12}_{\text{high}} W_{\text{high}} D^{-\frac12}_{\text{high}} + D^{-\frac12}_{\text{reg.}} W_{\text{low, exclusive}} D^{-\frac12}_{\text{reg.}}\right) + \mathcal{O}\left(\frac{S_{\text{reg.}}}{S_{\text{high}}} \right)
	\end{equation}
	which was to be established.

	%
	%
	%

\end{proof}

Apart from networks that make use of the symmetrically normalized graph Laplacian $\mathscr{L}$, some methods, such as most notably \cite{Kipf}, instead base their filters on the operator	  
\begin{equation}
T = \tilde{D}^{-\frac12}\tilde{W}\tilde{D}^{-\frac12},
\end{equation}
with 
\begin{equation}
\tilde{W} = (W+Id)
\end{equation}
and
\begin{equation}
\tilde{D} = D + Id.
\end{equation}
In analogy to Theorem \ref{sym_norm_limit_prop}, we here establish the limit propagation scheme determined by such operators:

\begin{Thm}\label{sym_norm_limit_prop_GCN}
	With 
	\begin{equation}
	T = \tilde{D}^{-\frac12}\tilde{W}\tilde{D}^{-\frac12},
	\end{equation}
	where $\tilde{W} = (W+Id)$ and $\tilde{D} = D + Id$ as well as the scale decompositionof Section \ref{scales}, 	we have that
	\begin{equation}\label{sym_norm_to_establish_GCN}
	\left\|T - \left( D^{-\frac12}_{\text{high}} W_{\text{high}} D^{-\frac12}_{\text{high}} + D^{-\frac12}_{\text{reg.}} \tilde{W}_{\text{low, exclusive}} D^{-\frac12}_{\text{reg.}}\right)\right\| =
	\mathcal{O}\left(\sqrt{\frac{S_{\text{reg.}}+1}{S_{\text{high}}}}\right)
	\end{equation}
	for  $S_{\text{high}} \gg S_{\text{reg.}}$. Here $\tilde{W}_{\text{low, exclusive}}$ is given as
	\begin{equation}
	\tilde{W}_{\text{low, exclusive}} := W_{\text{low, exclusive}} + \text{diag}\left( \mathds{1}_{G_{\text{low, exclusive}}}\right)
	\end{equation}
	and $\mathds{1}_{G_{\text{low, exclusive}}}$ denotes the vector whose entries are one for nodes in $G_{\text{low, exclusive}}$ and zero for all other nodes.
\end{Thm}
The difference to the result of Theorem \ref{sym_norm_limit_prop} is thus that applicability of the limit propagation scheme of Fig. \ref{sym_norm_limit_app}  for the GCN \cite{Kipf}   is not only contingent upon $S_{\text{high}}\gg S_{\text{reg.}}$ but also  $S_{\text{high}}\gg1$.

\begin{proof}
	To establish this -- as in the proof of Theorem \ref{sym_norm_limit_prop} -- we first decompose $T$:
	\begin{align}\label{kipfs_mode_deco}
	\tilde{D}^{-\frac12}\tilde{W}\tilde{D}^{-\frac12} &= \tilde{D}^{-\frac12}W_{\text{high}}\tilde{D}^{-\frac12} + \tilde{D}^{-\frac12}W_{\text{reg.}}\tilde{D}^{-\frac12} + \tilde{D}^{-\frac12}Id\tilde{D}^{-\frac12}\\
	&= \tilde{D}^{-\frac12}W_{\text{high}}\tilde{D}^{-\frac12} + \tilde{D}^{-\frac12}W_{\text{reg.}}\tilde{D}^{-\frac12} + \tilde{D}^{-1}
	\end{align}
	For the first term, we note
	\begin{equation}
	\left(\tilde{D}^{-\frac12}W_{\text{high}}\tilde{D}^{-\frac12}\right)_{ij} = \frac{1}{\sqrt{d_i+1}}\cdot (W_{\text{high}})_{ij}\cdot \frac{1}{\sqrt{d_j+1}}.
	\end{equation}	
	
	We then find
	\begin{equation}
	\frac{1}{\sqrt{d_i+1}} = \frac{1}{\sqrt{d^\text{high}_i}}\cdot \frac{1}{\sqrt{1+\frac{d^\text{reg.}_i+1}{d^\text{high}_i}}}.
	\end{equation}
	Analogously to the proof of Theorem \ref{sym_norm_limit_prop}, this yields 	
	\begin{equation}
	\left(\tilde D^{-\frac12} W_{\text{high}} \tilde D^{-\frac12}\right)_{ij} = \frac{1}{\sqrt{d^{\text{high}}_i}}\cdot (W_{\text{high}})_{ij}\cdot \frac{1}{\sqrt{d^{\text{high}}_j}} + \mathcal{O}\left(\frac{1+d^\text{reg.}_i}{d^\text{high}_i}\right).
	\end{equation}
	This implies 
	\begin{equation}
	\tilde{D}^{-\frac12} W_{\text{high}} \tilde{D}^{-\frac12} =  D^{-\frac12}_{\text{high}} W_{\text{high}} D^{-\frac12}_{\text{high}} + \mathcal{O}\left(\frac{S_\text{reg.}+1}{S_\text{high}}\right).
	\end{equation}
	Next we turn to the second summand in (\ref{kipfs_mode_deco}):
	
	\begin{equation}
	\left(\tilde D^{-\frac12} W_{\text{reg.}}\tilde D^{-\frac12}\right)_{ij} = \frac{1}{\sqrt{d_i+1}}\cdot (W_{\text{reg.}})_{ij}. \frac{1}{\sqrt{d_j+1}}.
	\end{equation}

	Suppose that either $i$ or $j$ is not in $G_{\text{low, exclusive}}$. Without loss of generality (since the matrix under consideration is symmetric), assume $i \notin G_{\text{low, exclusive}}$, but $(W_{\text{reg.}})_{ij} \neq 0$. We may again write
	\begin{equation}
	\frac{1}{\sqrt{ d_j+1}} = \frac{1}{\sqrt{ d^{\text{high}}_j}} \cdot \frac{1}{\sqrt{1+\frac{d^\text{reg.}_i+1}{d^\text{high}_i}}}.
	\end{equation}
	Since 
	\begin{equation}
	\frac{1}{\sqrt{1+\frac{d^\text{reg.}_i+1}{d^\text{high}_i}}} \leq 1,
	\end{equation}
	we have
	\begin{align}
	\left|\left( D^{-\frac12} W_{\text{reg.}} D^{-\frac12}\right)_{ij}\right| &\leq \left|\frac{1}{\sqrt{1+d_i}}\cdot (W_{\text{reg.}})_{ij}\right|\cdot \frac{1}{\sqrt{d_j^{\text{high}}}}\\
	&\leq \left|\frac{1}{\sqrt{d^{\text{reg.}}_i}}\cdot (W_{\text{reg.}})_{ij}\right|\cdot \frac{1}{\sqrt{d_j^{\text{high}}}}\\
	&= \mathcal{O}\left(\sqrt{\frac{S_{\text{reg.}}}{S_{\text{high}}}}\right).
	\end{align}

	If instead we have $i,j \in G_{\text{low, exclusive}}$, then clearly
	\begin{equation}
	\left( \tilde D^{-\frac12} W_{\text{reg.}} \tilde D^{-\frac12}\right)_{ij} 
	= \left( \tilde D^{-\frac12}_{\text{reg.}} W_{\text{low,exclusive}} \tilde D^{-\frac12}_{\text{reg.}}\right)_{ij}.
	\end{equation}

	Finally we note for the third term on the right-hand-side of (\ref{kipfs_mode_deco}) that
	\begin{equation}
	\frac{1}{d_i}\leq \frac{1}{d^{\text{high}}_i} = \mathcal{O}\left(\frac{1}{S_{\text{high}}}\right)
	\end{equation}
	if $i\notin G_{\text{low, exclusive}}$.

	In total we thus have found
	\begin{equation}
	\tilde{D}^{-\frac12}\tilde{W}\tilde{D}^{-\frac12} = \left( D^{-\frac12}_{\text{high}} W_{\text{high}} D^{-\frac12}_{\text{high}} + D^{-\frac12}_{\text{reg.}} \tilde{W}_{\text{low, exclusive}} D^{-\frac12}_{\text{reg.}}\right) + \mathcal{O}\left(\sqrt{\frac{S_{\text{reg.}}+1}{S_{\text{high}}}}\right);
	\end{equation}
	which was to be proved.
\end{proof}

\subsection{Spatial Convolutional Filters}\label{mpnns}

Apart from spectral methods, there of course also exist methods that purely operate in the spatial domain of the graph. Such methods most often fall into the paradigm of message passing neural networks (MPNNs) \cite{mpnncm,pyg}: 
With  $X^{\ell}_i\in\mathds{R}^F$ denoting the features of node $i$ in layer $\ell$ and $w_{ij}$ denoting edge features, a message passing neural network may be described by the update rule (c.f. \cite{mpnncm})
\begin{equation}\label{mpnn_def_eq}
X^{\ell+1}_i = \gamma\left(X^{\ell}_i, \coprod\limits_{j \in \mathcal{N}(i)} \phi\left(X_i^\ell, X_j^\ell, w_{ij}\right)\right).
\end{equation}
Here $ \mathcal{N}(i)$ denotes the neighbourhood of node $i$,   $\coprod$ denotes a differentiable and permutation invariant function (typically "sum", "mean" or "max") while $\gamma$ and $\phi$ denote differentiable functions such as multi-layer-perceptrons (MLPs) which might not be the same in each layer. 
\cite{pyg}.

Before we discuss corresponding limit-propagation schemes, we first establish that MPNNs are not able to reproduce the limit propagation scheme of Section \ref{resolv_arch} and are thus not stable to scale transitions and topological perturbations as discussed in Theorem \ref{graph_level_top_stab} and Section \ref{graph_level_desire}.

\subsubsection{Scale-Sensitivity of Message Passing Neural Networks}\label{spatial_scale_sensitivity}
As we established in Theorem \ref{varying_spaces} and Theorem \ref{graph_level_top_stab} (c.f. also the corresponding proofs in Appendix \ref{stab_theo} and Appendix \ref{top_stab_theo} respectively), the stability to scale-variations (such as coarse-graining) of ResolvNets arises from the reliance on \textit{resolvents} and the limit propagation scheme that they establish if separated weight-scales are present (c.f. 
Appendix \ref{main_thm_proof} below).

Here we establish that message passing networks (as defined in (\ref{mpnn_def_eq}) above) are unable to emulate this limit propagation scheme. Hence such architectures are also not stable to scale-changing topological perturbations such as coarse-graining procedures.
\noindent

\begin{minipage}{0.6\textwidth}	
	To this end, we consider a simple, fully connected graph $G$ on three nodes labeled $1$, $2$ and $3$ (c.f. Fig. \ref{three_graph}). We assume all node-weights to be equal to one ($\mu_i = 1$ for $i = 1,2,3$) and  edge weights 
	\begin{equation}
	w_{13}, w_{23} \leq S_{\text{reg.}}
	\end{equation}
	as well as 
	\begin{equation}
	w_{12} = S_{\text{high}}.
	\end{equation}
	We now assume $ S_{\text{high}} \gg S_{\text{reg.}}$.
\end{minipage}\hfill
\begin{minipage}{0.36\textwidth}
	\begin{figure}[H]
		\includegraphics[scale=0.6]{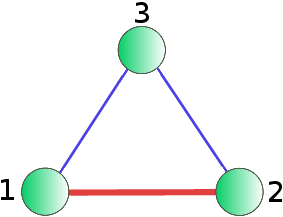}\hfill
		
		\captionof{figure}{Three node Graph $G$ with on large weight $w_{12}\gg1$.
		} 
		\label{three_graph}
	\end{figure}
\end{minipage}
Given states $\{X_1^\ell, X_2^\ell, X_3^\ell\}$ in layer $\ell$, the limit propagation scheme introduced in Section \ref{resolv_arch} would require the updated feature vector of node $3$ to be given by 
\begin{equation}
X_{3, \text{desired}}^{\ell +1} := \gamma\left(X^{\ell}_3, \phi\left(X_3^\ell, \frac{ X_1^\ell+X_2^\ell}{2}, (w_{31}+w_{32})\right) \right)
\end{equation}

However, the actual updated feature at node $3$ is  given as (c.f. (\ref{mpnn_def_eq})):
\begin{equation}\label{node_three_no_scale_dep_rem}
X_{3, \text{actual}}^{\ell +1} := \gamma\left(X^{\ell}_3, \phi\left(X_3^\ell, X_1^\ell, w_{31}\right) \coprod \phi\left(X_3^\ell, X_2^\ell, w_{32}\right)\right)
\end{equation}
Since there is no dependence on $S_{\text{high}}$ in equation (\ref{node_three_no_scale_dep_rem}) -- which defines  $X_{3, \text{actual}}^{\ell +1}$ -- the desired propagation scheme can not arise, unless it is paradoxically already present at all scales $S_{\text{high}}$. If it is present at all scales, there is however only propagation along edges in $\underline{G}$, even if $S_{\text{high}}\approx S_{\text{reg.}}$, which would imply that the message passing network would not respect the graph structure of $G$. 	Hence $X_{3, \text{actual}}^{\ell +1} \nrightarrow X_{3, \text{desired}}^{\ell +1}$ does not converge as $S_{\text{high}}$ increases.

\subsubsection{Limit Propagation Schemes}

%


%

The number of possible choices of message functions $\phi$, aggregation functions $\coprod$ and update functions $\gamma$ is clearly endless. Here we shall exemplarily discuss limit propagation schemes for two popular architectures: We first  discuss the most general case where the message function $\phi$ is given as a learnable perceptron. Subsequently we assume that node features are updated with an attention-type mechanism.

\paragraph{Generic message functions:}
We first consider the possibility that the message function $\phi$ in (\ref{node_three_no_scale_dep_rem}) 
is implemented via an MLP using ReLU-activations: Assuming (for simplicity in notation) a one-hidden-layer MLP mapping features $X_i^\ell \in \mathds{R}^{F_\ell}$ to features $X_i^{\ell+1} \in \mathds{R}^{F_\ell+1}$ 
we have
\begin{equation}
\phi(X^\ell_i,X^\ell_j,w_{ij}) = \text{ReLU}\left(W^\ell_1\cdot X^\ell_i+W^\ell_2\cdot X^\ell_2+ W^\ell_3\cdot w_{ij} +B^\ell    \right)
\end{equation}
with bias term $B^{\ell+1} \in \mathds{R}^{F_{\ell+1}}$ and  weight matrices $W^{\ell+1}_1,W^{\ell+1}_2 \in \mathds{R}^{F_{\ell+1}\times F_{\ell}}$ and $W^\ell_3 \in \mathds{R}^{F_{\ell+1}}$.

We will assume that the weight-vecor $W_3^{\ell+1}$ has no-nonzero entries.
This is not a severe limitation experimentally and in fact generically justified: The complementary event of at-least one entry of $W_3$ being assigned precisely zero during training has probability weight zero (assuming an absolutely continuous probability distribtuion according to which weights are learned).

Let us now assume that the edge $(ij)$ belongs to $\mathcal{E}_{\text{high}}$ and the corresponding weight $w_{ij}$ is large ($w_{ij} \gg 1$). The behaviour of entries $\phi(X^\ell_i,X^\ell_j,w_{ij})_a$ of the message $\phi(X^\ell_i,X^\ell_j,w_{ij}) \in \mathds{R}^{F_{\ell+1}}$ is then determined by the sign of the corresponding entry $\left(W_3^\ell\right)_a$ of the weight vector $W_3^\ell\in \mathds{R}^{F_{\ell+1}}$: 

If we have $\left(W_3^\ell\right)_a<0$, then $\phi(X^\ell_i,X^\ell_j,w_{ij})_a $ approaches zero for larger edge-weights $w_{ij}$:
\begin{equation}\label{uninformative_message}
\lim\limits_{w_{ij}\rightarrow \infty} \phi(X^\ell_i,X^\ell_j,w_{ij})_a = 0
\end{equation} 

If we have $\left(W_3^\ell\right)_a>0$, then $\phi(X^\ell_i,X^\ell_j,w_{ij})_a $ increasingly diverges for larger edge-weights $w_{ij}$:
\begin{equation}\label{dangerous_message}
\lim\limits_{w_{ij}\rightarrow \infty} \phi(X^\ell_i,X^\ell_j,w_{ij})_a = \infty
\end{equation} 
For either choice of  aggregation function  $\coprod$ in (\ref{mpnn_def_eq}) among  "max", "sum" or "mean" the behaviour in (\ref{dangerous_message}) leads to unstable networks if the update function $\gamma$ is also given as an MLP with ReLU activations. Apart from instabilities,
we also make the following observation: If  $S_{\text{high}} \gg S_{\text{reg.}}$, then by (\ref{dangerous_message}) and continuity of $\phi$ we can conclude that components $\phi(X^\ell_i,X^\ell_j,w_{ij})_a $ of messages propagated along $\mathcal{E}_{\text{high}}$  for which  $\left(W_3^\ell\right)_a>0$ dominate over messages propagated along edges in $\mathcal{E}_{\text{reg.}}$. By (\ref{uninformative_message}), the former clearly also dominate over components $\phi(X^\ell_i,X^\ell_j,w_{ij})_a $ of messages propagated along $\mathcal{E}_{\text{high}}$ for which $\left(W_3^\ell\right)_a<0$. This behaviour is irrespective of whether  "max", "sum" or "mean" aggregations are employed. Hence the limit propagation scheme essentially only takes into account message channels $\phi(X^\ell_i,X^\ell_j,w_{ij})_a $ for which $(ij)\in \mathcal{E}_{\text{high}}$ and $\left(W_3^\ell\right)_a>0$.

Similar considerations apply, if non-linearities are chosen as leaky ReLU. 
If instead of ReLU activations a sigmoid-nonlinearity $\sigma$ like $\tanh$ is employed,   messages  propagated along $\mathcal{E}_{\text{large}}$ become increasingly uninformative, since they are progressively more independent of features $X_i^\ell$ and weights $w_{ij}$.  
Indeed, for sigmoid activations, the limits (\ref{uninformative_message}) and (\ref{dangerous_message}) are given as follows:

If we have $\left(W_3^\ell\right)_a<0$, then we have for larger edge-weights $w_{ij}$ that
\begin{equation}\label{uninformative_message_sigma}
\lim\limits_{w_{ij}\rightarrow \infty} \phi(X^\ell_i,X^\ell_j,w_{ij})_a = \lim\limits_{y \rightarrow -\infty} \sigma(y).
\end{equation} 

If we have $\left(W_3^\ell\right)_a>0$, then 
\begin{equation}\label{dangerous_message_sigma}
\lim\limits_{w_{ij}\rightarrow \infty} \phi(X^\ell_i,X^\ell_j,w_{ij})_a = \lim\limits_{y \rightarrow \infty} \sigma(y).
\end{equation} 
In both cases, the messages $\phi(X^\ell_i,X^\ell_j,w_{ij})$ propagated along $\mathcal{E}_{\text{large}}$ become increasingly constant as the scale $S_{\text{high}}$ increases.

%
%
%
%
%
%
%
%
%
%
%
%
%

\paragraph{Attention based messages:}
Apart from general learnable message functions as above, we here also discuss an approach where edge weights are re-learned in an attention based manner. For this we modify the method \cite{GAT} to include edge weights. The resulting propagation scheme -- with a single attention head for simplicity and a non-linearity $\rho$ -- is given as
\begin{equation}
X_i^{\ell +1} = \rho\left(\sum\limits_{j \in \mathcal{N}(i)}\alpha_{ij}(WX_j^{\ell +1})\right).
\end{equation}
Here we have $W \in \mathds{R}^{F_{\ell+1}\times F_\ell}$ and
\begin{equation}\label{att_weight_mat}
\alpha_{ij}=\frac{\exp\left(\text{LeakyRelu}\left(\vec{a}^\top\left[WX_i^\ell\mathbin{\|}WX_j^\ell\mathbin{\|}w_{ij}\right]\right)\right)}{\sum\limits_{k \in \mathcal{N}(i)}\exp\left(\text{LeakyRelu}\left(\vec{a}^\top\left[WX_i^\ell\mathbin{\|}WX_k^\ell\mathbin{\|}w_{ik}\right]\right)\right)},
\end{equation}
with $\mathbin{\|}$ denoting concatenation. The weight vector $\vec{a}\in \mathds{R}^{2F_{\ell+1}+1}$ is assumed to have a non zero entry in its last component. Otherwise, this attention mechanism would correspond to the one proposed in \cite{GAT}, which does not take into account edge weights. Let us denote this entry of $\vec{a}$ ()determining attention on the weight $w_{ij}$)  by $a_w$. 

If  $a_w <0 $, we have for $(i,j)\in \mathcal{E}_{\text{high}}$ that
\begin{equation}
\exp\left(\text{LeakyRelu}\left(\vec{a}^\top\left[WX_i^\ell\mathbin{\|}WX_j^\ell\mathbin{\|}w_{ij}\right]\right) \right) \longrightarrow 0
\end{equation} 
as the weight $w_{ij}$ increases. Thus propagation along edges in $\mathcal{E}_{\text{high}}$ is essentially suppressed in this case.

If  $a_w >0 $, we have for $(i,j)\in \mathcal{E}_{\text{high}}$ that
\begin{equation}
\exp\left(\text{LeakyRelu}\left(\vec{a}^\top\left[WX_i^\ell\mathbin{\|}WX_j^\ell\mathbin{\|}w_{ij}\right]\right) \right) \longrightarrow \infty
\end{equation}
as the weight $w_{ij}$ increases.
Thus for edges $(i,j) \in \mathcal{E}_{\text{reg.}}$ (i.e. those that are \textit{not} in $\mathcal{E}_{\text{high}}$), we have
\begin{equation}
\alpha_{ij} \rightarrow 0,
\end{equation}
since the denominator in (\ref{att_weight_mat}) diverges. Hence in this case, propagation along $\mathcal{E}_{\text{reg.}}$ is essentially suppressed and features are effectively only propagated along $\mathcal{E}_{\text{high}}$.

\section{Proof of  Theorem \ref{main_resolvent_theorem}}\label{main_thm_proof}

In this section, we prove Theorem \ref{main_resolvent_theorem}. For convenience, we first restate the result -- together with the definitions leading up to it -- again:

%
%
%
%
%

\begin{Def}\label{app_lim_graph_def}
	Denote by 	$\underline{\mathcal{G}}$ the set of connected components in $G_{\text{high}}$. We give this set a graph structure as follows: Let $R$ and $P$ be elements of $\underline{\mathcal{G}}$ (i.e. connected components in $G_{\text{high}}$). We define the real number 
	\begin{equation}\label{new_W}
	\underline{W}_{RP} = \sum_{r\in R}\sum_{p\in P} W_{rp},
	\end{equation}
	with $r$ and $p$ nodes in the original graph $G$.
	We define the set of edges $\underline{\mathcal{E}}$ on $\underline{G}$ as  
	\begin{equation}
	\underline{\mathcal{E}} = \{(R,P)\in\underline{\mathcal{G}}\times\underline{\mathcal{G}}: \underline{W}_{RP} >0 \}
	\end{equation}
	and assign $\underline{W}_{RP}$ as weight to such edges.
	Node weights of limit nodes are defined similarly as aggregated weights of all nodes $r$ (in $G$) contained in the component $R$ as
	\begin{equation}
	\underline{\mu}_R = \sum_{r \in R} \mu_r.
	\end{equation}
\end{Def}
In order to translate signals between the original graph $G$ and the limit description $\underline{G}$, we need translation operators mapping signals from one graph to the other:
\begin{Def}
	Denote by $\mathds{1}_R$ the vector that has $1$ as entries on nodes $r$ belonging to the connected (in $G_{\text{hign}}$) component $R$ and has entry zero for all nodes not in $R$. We define the down-projection operator $J^\downarrow$ component-wise via evaluating at node $R$ in $\underline{\mathcal{G}}$ as
	\begin{equation}\label{J_down}
	(J^\downarrow x)_R = \langle \mathds{1}_R, x\rangle/\underline{\mu}_R.
	\end{equation}
	The upsampling operator $J^\uparrow$ 
	is defined as 
	\begin{equation}\label{J_up}
	J^\uparrow u = \sum_R u_R \cdot\mathds{1}_R ;
	\end{equation}
	where $u_R$ is a scalar value (the component entry of $u$ at $R\in \underline{\mathcal{G}}$) and the sum is taken over all connected components in $G_{\text{high}}$.
\end{Def}

The result we then have to prove is the following:

\begin{Thm}\label{main_resolvent_theorem_app}
	We have 
	\begin{equation}\label{resolvent_closeness_app}
	\left\|R_z(\Delta) -  J^\uparrow R_z(\underline{\Delta}) J^\downarrow\right\| =\mathcal{O}\left(  \frac{\|\Delta_{\text{reg.}}\|}{\lambda_1(\Delta_{\text{high}})}\right)
	\end{equation}
	holds; with $\lambda_1(\Delta_{\text{high}})$ denoting the first non-zero eigenvalue of $\Delta_{\text{high}}$.
\end{Thm}
Note that this then indeed proves Theorem \ref{main_resolvent_theorem}, since we have
\begin{equation}
\lambda_{\max}(\Delta_{\text{reg.}}) = \|\Delta_{\text{reg.}}\|.
\end{equation}

\begin{proof}
	We will split the proof of this result into multiple steps.
	For $z<0$ Let us denote by
	\begin{align}
	R_z(\Delta) &= (\Delta - z Id)^{-1},\\
	R_z(\Delta_{\textit{high}}) &= (\Delta_{\textit{high}} - z Id)^{-1}\\
	R_z(\Delta_{\textit{reg.}}) &= (\Delta_{\textit{reg.}} - z Id)^{-1}
	\end{align}
	the resolvents correspodning to $\Delta$, $\Delta_{\textit{high}}$ and $\Delta_{\textit{reg.}}$ respectively.\\
	Our first goal is establishing that we may write
	\begin{equation}
	R_z(\Delta) = \left[Id + R_z(\Delta_{\textit{high}})\Delta_{\textit{reg.}} \right]^{-1}\cdot R_z(\Delta_{\textit{high}})
	\end{equation}
	This will follow as a consequence of what is called the second resolvent formula \cite{Teschl}: 
	
	"Given self-adjoint operators $A,B$, we may write
	\begin{equation}
	R_z(A+B) - R_z(A) =  - R_z(A)BR_z(A+B)."
	\end{equation}

	In our case, this translates to
	\begin{equation}
	R_z(\Delta) - R_z(\Delta_{\textit{high}}) = -R_z(\Delta_{\textit{high}}) \Delta_{\text{reg.}}   R_z(\Delta) 
	\end{equation}
	or equivalently
	\begin{equation}
	\left[Id + R_z(\Delta_{\textit{high}})\Delta_{\text{reg.}} \right]R_z(\Delta) =  R_z(\Delta_{\textit{high}}).
	\end{equation}
	Multiplying with $\left[Id +R_z(\Delta_{\textit{high}}) \Delta_{\text{reg.}} \right]^{-1}$ from the left then yields 
	\begin{equation}
	R_z(\Delta) = \left[Id + R_z(\Delta_{\textit{high}})\Delta_{\textit{reg.}} \right]^{-1}\cdot R_z(\Delta_{\textit{high}})
	\end{equation}
	as desired. \\
	Hence we need to establish that $\left[Id + R_z(\Delta_{\textit{high}}) \Delta_{\textit{reg.}} \right]$ is invertible for $z<0$.\\
	
	To establish a contradiction, assume it is not invertible. Then there is a signal $x$ such that
	\begin{equation}
	\left[Id + R_z(\Delta_{\textit{high}}) \Delta_{\textit{reg.}} \right]x = 0.
	\end{equation}
	Multiplying with $(\Delta_{\text{high}} - z Id)$ from the left yields
	\begin{equation}
	(\Delta_{\text{high}} + \Delta_{\text{reg.}} - z Id)x = 0
	\end{equation}
	which is precisely to say that 
	\begin{equation}
	(\Delta - z Id)x = 0
	\end{equation}
	But since $\Delta$ is a graph Laplacian, it only has non-negative eigenvalues. Hence we have reached our contradiction and established
	\begin{equation}
	R_z(\Delta) = \left[Id +R_z(\Delta_{\textit{high}}) \Delta_{\textit{reg.}} \right]^{-1}R_z(\Delta_{\textit{high}}).
	\end{equation}
	\ \\
	Our next step is to establish that 
	\begin{equation}
	R_z(\Delta_{\textit{high}}) \rightarrow \frac{P^{\text{high}}_0}{-z},
	\end{equation}
	where $P^{\text{high}}_0$ is the spectral projection onto the eigenspace corresponding to the lowest lying eigenvalue $\lambda_0(\Delta_{\textit{high}}) = 0$ of $\Delta_{\textit{high}}$. Indeed, by the spectral theorem for finite dimensional operators (c.f. e.g. \cite{Teschl}), we may write
	\begin{equation}
	R_z(\Delta_{\textit{high}}) \equiv (\Delta_{\textit{high}} - z Id)^{-1} = \sum\limits_{\lambda \in \sigma(\Delta_{\textit{high}})} \frac{1}{\lambda-z}\cdot P^{\textit{high}}_{\lambda}.
	\end{equation}
	Here $\sigma(\Delta_{\textit{high}})$ denotes the spectrum (i.e. the collection of eigenvalues) of $\Delta_{\textit{high}}$ and the $\{P_\lambda^{\textit{high}}\}_{\lambda \in \sigma(\Delta_{\textit{high}})}$ are the corresponding (orthogonal) eigenprojections onto the eigenspaces of the respective eigenvalues.
	Thus we find
	\begin{equation}
	\left\|R_z(\Delta_{\textit{high}}) - \frac{P_0^{\textit{high}}}{-z}\right\| = \left|\sum\limits_{0<\lambda \in \sigma(\Delta_{\textit{high}})} \frac{1}{\lambda-z}\cdot P^{\textit{high}}_{\lambda}\right\|;
	\end{equation}
	where the sum on the right hand side now excludes the eigenvalue $\lambda = 0$.

	Using orthonormality of the spectral projections, the fact that $z <0$ and monotonicity of $1/(\cdot+|z|)$ we find
	\begin{equation}
	\left\|R_z(\Delta_{\textit{high}}) - \frac{P_0^{\textit{high}}}{-z}\right\| = \frac{1}{\lambda_1(\Delta_{\textit{high}})+|z|}.
	\end{equation}
	Here $\lambda_1(\Delta_{\textit{high}})$ is the firt non-zero eigenvalue of $(\Delta_{\textit{high}})$.\\
	Non-zero eigenvalues scale linearly with the weight scale since we have
	\begin{equation}
	\lambda(S\cdot \Delta) = S \cdot \lambda(\Delta)
	\end{equation}
	for any graph Laplacian (in fact any matrix) $\Delta$ with eigenvalue $\lambda$. Thus we have
	\begin{equation}
	\left\|R_z(\Delta_{\textit{high}}) - \frac{P_0^{\textit{high}}}{-z}\right\| = \frac{1}{\lambda_1(\Delta_{\textit{high}})+|z|}  \leq \frac{1}{\lambda_1(\Delta_{\textit{high}})} \longrightarrow 0
	\end{equation}
	as $\lambda_1(\Delta_{\textit{high}})\rightarrow \infty$.\\
	
	Our next task is to use this result in order to bound the difference  
	\begin{equation}
	I := \left\| \left[Id + \frac{P_0^{\textit{high}}}{-z}\Delta_{\textit{reg.}} \right]^{-1}\frac{P_0^{\textit{high}}}{-z} - \left[Id + R_z(\Delta_{\textit{high}})\Delta_{\textit{reg.}} \right]^{-1} R_z(\Delta_{\textit{high}})\right\|.
	\end{equation}
	To this end we first note that the relation
	\begin{equation}
	[A+B-zId]^{-1} = [Id+R_z(A)B]^{-1}R_z(A)
	\end{equation}
	provided to us by the second resolvent formula, implies 
	\begin{equation}
	[Id+R_z(A)B]^{-1} = Id -B [A+B-zId]^{-1} .
	\end{equation}
	Thus we have
	\begin{align}
	\left\|\left[Id + R_z(\Delta_{\textit{high}})\Delta_{\textit{reg.}} \right]^{-1}\right\| &\leq  1 + \|\Delta_{\text{reg.}}\|\cdot\|R_z(\Delta)\|\\
	&\leq 1+ \frac{\|\Delta_{\text{reg.}}\|}{|z|}.
	\end{align}
	With this, we have

	\begin{align}
	&\left\| \left[Id + \frac{P_0^{\textit{high}}}{-z}\Delta_{\textit{reg.}} \right]^{-1}\cdot\frac{P_0^{\textit{high}}}{-z} -R_z(\Delta)\right\|\\
	=&
	\left\| \left[Id + \frac{P_0^{\textit{high}}}{-z}\Delta_{\textit{reg.}} \right]^{-1}\cdot\frac{P_0^{\textit{high}}}{-z} - \left[Id + R_z(\Delta_{\textit{high}})\Delta_{\textit{reg.}} \right]^{-1}\cdot R_z(\Delta_{\textit{high}})\right\|\\
	\leq&\left\|\frac{P_0^{\textit{high}}}{-z} \right\|\cdot \left\| \left[Id + \frac{P_0^{\textit{high}}}{-z}\Delta_{\textit{reg.}} \right]^{-1} - \left[Id + R_z(\Delta_{\textit{high}})\Delta_{\textit{reg.}} \right]^{-1}\right\| + \left\|\frac{P_0^{\textit{high}}}{-z} - R_z(\Delta_{\text{high}})\right\|\cdot \left\|\left[Id + R_z(\Delta_{\textit{high}})\Delta_{\textit{reg.}} \right]^{-1} \right\|\\
	\leq& \frac{1}{|z|} \left\| \left[Id + \frac{P_0^{\textit{high}}}{-z}\Delta_{\textit{reg.}} \right]^{-1} - \left[Id + R_z(\Delta_{\textit{high}})\Delta_{\textit{reg.}} \right]^{-1}\right\| + \left( 1+ \frac{\|\Delta_{\text{reg.}}\|}{|z|}\right) \cdot \frac{1}{\lambda_1(\Delta_{\textit{high}})}.
	\end{align}
	Hence it remains to bound the left hand summand. For this we use the following fact (c.f. \cite{horn}, Section 5.8. "Condition numbers: inverses and linear systems"):
	\ \\
	\ \\ 
	Given square matrices $A,B,C$ with $C = B-A$ and $\|A^{-1}C\|<1$, we have
	\begin{equation}
	\|A^{-1} - B^{-1}\|\leq \frac{\|A^{-1}\|\cdot \|A^{-1}C\|}{1 - \|A^{-1}C\|}.
	\end{equation}
	In our case, this yields (together with $\|P_0^{\textit{high}}\| = 1$) that
	\begin{align}
	&\left\| \left[Id + P_0^{\textit{high}}/(-z)\cdot\Delta_{\textit{reg.}} \right]^{-1} - \left[Id + R_z(\Delta_{\textit{high}})\Delta_{\textit{reg.}} \right]^{-1}\right\|\\
	\leq&\frac{\left(1 + \|\Delta_{\text{reg.}}\|/|z|\right)^2\cdot\|\Delta_{\text{reg.}}\|\cdot\|\frac{P_0^\text{high}}{-z} - R_z(\Delta_{\text{high}})\|}{1-\left(1 + \|\Delta_{\text{reg.}}\|/|z|\right)\cdot\|\Delta_{\text{reg.}}\|\cdot\|\frac{P_0^\text{high}}{-z} - R_z(\Delta_{\text{high}})\|}
	\end{align}
	For $S_{\text{high}}$ sufficiently large, we have
	\begin{equation}
	\|-P_0^\text{high}/z - R_z(\Delta_{\text{high}})\| \leq \frac{1}{2 \left(1 + \|\Delta_{\text{reg.}}\|/|z|\right)}
	\end{equation}
	so that we may estimate
	
	\begin{align}
	&\left\| \left[Id + \Delta_{\textit{reg.}}\frac{P_0^{\textit{high}}}{-z} \right]^{-1} - \left[Id + \Delta_{\textit{reg.}}R_z(\Delta_{\textit{high}}) \right]^{-1}\right\|\\
	\leq& 2\cdot(1+\|\Delta_{\text{reg.}}\|)\cdot\|\frac{P_0^\text{high}}{-z} - R_z(\Delta_{\text{high}})\|\\
	=&2\frac{1+\|\Delta_{\text{reg.}}\|/|z|}{\lambda_1(\Delta_{\text{high}})}
	\end{align}
	Thus we have now established
	\begin{equation}
	\left| \left[Id +\frac{P_0^{\textit{high}}}{-z}  \Delta_{\textit{reg.}} \right]^{-1}\cdot\frac{P_0^{\textit{high}}}{-z} -R_z(\Delta)\right| = \mathcal{O}\left(\frac{\|\Delta_{\text{reg.}}\|}{\lambda_1(\Delta_{\text{high}})}\right).
	\end{equation}
	\ \\
	Hence we are done with the proof, as soon as we can establish
	\begin{equation}
	\left[-z Id + P_0^{\textit{high}}\Delta_{\textit{reg.}} \right]^{-1}P_0^{\textit{high}} = J^\uparrow R_z(\underline{\Delta})J^\downarrow,
	\end{equation}
	with $J^\uparrow, \underline{\Delta},  J^\downarrow$ as defined above.
	To this end, we first note that
	\begin{equation}\label{J_proj}
	J^\uparrow\cdot J^\downarrow = P_0^{\textit{high}}
	\end{equation}
	and
	\begin{equation}\label{J_id}
	J^\downarrow\cdot J^\uparrow = Id_{\underline{G}}.
	\end{equation}
	Indeed,the relation (\ref{J_proj}) follows from the fact that the  eigenspace corresponding to the eignvalue zero is spanned by the vectors $\{\mathds{1}_R\}_{R}$, with $\{R\}$ the connected components of $G_{\text{high}}$. Equation (\ref{J_id}) follows from the fact that
	\begin{equation}
	\langle \mathds{1}_R,\mathds{1}_R\rangle = \underline{\mu}_R.
	\end{equation}
	With this we have
	\begin{equation}
	\left[Id + P_0^{\textit{high}}\Delta_{\textit{reg.}} \right]^{-1}P_0^{\textit{high}} = \left[Id + J^\uparrow J^\downarrow \Delta_{\textit{reg.}} \right]^{-1}J^\uparrow J^\downarrow.
	\end{equation}
	To proceed, set 
	\begin{equation}
	\underline{x} := F^\downarrow x
	\end{equation}
	and
	\begin{equation}
	\mathscr{X} = \left[P_0^{\textit{high}}\Delta_{\textit{reg.}} -z Id  \right]^{-1}P_0^{\textit{high}} x.
	\end{equation}
	Then 
	\begin{equation}
	\left[P_0^{\textit{high}}\Delta_{\textit{reg.}} -z Id  \right] \mathscr{X} = P_0^{\textit{high}}x
	\end{equation}
	and hence $\mathscr{X} \in \text{Ran}(P_0^{\textit{high}})$.
	Thus we have
	\begin{equation}
	J^\uparrow J^\downarrow (\Delta_{\text{reg.}}-z Id)J^\uparrow J^\downarrow\mathscr{X} = J^\uparrow J^\downarrow x.
	\end{equation}
	Multiplying with $J^\downarrow$ from the left yields
	\begin{equation}
	J^\downarrow (\Delta_{\text{reg.}}-z Id)J^\uparrow J^\downarrow\mathscr{X} = J^\downarrow x.
	\end{equation}
	Thus we have
	\begin{equation}
	(J^\downarrow\Delta_{\text{reg.}}J^\uparrow-z Id)J^\uparrow J^\downarrow\mathscr{X} = J^\downarrow x.
	\end{equation}
	This -- in turn -- implies 
	\begin{equation}
	J^\uparrow J^\downarrow  \mathscr{X} 
	= \left[J^\downarrow\Delta_{\text{reg.}}J^\uparrow - z Id \right]^{-1}J^\downarrow x.
	\end{equation}
	Using 
	\begin{equation}
	P_0^{\textit{high}} \mathscr{X} = \mathscr{X},
	\end{equation}
	we then have
	\begin{equation}
	\mathscr{X} = J^\uparrow\left[J^\downarrow\Delta_{\text{reg.}}J^\uparrow - z Id \right]^{-1}J^\downarrow x.
	\end{equation}
	We have thus concluded the proof if we can prove that $J^\downarrow\Delta_{\text{reg.}}J^\uparrow$ is the Laplacian corresponding to the graph $\underline{G}$ defined in Definition \ref{app_lim_graph_def}. But this is a straightforward calculation.
\end{proof}

As a corollary, we find

\begin{Cor}\label{polynom_cor}
	We have
	\begin{equation}
	R_z(\Delta)^k \rightarrow J^\uparrow R^k(\underline{\Delta})J^\downarrow
	\end{equation}
\end{Cor}
\begin{proof}
	This follows directly from the fact that
	\begin{equation}
	J^\downarrow J^\uparrow = Id_{\underline{G}}.
	\end{equation}
\end{proof}

\section{Proof of Theorem \ref{approx_theorem}}\label{approx_thm_proof}
Here we prove Theorem \ref{approx_theorem}, which we restate for convenience:
\begin{Thm}	Fix $\epsilon > 0$ and $z<0$. For arbitrary functions $g,h : [0,\infty] \rightarrow \mathds{R}$ with $\lim_{\lambda\rightarrow\infty}g(\lambda)=\text{const.}$ and $\lim_{\lambda\rightarrow\infty}h(\lambda)=0$, there are filters $f_{z,\theta}^0, f_{z,\theta}^I$ of Type-$0$ and Type-I  respectively such that $\|f_{z,\theta}^0-g\|_{\infty}, \|f_{z,\theta}^I-h\|_{\infty} < \epsilon$.
\end{Thm}
\begin{proof}
	The Stone-Weierstrass theorem (see e.g. \cite{Teschl}) states that any sub-algebra of continuous functions that are constant at infinity is already dense (in the topoloogy of uniform convergence) if this sub-algebra separates points.		
	
	Thus -- using the Stone-Weierstrass Theorem -- all we have to prove to establish the claim is that for every pair of points $x,y \geq 0$ there is a function $f_{\theta}$ with
	\begin{equation}
	f_{\theta}(x) \neq f_{\theta}(y).
	\end{equation}
	But this is clear since (for $z < 0$) the function
	\begin{equation}
	\frac{1}{\cdot - z}: [0, \infty ) \longrightarrow \mathds{R}
	\end{equation}
	(which generates the algebra of functions we consider) is already everywhere defined and injective.
\end{proof}

\section{Stability Theory}\label{stab_theo}
Here we provide stability results to input- and edge-weight- perturbations for our architecture. For convenience, we restate our layer-wise update rule here again:

Given a feature matrix $X^\ell \in \mathds{R}^{N \times F_\ell}$ in layer $\ell$, with column vectors $\{X_j^\ell\}_{j=1}^{F_\ell}$, the feature vector $X^{\ell+1}_i$  in layer $\ell+1$ is calculated as 
$
X^{\ell+1}_i = \text{ReLU}\left( \sum_{j=1}^{F_{\ell+1}}  f_{z,\theta^{\ell+1}_{ij}}(\Delta) \cdot X^{\ell}_j + b_i^{\ell+1} \right)
$
with a learnable bias vector $b^{\ell+1}_i$. Collecting biases into a matrix $B^{\ell+1} \in \mathds{R}^{F_{\ell+1}\times N}$, we  efficiently implement this using matrix-multiplications as
\begin{equation}\label{matriximplementation:appendix}
X^{\ell+1} = \text{ReLU}\left( \sum\limits_{k = a}^{K}  (T - \omega Id)^{-k} \cdot X^{\ell}  \cdot W^{\ell+1}_{k}  + B^{\ell+1} \right)
\end{equation}
with weight matrices $\{W_k^{\ell+1}\}$ in $\mathds{R}^{F_{\ell}\times F_{\ell+1}}$. 
Biases are implemented as $b_i = \beta_i\cdot\mathds{1}_G$, with $\mathds{1}_G$ the vector of all ones on $G$ and $\beta_i \in \mathds{R}$ learnable.

Our first result main-body of the paper then concerns stability to perturbations of input signals:
\begin{Thm}\label{node_lev_stab_appendix}
	Let $\Phi_L$ be the map associated to an $L$-layer deep ResolvNet. Denote the collection of weight matrices in layer $\ell$ by $\mathscr{W}^\ell := \{W_k\}_{K=a}^{K_\ell}$. 
	We have 
	\begin{equation}\label{node_stab_app}
	\|\Phi_L(X) - \Phi_L(Y)\|_2 \leq \|X - Y\|_2\cdot  \prod_{\ell=1}^L\|\mathscr{W}^\ell\|_{z},
	\end{equation}
	with 	
	\begin{equation}
	\|\mathscr{W}^\ell\|_{z} := \sum_{k = a}^K \frac{1}{|z|^k}\|W^\ell_k\|
	\end{equation}
	aggregating singular values of  weight matrices.
\end{Thm}
\begin{proof}
	Let us denote (hidden) feature matrices in layer $\ell$	by $X^\ell$ (resp. $Y^\ell$).

	We note the following:
	\begin{align}
	\|X^L - Y^L\| &= \left\|\text{ReLU}\left( \sum\limits_{k=a}^K R_z^k(\Delta) X^{L-1}W^L_k + B^L\right) - \text{ReLU}\left( \sum\limits_{k=a}^K R_z^k(\Delta) Y^{L-1}W^L_k + B^L\right)\right\|\\
	&\leq \left\|\left( \sum\limits_{k=a}^K R_z^k(\Delta) X^{L-1}W^L_k + B^L\right) - \left( \sum\limits_{k=a}^K R_z^k(\Delta) Y^{L-1}W^L_k + B^L\right)\right\|\\
	&\leq \left\| \sum\limits_{k=a}^K R_z^k(\Delta) X^{L-1}W_k  -  \sum\limits_{k=a}^K R_z^k(\Delta) Y^{L-1}W^L_k\right\|\\
	&\leq \sum\limits_{k=a}^K \left\| R_z^k(\Delta)\right| \cdot \left\| X^{L-1}  -   Y^{L-1}\right\|\cdot\left\|W^L_k\right\|\\
	&=\sum\limits_{k=a}^K \frac{1}{|z|^k} \cdot \left\| X^{L-1}  -   Y^{L-1}\right\|\cdot\left\|W^L_k\right\| \\
	&\leq  \|\mathscr{W}^L\|_z \cdot \left\| X^{L-1}  -   Y^{L-1}\right\|.
	\end{align}
	Iterating through the layers yields the desired inequality (\ref{node_stab_app}).
\end{proof}

%

In preparation for our next result --  Theorem \ref{varying_Laplace} below --  we note the following:
\begin{Lem}\label{norm_bound}
	Let $\Phi_L$ be the map associated to an $L$-layer deep ResolvNet. With weights and biases denoted as above, we have
	\begin{align}\label{single_stab_app}
	\|\Phi_L(X)\| \leq \|B^L\|+\sum\limits_{m=0}^L\left( \prod\limits_{j=0}^m \|\mathscr{W}^{L-1-j}\|_z\right)\|B^{L-1-j}\| + \left(\prod\limits_{\ell=1}^L\|\mathscr{W}^\ell\|_z\right)\cdot\|X\|_2
	\end{align}
\end{Lem}

\begin{proof}
	We have
	\begin{align}
	\|X\|^L \leq& \left\|\text{ReLU}\left( \sum\limits_{k=a}^K R_z^k(\Delta) X^{L-1}W_k + B^L\right) \right\|\\
	\leq& \left\| \sum\limits_{k=a}^K R_z^k(\Delta) X^{L-1}W^L_k + B^L \right\|\\
	\leq&\left\| \sum\limits_{k=a}^K R_z^k(\Delta)X^{L-1}W^L_k \right\| +\left\| B^L \right\|\\
	\leq&\sum\limits_{k=a}^K\|R_z^k(\Delta)\|\cdot\|X^{L-1}\|\cdot \|W^L_k\|+\left\| B^L \right\|\\
	\leq&\left(\sum\limits_{k=a}^K\frac{\|W^L_k\|}{|z|^k}\right)\cdot \|X^{L-1}\| + \|B^L\|.
	\end{align}
	Iterating this through all layers, we obtain (\ref{single_stab_app}).
\end{proof}

Before we can establish Theorem \ref{varying_Laplace} below, we need two additional (related) preliminary results:

\begin{Lem}\label{one_to_minus_one_app}
	Let us use the notation $\widetilde{R}_z := (\widetilde{\Delta} - z Id)^{-1}$ and $R_z := (\Delta - z Id)^{-1}$ for resulvents corresponding to two different Laplacians $\Delta$ and $\widetilde{\Delta}$. We have
	
	\begin{equation}
	\|R_z - \widetilde{R}_z\| \leq \frac{1}{|z|^3} \|\Delta - \widetilde{\Delta}\|
	\end{equation}
	%
	%
\end{Lem}
\begin{proof}
	Let $T$ and $\widetilde T$ be (finite dimensional) operators. Choose $z$ so that it is neither an eigenvalue of $T$ nor $\widetilde T$.
	
	To showcase the principles underlying the proof, let us use the notation
	\begin{equation}
	R_z(T) \equiv \frac{1}{T - z}.
	\end{equation}

	We note the following
	\begin{align}
	&\frac{1}{\widetilde{T} - z}(\widetilde{T}  -  T) \frac{1}{T-z}\\
	=& \frac{1}{\widetilde{T} - z}\widetilde{T} \frac{1}{T-z} - \frac{1}{\widetilde{T} - z} T \frac{1}{T-z}\\
	=&\left[ \frac{1}{\widetilde T-z} (\widetilde{T} - z) + \frac{z}{\widetilde T - z}\right] \frac{1}{T-z} - \frac{1}{\widetilde T-z} \left[ \frac{1}{T-z} (T - z) + \frac{z}{ T - z}\right] \\
	=& z \left(  \frac{1}{T - z} - \frac{1}{\widetilde{T} - z}\right).
	\end{align}
	Rearranging and using 
	\begin{equation}
	\|R_z(\Delta)\| = \|R_z(\widetilde(\Delta))\| = \frac{1}{|z|}
	\end{equation}
	together with the sub-multiplicativity of the operator-norm $\|\cdot\|$ yields the claim.
\end{proof}
We also note the following estimate on differences of powers of resolvents:

\begin{Lem}\label{no_J_k_to_lin}
	Let $\widetilde{R}_z := (\widetilde{\Delta} - z Id)^{-1}$ and $R_z := (\Delta - z Id)^{-1}$.
	For any natural number $k$, we have
	\begin{equation}
	\|\widetilde{R}^k_z - R^k_z\| \leq \frac{k}{|z|^{k-1}} \|\widetilde{R}_z - R_z\|
	\end{equation}
	
\end{Lem}
\begin{proof}
	We note that for arbitrary matrices $T,\widetilde{T}$, we have	
	\begin{align}
	&\widetilde{T}^k  -  T^k = \widetilde{T}^{k-1}(\widetilde{T} -  T ) +  (\widetilde{T}^{k-1} -  T^{k-1} )T\\
	= & \widetilde{T}^{k-1}(\widetilde{T} -  T ) +  \widetilde{T}^{k-2}(\widetilde{T} -  T )T + (\widetilde{T}^{k-2} -  T^{k-2} )T^2.
	\end{align}
	Iterating this and using
	\begin{equation}
	\|R_z(\Delta)\| = \|R_z(\widetilde{\Delta})\|=\frac{1}{|z|}
	\end{equation}	
	for $z < 0$ then yields the claim.
\end{proof}
Having established the preceding lemmata, we can now establish stability to perturbations of the edge weights:
\begin{Thm}\label{varying_Laplace}
	Let $\Phi_L$ and $\widetilde\Phi_L$	be the maps associated to ResolvNets with the same network architecture, but based on Laplacians $\Delta$ and $\widetilde{\Delta}$ respectively. We have
	\begin{equation}\label{edge_variations}
	\|\Phi_L(X) - \widetilde{\Phi}_L(X)\|_2 \leq \left(C_1(\mathscr{W})\cdot\|X \|_2+C_2(\mathscr{W},\mathscr{B})\right)  \cdot \|\Delta - \widetilde{\Delta}\|.
	\end{equation}
	Here, the stability constants  $C_{1}(\mathscr{W})$ and $C_2(\mathscr{W},\mathscr{B})$ are polynomials in (the largest) singular values of weight matrices and weight matrices as well as bias matrices, respectively.
\end{Thm}
\begin{proof}
	Denote by $X^\ell$ and $\widetilde{X}^\ell$ the (hidden) feature matrices generated in layer $\ell$ for networks based on Laplacians $\Delta$ and $\widetilde{\Delta}$ respectively: I.e. we have 
	\begin{equation}
	X^\ell = \text{ReLU}\left( \sum\limits_{k=a}^K R_z^k(\Delta) X^{\ell-1}W_k + B^\ell\right)
	\end{equation}
	and
	\begin{equation}
	\widetilde{X}^\ell =\text{ReLU}\left( \sum\limits_{k=a}^K R_z^k(\widetilde{\Delta}) \widetilde{X}^{\ell-1}W_k + B^\ell\right).
	\end{equation}
	Using the fact that $\text{ReLU}(\cdot)$ is Lipschitz continuous with Lipschitz constant $D=1$, we have
	\begin{align}
	&\|X^L - \widetilde{X}^L\|  \\
	=&\left\| \text{ReLU}\left( \sum\limits_{k=a}^K R_z^k(\Delta) X^{L-1}W^L_k + B^L\right) - \text{ReLU}\left( \sum\limits_{k=a}^K R_z^k(\widetilde{\Delta}) \widetilde{X}^{L-1}W^L_k + B^L\right) \right\|\\
	\leq&\left\| \left( \sum\limits_{k=a}^K R_z^k(\Delta) X^{L-1}W^L_k + B^L\right) - \left( \sum\limits_{k=a}^K R_z^k(\widetilde{\Delta}) \widetilde{X}^{L-1}W^L_k + B^L\right) \right\|\\
	\leq&\left\|  \sum\limits_{k=a}^K R_z^k(\Delta) X^{L-1}W^L_k  - \sum\limits_{k=a}^K R_z^k(\widetilde{\Delta}) \widetilde{X}^{L-1}W^L_k  \right\|\\
	\leq &\left\|  \sum\limits_{k=a}^K (R_z^k(\Delta) - R_z^k(\widetilde\Delta) ) X^{L-1}W^L_k  \right\| + \sum\limits_{k=a}^K\|R_z(\widetilde{\Delta})\|\cdot\|\widetilde{X}^{L-1} - X^{L-1}\|\cdot\|W^L_k\|\\
	\leq &\left\|  \sum\limits_{k=a}^K (R_z^k(\Delta) - R_z^k(\widetilde\Delta) ) X^{L-1}W^L_k  \right\| + \|\mathscr{W}^L\|_z\cdot\|\widetilde{X}^{L-1} - X^{L-1}\|\\
	\leq &  \sum\limits_{k=a}^K \left\|R_z^k(\Delta) - R_z^k(\widetilde\Delta)\right\|\cdot\left\| X^{L-1}\right\|\cdot \left\|W^L_k  \right\| + \|\mathscr{W}^L\|_z\cdot\|\widetilde{X}^{L-1} - X^{L-1}\|\\
	\end{align}
	
	Applying Lemma \ref{no_J_k_to_lin} yields
	\begin{align}
	&\|X^L - \widetilde{X}^L\|  \\
	\leq &  \left(\sum\limits_{k=a}^K \frac{k}{|z|^{k-1}}\left\|W^L_k  \right\|\right)\cdot\left\| X^{L-1}\right\|\cdot\left\|R_z(\Delta) - R_z(\widetilde\Delta)\right\|  + \|\mathscr{W}^L\|_z\cdot\|\widetilde{X}^{L-1} - X^{L-1}\|.\\
	\end{align}
	
	Using Lemma \ref{one_to_minus_one_app}, we then have
	
	\begin{align}
	&\|X^L - \widetilde{X}^L\|  \\
	\leq &  \left(\sum\limits_{k=a}^K \frac{k}{|z|^{k+2}}\left\|W^L_k  \right\|\right)\cdot\left\| X^{L-1}\right\|\cdot\left\|\Delta - \widetilde\Delta\right\|  + \|\mathscr{W}^L\|_z\cdot\|\widetilde{X}^{L-1} - X^{L-1}\|.\\
	\end{align}

	Lemma \ref{norm_bound} then yields
	\begin{align}
	&\|X^L - \widetilde{X}^L\|  \\
	\leq &  \left(\sum\limits_{k=a}^K \frac{k}{|z|^{k+2}}\left\|W^L_k  \right\|\right)\cdot\\
	\cdot&\left[\|B^L\|+\sum\limits_{m=0}^L\left( \prod\limits_{j=0}^m \|\mathscr{W}^{L-1-k}\|_z\right)\|B^{L-1-k}\|
	+ \left(\prod\limits_{\ell=1}^L\|\mathscr{W}^\ell\|_z\right)\cdot\|X\|_2 \right]\cdot\| \widetilde{\Delta} - \Delta\|\\
	+& \|\mathscr{W}^L\|_z\cdot\|\widetilde{X}^{L-1} - X^{L-1}\|.
	\end{align}
	Iterating this through the layers and collecting summands yields the desired relation (\ref{edge_variations}).
\end{proof}

%

\newpage
\section{Stability under Scale Variations}\label{top_stab_theo}
Here we provide details on the scale-invariance results discussed in Section \ref{stability}.

In preparation, we will first need to prove a lemma relating powers of resolvents on the original graph $G$ and its limit-description $\underline{G}$:

\begin{Lem}\label{J_k_to_lin}
	Let $\underline{R}_z := (\underline{\Delta} - z Id)^{-1}$ and $R_z := (\Delta - z Id)^{-1}$.
	For any natural number $k$, we have
	\begin{equation}
	\|J^\uparrow\underline{R}^k_zJ^\downarrow - R^k_z\| \leq \frac{k}{|z|^{k-1}} \|J^\uparrow\underline{R}_zJ^\downarrow - R_z\|
	\end{equation}
	
\end{Lem}
The proof proceeds in analogy to that of Lemma \ref{no_J_k_to_lin}:
\begin{proof}
	We note that for arbitrary matrices $T,\widetilde{T}$, we have	
	\begin{align}
	&\widetilde{T}^k  -  T^k = \widetilde{T}^{k-1}(\widetilde{T} -  T ) +  (\widetilde{T}^{k-1} -  T^{k-1} )T\\
	= & \widetilde{T}^{k-1}(\widetilde{T} -  T ) +  \widetilde{T}^{k-2}(\widetilde{T} -  T )T + (\widetilde{T}^{k-2} -  T^{k-2} )T^2.
	\end{align}
	Iterating this, using
	\begin{equation}
	\|R_z(\Delta)\| = \|J^\uparrow R_z(\underline{\Delta})J^\downarrow\|=\frac{1}{|z|}
	\end{equation}	
	for $z < 0$ together with $\|J^\uparrow\|,\|J^\downarrow\|\leq1$ and
	\begin{equation}
	J^\uparrow\underline{R}^k_zJ^\downarrow = \left(J^\uparrow\underline{R}_zJ^\downarrow\right)^k
	\end{equation}
	(which holds since $J^\downarrow J^\uparrow = Id_{\underline{G}} $) then yields the claim.
	
	Note that the equation 
	\begin{equation}
	\|J^\uparrow R_z(\underline{\Delta})J^\downarrow\|=\frac{1}{|z|}
	\end{equation}	
	holds, because we may write
	\begin{align}
	\|J^\uparrow R_z(\underline{\Delta})J^\downarrow\| 
	=	\|\lim\limits_{\lambda_1(\Delta_{\text{high}})\rightarrow \infty}  R_z(\Delta)\| 	
	=\lim\limits_{\lambda_1(\Delta_{\text{high}})\rightarrow \infty} 	\| R_z(\Delta)\| 	 		 
	=\lim\limits_{\lambda_1(\Delta_{\text{high}})\rightarrow \infty} \frac{1}{|z|}
	= \frac{1}{|z|}.
	\end{align}
	
\end{proof}

Hence let us now prove Stability-Theorem \ref{varying_spaces}, which we restate here for convenience:
\begin{Thm}\label{varying_spaces_app}
	Let $\Phi_L$ and $\underline{\Phi}_L$	be the maps associated to ResolvNets with the same learned weight matrices and biases but
	deployed on graphs $G$ and $\underline{G}$ as defined in Section \ref{graph_level_desire} . We have	
	\begin{equation}\label{stab_eq_0_app}
	\|\Phi_L(J^\uparrow \underline{X}) -  J^\uparrow\underline{\Phi}_L(\underline{X})\|_2 \leq \left(C_1(\mathscr{W})\cdot\|X \|_2+C_2(\mathscr{W},\mathscr{B})\right)  \cdot  	\left\|R_z(\Delta) -  J^\uparrow R_z(\underline{\Delta}) J^\downarrow\right\|
	\end{equation}
	if the network is based on Type-$0$ resolvent filters (c.f. Section \ref{resolv_arch}). Additionally, we have
	\begin{equation}\label{stab_eq_I_app}
	\|\Phi_L(X) - J^\uparrow\underline{\Phi}_L(J^\downarrow X)\|_2 \leq \left(C_1(\mathscr{W})\cdot\|X \|_2+C_2(\mathscr{W},\mathscr{B})\right)  \cdot  	\left\|R_z(\Delta) -  J^\uparrow R_z(\underline{\Delta}) J^\downarrow\right\|
	\end{equation}
	if only Type-I filters are used in the network. Here $C_1(\mathscr{W})$ and $C_2(\mathscr{W},\mathscr{B})$ are constants that depend polynomially on singular values of learned weight matrices $\mathscr{W}$ and biases $\mathscr{B}$.
\end{Thm}

\begin{proof}
	Let us first prove (\ref{stab_eq_I_app}).
	To this end, let us define 
	\begin{equation}
	\underline{X} := J^\downarrow X.
	\end{equation}
	Let us further use the notation $\underline{R}_z := (\underline{\Delta} - z Id)^{-1}$ and $R_z := (\Delta - z Id)^{-1}$.

	Denote by $X^\ell$ and $\widetilde{X}^\ell$ the (hidden) feature matrices generated in layer $\ell$ for networks based on resolvents $R_z$ and $\underline{R}_z$ respectively: I.e. we have 
	\begin{equation}
	X^\ell = \text{ReLU}\left( \sum\limits_{k=a}^K R_z^k X^{\ell-1}W_k + B^\ell\right)
	\end{equation}
	and
	\begin{equation}
	\widetilde{X}^\ell =\text{ReLU}\left( \sum\limits_{k=a}^K \underline{R}_z^k \widetilde{X}^{\ell-1}W_k + \underline{B}^\ell\right).
	\end{equation}
	Here, since bias terms are proportional to constant vectors on the graphs, as detailed in Section \ref{resolv_arch}, we have
	\begin{equation}
	J^\downarrow B = \underline{B}
	\end{equation}
	and 
	\begin{equation}\label{bias_up}
	J^\uparrow  \underline{B} = B
	\end{equation}
	for bias matrices $B$ and $\underline{B}$ in networks deployed on $G$ and $\underline{G}$ respectively.

	We then have

	\begin{align}
	&\|\Phi_L(X) - J^\uparrow\underline{\Phi}_L(J^\downarrow X)\|\\
	=&\|X^L - J^\uparrow \widetilde{X}^L\|\\
	=&\left\| \text{ReLU}\left( \sum\limits_{k=a}^K R_z^k X^{L-1}W^L_k + B^L\right) - J^\uparrow\text{ReLU}\left( \sum\limits_{k=a}^K \underline{R}_z^k \widetilde{X}^{L-1}W^L_k + \underline{B}^L\right) \right\|\\
	=&\left\| \text{ReLU}\left( \sum\limits_{k=a}^K R_z^k X^{L-1}W^L_k + B^L\right) - \text{ReLU}\left( \sum\limits_{k=a}^K J^\uparrow\underline{R}_z^k \widetilde{X}^{L-1}W^L_k + B^L\right) \right\|.
	\end{align}
	Here we used the fact that since $\text{ReLU}(\cdot)$ maps positive entries to positive entries and acts pointwise, it commutes with $J^\uparrow$. We also made use of (\ref{bias_up}).\\
	Using the fact that $\text{ReLU}(\cdot)$ is  Lipschitz-continuous with Lipschitz constant $D=1$, we can establish
	\begin{align}
	&\|\Phi_L(X) - J^\uparrow\underline{\Phi}_L(J^\downarrow X)\|
	\leq\left\|  \sum\limits_{k=a}^K R_z^k X^{L-1}W^L_k  -  \sum\limits_{k=a}^K J^\uparrow\underline{R}_z^k \widetilde{X}^{L-1}W^L_k  \right\|.
	\end{align}
	Using the fact that $J^\downarrow J^\uparrow = Id_{\underline{G}}$, we have
	\begin{align}
	&\|\Phi_L(X) - J^\uparrow\underline{\Phi}_L(J^\downarrow X)\|
	\leq\left\|  \sum\limits_{k=1}^K R_z^k X^{L-1}W^L_k  -  \sum\limits_{k=1}^K (J^\uparrow\underline{R}_z^kJ^\downarrow) J^\uparrow \widetilde{X}^{L-1}W^L_k  \right\|.
	\end{align}
	From this, we find	(using $\|J^\uparrow\|,\|J^\downarrow\|\leq1$ ), that
	
	\begin{align}
	&\|X^L - J^\uparrow\widetilde{X}^L\|  \\
	\leq&\left\|  \sum\limits_{k=0}^K R_z^k X^{L-1}W^L_k  - \sum\limits_{k=1}^K (J^\uparrow\underline{R}_z^kJ^\downarrow) J^\uparrow \widetilde{X}^{L-1}W^L_k  \right\|\\
	\leq &\left\|  \sum\limits_{k=1}^K (R_z^k - (J^\uparrow\underline{R}_z^kJ^\downarrow) ) X^{L-1}W^L_k  \right\| + \sum\limits_{k=1}^K\|J^\uparrow \underline{R}_zJ^\downarrow\|\cdot\|J^\uparrow\widetilde{X}^{L-1} - X^{L-1}\|\cdot\|W^L_k\|\\
	\leq &\left\|  \sum\limits_{k=1}^K (R_z^k - (J^\uparrow\underline{R}_z^kJ^\downarrow) ) X^{L-1}W^L_k  \right\| + \|\mathscr{W}^L\|_z\cdot\|J^\uparrow\widetilde{X}^{L-1} - X^{L-1}\|\\
	\leq &  \sum\limits_{k=1}^K \left\|R_z^k - (J^\uparrow\underline{R}_z^kJ^\downarrow)\right\|\cdot\left\| X^{L-1}\right\|\cdot \left\|W^L_k  \right\| + \|\mathscr{W}^L\|_z\cdot\|J^\uparrow\widetilde{X}^{L-1} - X^{L-1}\|\\
	\end{align}
	
	Applying Lemma \ref{J_k_to_lin} yields
	\begin{align}
	&\|X^L - J^\uparrow\widetilde{X}^L\|  \\
	\leq &  \left(\sum\limits_{k=1}^K \frac{k}{|z|^{k-1}}\left\|W^L_k  \right\|\right)\cdot\left\|R_z - (J^\uparrow\underline{R}_zJ^\downarrow)\right\|\cdot\left\| X^{L-1}\right\|  + \|\mathscr{W}^L\|_z\cdot\|J^\uparrow\widetilde{X}^{L-1} - X^{L-1}\|.\\
	\end{align}
	
	%
	%
	%
	%
	%
	%
	Lemma then \ref{norm_bound} in Appendix \ref{stab_theo} established that we have
	\begin{align}\label{will_be_aggregated}
	\|X^L\| \leq \|B^L\|+\sum\limits_{m=0}^L\left( \prod\limits_{j=0}^m \|\mathscr{W}^{L-1-k}\|_z\right)\|B^{L-1-k}\| + \left(\prod\limits_{\ell=1}^L\|\mathscr{W}^\ell\|_z\right)\cdot\|X\|.
	\end{align}
	Hence the summand on the left-hand-side can be bounded in terms of a polynomial in singular values of bias- and weight matrices, as well as $\|X\|$ and most importantly the factor $\|R_z - (J^\uparrow\underline{R}_zJ^\downarrow)\|$ which tends to zero.\\
	For the summand on the right-hand-side, we can iterate the above procedure (aggregating terms like (\ref{will_be_aggregated}) multiplied by $\|R_z - (J^\uparrow\underline{R}_zJ^\downarrow)\|$) until reaching the last layer $L=1$. There we observe 
	\begin{align}
	&\|X^1 - J^\uparrow\widetilde{X}^1\|\\
	=&\left\| \text{ReLU}\left( \sum\limits_{k=1}^K R_z^k X W^1_k + B^1\right) - J^\uparrow\text{ReLU}\left( \sum\limits_{k=1}^K \underline{R}_z^k J^\downarrow XW^1_k + \underline{B}^1\right) \right\|\\
	\leq&\left\|  \sum\limits_{k=1}^K R_z^k X W^1_k -  \sum\limits_{k=1}^K J^\uparrow\underline{R}_z^k J^\downarrow XW^1_k  \right\|\\
	\leq&\left\|  \sum\limits_{k=1}^K ( R_z^k - J^\uparrow\underline{R}_z^k J^\downarrow ) X W^1_k  \right\|\\
	\leq &  \left(\sum\limits_{k=1}^K \frac{k}{|z|^{k-1}}\left\|W^1_k  \right\|\right)\cdot\left\|R_z - (J^\uparrow\underline{R}_zJ^\downarrow)\right\|\cdot\left\| X\right\|
	\end{align}
	The last step is only possible because we let the sums over powers of resolvents start at $a=1$ as opposed to $a=0$. In the latter case, there would have remained a term $\|X - J^\uparrow J^\downarrow X\|$, which would not decay as $\lambda_1(\Delta_{\textit{high}})\rightarrow\infty$.\\
	Aggregating terms, we build up the polynomial stability constants of (\ref{stab_eq_I_app}) layer by layer, and complete the proof.\\
	\ \\
	\ \\
	The proof of (\ref{stab_eq_0_app}) proceeds in complete analogy upon defining 
	\begin{equation}
	X := J^\uparrow \underline{X}.
	\end{equation}
	Note that starting with $\underline{X}$ on $\underline{G}$, implies that we have
	\begin{equation}
	J^\uparrow J^\downarrow X \equiv J^\uparrow J^\downarrow (J^\uparrow \underline{X}) = J^\uparrow \underline{X} \equiv X.
	\end{equation}
	This avoids any complications arising from employing Type-$0$ filters in this setting.

	
\end{proof}

Next we transfer the previous result to the graph level setting:

\begin{Thm}\label{graph_level_top_stab_app}
	Denote by $\Psi$ the aggregation method introduced in Section \ref{resolv_arch}. With $\mu(G) = \sum_{i =1}^N \mu_i$ the total weight of the graph $G$, we have in the setting of  Theorem \ref{varying_spaces} with Type-I filters, that
	\begin{equation}
	\|\Psi\left(\Phi_L(X)\right) - \Psi\left(\underline{\Phi}_L(J^\downarrow X)\right)\|_2 \leq \sqrt{\mu(G)}\cdot \left(C_1(\mathscr{W})\cdot\|X \|_2+C_2(\mathscr{W},\mathscr{B})\right)  \cdot  	\left\|R_z(\Delta) -  J^\uparrow R_z(\underline{\Delta}) J^\downarrow\right\|.
	\end{equation}
\end{Thm}
\begin{proof}
	Let us first recall that our aggregation scheme $\Psi$ mapped a feature matrix $X \in \mathds{R}^{N \times F}$ to a graph-level feature vector  $\Psi(X) \in \mathds{R}^{F}$  defined component-wise as
	\begin{equation}
	\Psi(X)_j = \sum_{i = 1}^N |X_{ij}|\cdot\mu_i.
	\end{equation}
	In light of Theorem \ref{varying_spaces_app}, we are done with the proof, once we have established that
	\begin{equation}
	\|\Psi\left(\Phi_L(X)\right) - \Psi\left(\underline{\Phi}_L(J^\downarrow X)\right)\|_2 \leq \sqrt{\mu(G)} \cdot \|\Phi_L(X) - J^\uparrow\underline{\Phi}_L(J^\downarrow X)\|_2.
	\end{equation}
	To this end, we first note that
	\begin{equation}
	\Psi(J^\uparrow \underline{X}) = \Psi(\underline{X}).
	\end{equation}
	Indeed, this follows from the fact that given a connected component $R$ in $G_{\text{high}}$, the map $J^\uparrow$ assigns the same feature vector to each node $r\in R \subseteq G$ (c.f. (\ref{J_up})), together with the fact that
	\begin{equation}
	\underline{\mu}_R = \sum\limits_{r \in R} \mu_r.
	\end{equation}
	Thus we have
	\begin{equation}
	\|\Psi\left(\Phi_L(X)\right) - \Psi\left(\underline{\Phi}_L(J^\downarrow X)\right)\|_2
	=
	\|\Psi\left(\Phi_L(X)\right) - \Psi\left(J^\uparrow\underline{\Phi}_L(J^\downarrow X)\right)\|_2.
	\end{equation}
	Next let us simplify notation and write 
	\begin{equation}
	A = \Phi_L(X)
	\end{equation}
	and 
	\begin{equation}
	B = J^\uparrow\underline{\Phi}_L(J^\downarrow X)
	\end{equation}
	with $A,B \in \mathds{R}^{N \times F}$. 
	We  note:
	\begin{equation}
	\|\Psi\left(\Phi_L(X)\right) - \Psi\left(J^\uparrow\underline{\Phi}_L(J^\downarrow X)\right)\|_2^2 = \sum\limits_{j=1}^F\left( \sum\limits_{i=1}^N (|A_{ij}|-|B_{ij}|)\cdot\mu_i\right)^2.
	\end{equation}
	By means of the Cauchy-Schwarz inequality together with the inverse triangle-inequality, we have
	\begin{align}
	\sum\limits_{j=1}^F\left( \sum\limits_{i=1}^N (|A_{ij}|-|B_{ij}|)\cdot\mu_i\right)^2 
	\leq&
	\sum\limits_{j=1}^F\left[\left( \sum\limits_{i=1}^N |A_{ij}-B_{ij}|^2\cdot\mu_i\right) \cdot \left(\sum\limits_{i=1}^N \mu_i\right)\right]\\
	=&\sum\limits_{j=1}^F\left( \sum\limits_{i=1}^N |A_{ij}-B_{ij}|^2\cdot\mu_i\right) \cdot \mu(G).
	\end{align}
	Since we have 
	\begin{equation}
	\|\Phi_L(X) - J^\uparrow\underline{\Phi}_L(J^\downarrow X)\|_2^2 =\sum\limits_{j=1}^F\left( \sum\limits_{i=1}^N |A_{ij}-B_{ij}|^2\cdot\mu_i\right),
	\end{equation}
	the claim is established.
\end{proof}

\section{Additional Details on Experiments:}\label{exp_det}
All experiments were performed on a single NVIDIA Quadro RTX 8000 graphics card.
\subsection{Node Classification}\label{nodeexpdet}

\paragraph{Datasets:}

%
%
%
We test our approach for the task of node-classification on eight different standard datasets across the entire homophily-spectrum.
Among these,
\textsc{Citeseer}  \cite{Citeseer},  \textsc{Cora-ML} \cite{Cora} and \textsc{PubMed} \cite{Pubmed} are citation graphs. Here each node represents a paper and edges correspond to citations.
We also test on the  \textsc{Microsoft Academic} graph \cite{MSAcademic} where an edge that is present corresponds to co-authorship.  
Bag-of-word representations act as node features.
%
%
The \textsc{WebKB} datasets \textsc{Cornell} and \textsc{Texas}  are  datasets modeling links between websites at computer science departments of various universities\cite{geom}.  Node features are bag-of-words representation of the   respective web pages.
We also consider the actor co-occurence dataset \textsc{Actor} \cite{actor} as well as the Wikipedia based dataset \textsc{Squirrel} \cite{Squirrel}.

\paragraph{Experimental setup}
We closely follow the experimental setup of \cite{PredictThenPropagate} on which our codebase builds:
All models are trained for a fixed maximum (and unreachably high) number of $n=10000$
epochs. Early stopping is performed when the validation performance has not improved for $100$ epochs. Test-results for the parameter set achieving the highest validation-accuracy are then reported. Ties are broken by
selecting the lowest loss (c.f. \cite{GAT,PPNP}).
Confidence intervals are calculated  over multiple splits and  random seeds at the $95\%$ confidence level via bootstrapping.

\paragraph{Additional details on training and models:}
We train all models on a fixed learning rate of
\begin{equation}
\text{lr} = 0.1.
\end{equation} 
Global dropout probability $p$ of all models is optimized individually over
\begin{equation}
p  \in \{0.3, 0.35, 0.4, 0.45, 0.5\}.
\end{equation}
We use $\ell^2$ weight decay and optimize the weight decay parameter $\lambda$ for all models over
\begin{equation}
\lambda \in \{0.0001, 0.0005\}.
\end{equation}
Where applicable (i.e. not for \cite{PPNP, bernnet}) we choose a two-layer deep convolutional  architecture with the dimensions of hidden features optimized over 
\begin{equation}\label{hidden_dims}
K_\ell \in \{32,64,128\}.
\end{equation}

%
%
%
%
%
%
%

In addition to the hyperparemeters specified above, some baselines have additional hyperparameters, which we detail here:
BernNet uses an additional in-layer dropout rate of $\text{dp\_rate} = 0.5$ and for its filters a polynomial order of $K=10$  as suggested in \cite{bernnet}.
As suggested in \cite{PPNP}, the  hyperparameter $\alpha$ of PPNP is set to $\alpha = 0.2$ on the \textsc{MS\_Academic} dataset and to $\alpha=0.1$ on other datasets. Hyperparameters depth $T$ and number of stacks $K$ of the ARMA convolutional layer \cite{ARMA} are set to $T=1$ and $K=2$.  ChebNet also uses $K=2$ to avoid the known over-fitting issue \cite{Kipf} for higher polynomial orders. For MagNet we use $K=1$ as suggested in \cite{magnet} and choose the parameter $q$ as given in Table 1 of \cite{magnet} for the respective datasets. The graph attention network \cite{GAT} uses $8$ attention heads, as suggested in \cite{GAT}.

For our ResolvNet model, we choose a depth of $L=1$ with hidden feature dimension optimized over the values in (\ref{hidden_dims}) as for baselines. 
%
%
We empirically observed in the setting of \textit{unweighted} graphs, that rescaling the Laplacian as
\begin{equation}
\Delta_{nf} := \frac{1}{c_{nf}}\Delta 
\end{equation}
with a normalizing factor $c_{nf}$ before calculating the resolvent
\begin{equation}\label{resc_res_def}
R_z(\Delta_{nf}) := (\Delta_{nf} - z\cdot Id)^{-1}
\end{equation}
on which we base our ResolvNet architectures improved performance.

For our ResolvNet architecture,  we express this normalizing factor in terms of the largest singular value $\|\Delta\|$ of the (non-normalized) graph Laplacian. It is then selected among
\begin{equation}
c_{nf}/\|\Delta\| \in \{0.001,0.01,0.1,2\}.
\end{equation}
The value $z$ in (\ref{resc_res_def}) is selected among
\begin{equation}
(-z) \in \{0.14,0.15, 0.2, 0.25\}.
\end{equation}

We base our ResolvNet architecture  on Type-0 filters and choose the maximum resolvent-exponent $K$ as $K=1$.
%
%

%

\subsection{Graph Regression}\label{qm9_experiment}

\paragraph{Datasets:}
The first dataset we consider is the \textbf{QM}$\mathbf7$ dataset, introduced in \cite{blum, rupp}. This dataset contains descriptions of $7165$ organic molecules, each with up to seven heavy atoms, with all non-hydrogen atoms being considered heavy. A molecule is represented by its Coulomb matrix $C^{\text{Clmb}}$, whose off-diagonal elements
\begin{equation}
C^{\text{Clmb}}_{ij}	 = \frac{Z_iZ_j}{|R_i-R_j|}
\end{equation}
correspond to the Coulomb-repulsion between atoms $i$ and $j$. We discard	diagonal entries of Coulomb matrices; which would encode a polynomial fit of atomic energies to nuclear charge \cite{rupp}.

For each atom in any given molecular graph, the individual Cartesian coordinates $R_i$ and the atomic charge $Z_i$ are also accessible individually. 
To each molecule an atomization energy - calculated via density functional theory - is associated. The objective is to predict this quantity. The performance metric is mean absolute error. Numerically, atomization energies are negative numbers in the range $-600$ to $-2200$. The associated unit is $[\textit{kcal/mol}]$.

The second dataset we consider is the \textbf{QM}$\mathbf 9$ dataset \cite{QM9}, which consists of roughly 130 000 molecules in equilibrium. Beyond atomization energy, there are in total $19$ targets available on  \textbf{QM}$\mathbf 9$. We provide a complete list of targets together with abbreviations in Table \ref{qm9_targets} below:

\begin{table*}[h!]
	\caption{Targets of QM$9$}
	\scalebox{1}{
		\begingroup
		
		\setlength{\tabcolsep}{1pt} 
		\renewcommand{\arraystretch}{1} 

		\begin{tabular}{lcc}
			\toprule
			
			Symbol   & Property &Unit \\
			
			\midrule
			$U_0$ &  Internal energy at $0K$ & $eV$\\
			$U$ & Internal energy at $298.15K$ & $eV$\\
			$H$ & Enthalpy at $298.15K$ & $eV$\\
			$G$ & Free energy at $298.15K$ & $eV$\\
			$U_0^{\text{ATOM}}$ &  Atomization energy at $0K$ & $eV$\\
			$U^{\text{ATOM}}$ & Atomization energy at $298.15K$ & $eV$\\
			$H^{\text{ATOM}}$ & Atomization enthalpy at $298.15K$ & $eV$\\
			$G^{\text{ATOM}}$ & Atomization free energy at $298.15K$ & $eV$\\
			$c_v$ & Heat capacity at $298.15K$ & $\frac{\text{cal}}{\text{mol}\cdot\text{K}}$\\
			$\mu$ & Dipole moment & $D$\\
			$\alpha$ & Isotropic polarizability & $\alpha_0^3$\\
			$\epsilon_{\text{HOMO}}$ & Highest occupied molecular orbital energy & $eV$\\
			$\epsilon_{\text{LUMO}}$ & Lowest unoccupied molecular orbital energy & $eV$\\
			$\Delta \epsilon$ & Gap between $\epsilon_{\text{HOMO}}$ and $\epsilon_{\text{LUMO}}$ & $eV$\\
			$\langle R^2 \rangle$ & Electronic spatial extent & $\alpha_0^2$\\
			ZPVE & Zero point vibrational energy & $eV$\\
			A & Rotational constant & $GHz$\\
			B & Rotational constant & $GHz$\\
			C & Rotational constant & $GHz$\\

			\bottomrule
		\end{tabular}
		\endgroup
	}
	\vskip -0.1in
	\label{qm9_targets}
\end{table*}

Molecules in QM$9$ are not directly encoded via their Coulomb-matrices, as in QM$7$. However, positions and charges of individual molecules are available, from which the Coulomb matrix description is calculated for each molecule.

\paragraph{Experimental Setup:}
On both datasets, we randomly select $1500$ molecules for testing and train on the remaining graphs. On QM$7$ we run experiments for $23$ different random random seeds and report mean and standard deviation. Due to computational limitations we run experiments for $3$ different random seeds on the larger QM$9$ dataset, and report mean and standard deviation.

\paragraph{Additional details on training and models:}
All considered  convolutional layers are incorporated into a two layer deep and fully connected graph convolutional architecture.
In each hidden layer, we set the width (i.e. the hidden feature dimension) to 
\begin{equation}
F_1 = F_2 =64.
\end{equation}
For BernNet, we set the polynomial order to $K=3$ to combat appearing numerical instabilities. ARMA is set to $K=2$ and $T=1$.  ChebNet uses $K=2$. 
For all baselines, the standard mean-aggregation scheme is employed after the graph-convolutional layers to generate graph level features. Finally, predictions are generated via an MLP.

For our model, we choose a two-layer deep instantiation of our ResolvNet architecture introduced in Section \ref{resolv_arch}. We choose Type-I filters and set $z=-1$. Laplacians are \textit{not} rescaled and resolvents are thus given as
\begin{equation}
R_{-1}(\Delta) = (\Delta + Id)^{-1}.
\end{equation}

As aggregation, we employ the graph level feature aggregation scheme introduced at the end of Section \ref{resolv_arch} with node weights set to atomic charges of individual atoms. Predictions are then generated via a final MLP with the same specifications as the one used for baselines.

All models are trained independently on each respective target.

\paragraph{Results:} Beyond the results already showcased in the main body of the paper, we here provide results for ResolvNet as well as baselines on all targets of Table \ref{qm9_targets}. These results are collected in Table \ref{full_qm9_table_app_I}, Table \ref{full_qm9_table_app_II} and Table \ref{full_qm9_table_app_III} below. 

As is evident from the tables, the ResolvNet architecture produces mean-absolute-errors comparable to those of baselines on $1/4$ of targets, while it performs significantly better on $3/4$ of targets.

The difference in performance is especially significant on the (extensive) energy targets of Table \ref{full_qm9_table_app_I}. In this Table, baselines are out-performed by factors varying between $4$ and $15$.

Table \ref{full_qm9_table_app_II} contains three additional targets where MAEs produced by ResolvNet are lower  by  factors varying between roughly two and four, when compared to baselines. 

Table \ref{full_qm9_table_app_III} finally contains MAEs corresponding to predictions of rotational constants. Here our model yields a comparable error on one target and provides better results than baselines on two out of three targets. 

\begin{table*}[h!]
	\caption{Energy prediction MAEs $[eV]$. Our Model is marked \textbf{R.N.} for \textbf{ResolvNet}.}
	\begin{scriptsize}
		\scalebox{.99}{
			\begingroup
			
			\setlength{\tabcolsep}{1pt} 
			\renewcommand{\arraystretch}{1} 
			\begin{tabular}{lcccccccccc}
				\toprule
				
				\textbf{Property}  & $U_0$ & $U$ & $H$ & $G$ & $U_0^{\text{ATOM}}$ & $U^{\text{ATOM}}$ & $H^{\text{ATOM}}$ & $G^{\text{ATOM}}$\\

				\midrule

				BernNet    & $370.42$\tiny{$\pm 38.91$}  & $382.64$\tiny{$\pm 36.52$} & $398.32$\tiny{$\pm 46.00$} & $362.69$\tiny{$\pm 24.84$}  
				& $ 3.112$\tiny{$\pm 0.285$}  & $ 3.096$\tiny{$\pm0.249 $} 
				& $3.046$\tiny{$\pm 0.277$} & $2.919$ \tiny{$\pm 0.375$} \\


				GCN  & $381.41$\tiny{$\pm 0.42$} & $376.41$\tiny{$\pm 7.10$}  & $368.01$\tiny{$\pm 16.77$} &$380.65$\tiny{$\pm 6.67$} 
				& $2.766$\tiny{$\pm 0.081$} &$2.828$\tiny{$\pm 0.091$}   
				& $2.803$\tiny{$\pm 0.077$} & $2.575$\tiny{$\pm 0.084$} \\
				

				ChebNet & $345.74$\tiny{$\pm12.30$}& $346.39$\tiny{$\pm 19.11$} &$398.32$\tiny{$\pm 22.48$} &  $350.22$\tiny{$\pm 12.32$} &
				$2.665$\tiny{$\pm0.040$} & $2.672$\tiny{$\pm0.056$} &
				$2.745$\tiny{$\pm 0.104$} & $2.477$\tiny{$\pm 0.036$}\\

				ARMA  & $327.62$\tiny{$\pm 19.83$} &  $316.09$\tiny{$\pm 18.06$} & $322.74$\tiny{$\pm 16.32$} & $320.72$\tiny{$\pm 11.98$} 
				& $2.588 $\tiny{$\pm 0.117$}  & $2.570$\tiny{$\pm0.088$}  
				& $2.600$\tiny{$\pm0.096$} & $2.326$\tiny{$\pm 0.101$} \\

				\midrule

				R.N. &  $\mathbf{21.72}$\tiny{$\pm 5.79$}   &  $\mathbf{19.14}$\tiny{$\pm 7.19$} &   $\mathbf{31.18}$\tiny{$\pm 8.622$}  & $\mathbf{53.50}$\tiny{$\pm 4.58$}
				& $\mathbf{0.605}$\tiny{$\pm 0.015$}  & $\mathbf{0.588}$\tiny{$\pm0.024$}  
				&   $\mathbf{0.593}$\tiny{$\pm 0.025$} & $\mathbf{0.607}$\tiny{$\pm 0.041$}   
				\\

				\bottomrule
			\end{tabular}
			\endgroup
		}
	\end{scriptsize}
	\vskip -0.1in
	\label{full_qm9_table_app_I}
\end{table*}

\begin{table*}[h!]
	\caption{Various target prediction MAEs. Our Model is marked \textbf{R.N.} for \textbf{ResolvNet}.}
	\begin{scriptsize}
		\scalebox{.99}{
			\begingroup
			
			\setlength{\tabcolsep}{1pt} 
			\renewcommand{\arraystretch}{1} 

			\begin{tabular}{lccccccccc}
				\toprule
				
				\textbf{Property}  & $c_v$ $\left[\frac{\text{cal}}{\text{mol}\cdot\text{K}}\right]$& $\mu$ $[D]$& $\alpha$ $[\alpha_0^3]$& $\epsilon_{\text{HOMO}}$ $[eV]$ & $\epsilon_{\text{LUMO}}$ $[eV]$ & $\Delta \epsilon$ $[eV]$ & $\langle R^2\rangle$ $[\alpha_0^2]$ & ZPVE $[eV]$\\

				\midrule

				BernNet  & $2.610$\tiny{$\pm0.986$} & $0.948$\tiny{$\pm 0.042$} & $3.519$\tiny{$\pm 0.288$}  
				& $ 0.376$\tiny{$\pm 0.028$}  & $0.649$\tiny{$\pm0.092 $} 
				& $0.841$\tiny{$\pm 0.085$} & $157.982$ \tiny{$\pm 34.804$} & $0.237$ \tiny{$\pm 0.032$} \\


				GCN & $1.521$\tiny{$\pm0.038$} & $0.936$\tiny{$\pm 0.003$} &$3.114$\tiny{$\pm 0.112$} 
				& $0.301$\tiny{$\pm 0.009$} &$0.523$\tiny{$\pm 0.018$}   
				& $0.566$\tiny{$\pm 0.016$} & $130.461$\tiny{$\pm 5.445$}& $0.185$\tiny{$\pm 0.004$} \\
				

				ChebNet &$1.455$\tiny{$\pm0.053$} &$0.881$\tiny{$\pm 0.007$} &  $3.049$\tiny{$\pm 0.092$} &
				$0.234$\tiny{$\pm0.005$} & $0.433$\tiny{$\pm0.018$} &
				$0.515$\tiny{$\pm 0.010$} & $132.695$\tiny{$\pm 2.218$}& $0.180$\tiny{$\pm 0.005$} \\

				ARMA &$1.327$\tiny{$\pm0.034$} & $0.806$\tiny{$\pm 0.031$} & $2.676$\tiny{$\pm 0.087$} 
				& $\mathbf{0.228}$\tiny{$\pm 0.010$}  & $\mathbf{0.333}$\tiny{$\pm0.009$}  
				& $\mathbf{0.380}$\tiny{$\pm0.007$} & $\mathbf{93.760}$\tiny{$\pm 4.122$}& $0.152$\tiny{$\pm 0.006$}  \\

				\midrule

				R.N.&$\mathbf{0.747}$\tiny{$\pm0.015$} & $\mathbf{0.776}$\tiny{$\pm 0.018$}  & $\mathbf{1.308}$\tiny{$\pm 0.034$}
				& $0.313$\tiny{$\pm 0.002$}  & $0.423$\tiny{$\pm0.011$}  
				&   $0.531$\tiny{$\pm 0.016$} & $97.614$\tiny{$\pm 2.308$}& $\mathbf{0.041}$\tiny{$\pm 0.008$}  
				\\

				\bottomrule
			\end{tabular}
			\endgroup
		}
	\end{scriptsize}
	\vskip -0.1in
	\label{full_qm9_table_app_II}
\end{table*}

\begin{table*}[h!]
	\caption{Rotational constants prediction MAEs. Our Model is marked \textbf{R.N.} for \textbf{ResolvNet}.}
	\begin{scriptsize}
		\scalebox{.99}{
			\begingroup
			
			\setlength{\tabcolsep}{1pt} 
			\renewcommand{\arraystretch}{1} 

			\begin{tabular}{lccc}
				\toprule
				
				\textbf{Property}   & $A$ $[GHz]$ & $B$ $[GHz]$ & $C$ $[GHz]$\\

				\midrule

				BernNet    & $0.888$\tiny{$\pm 0.034$} & $0.342$\tiny{$\pm 0.002$}  
				& $ 0.243$\tiny{$\pm 0.002$}  \\


				GCN  & $0.848$\tiny{$\pm 0.027$} &$0.281$\tiny{$\pm 0.004$} 
				& $0.183$\tiny{$\pm 0.002$}  \\
				

				ChebNet  &$0.797$\tiny{$\pm 0.034$} &  $0.262$\tiny{$\pm 0.003$} &
				$0.171$\tiny{$\pm0.003$}  \\

				ARMA & $\mathbf{0.715}$\tiny{$\pm 0.017$} & $0.259$\tiny{$\pm 0.004$} 
				& $0.168$\tiny{$\pm 0.004$} \\

				\midrule

				R.N. & $0.783$\tiny{$\pm 0.802$}  & $\mathbf{0.249}$\tiny{$\pm 0.002$}
				& $\mathbf{0.158}$\tiny{$\pm 0.001$} 
				\\

				\bottomrule
			\end{tabular}
			\endgroup
		}
	\end{scriptsize}
	\vskip -0.1in
	\label{full_qm9_table_app_III}
\end{table*}

\subsection{Scale Invariance}\label{stability_exp_details}
\paragraph{Dataset:} Again, we make use of the QM$7$ dataset \cite{rupp} and its Coulomb matrix description 
\begin{equation}\label{offdiagII}
C^{\text{Clmb}}_{ij}	 = \frac{Z_iZ_j}{|R_i-R_j|}
\end{equation}
of molecules.

\paragraph{Details on collapsing procedure:}
We modify (all) molecular graphs in QM$7$
by deflecting hydrogen atoms (H) out of their equilibrium positions towards the respective nearest heavy atom. This is possible since the QM$7$ dataset also contains the Cartesian coordinates of individual atoms.

This introduces a two-scale setting precisely as discussed in section \ref{scales}: Edge weights between heavy atoms remain the same, while  Coulomb repulsions  between H-atoms and respective nearest heavy atom increasingly diverge; as is evident from (\ref{offdiagII}).

Given an original molecular graph $G$ with node weights $\mu_i=Z_i$, the corresponding limit graph $\underline{G}$ corresponds to a coarse grained description, where heavy atoms and surrounding H-atoms are aggregated into single super-nodes in the sense of Section \ref{graph_level_desire} . 

Mathematically, $\underline{G}$ is obtained by removing all nodes corresponding to H-atoms from $G$, while adding the corresponding charges $Z_H = 1$ to the node-weights of the respective 
nearest heavy atom.
Charges 
in (\ref{offdiagII}) are modified similarly to generate the weight matrix $\underline{W}$.


On original molecular graphs, atomic charges are provided via one-hot encodings. For the graph of methane -- consisting of one carbon atom with charge $Z_C =6$ and four hydrogen atoms of charges $Z_H =1$ -- the corresponding node-feature-matrix is e.g. given as
\begin{align}
X = 
\begin{pmatrix}
0 & 0 &\cdots &0 &1 & 0 \cdots \\
1 & 0 & \cdots&0 &0 & 0 \cdots \\
1 & 0 & \cdots&0 &0 & 0 \cdots \\
1 & 0 & \cdots&0 &0 & 0 \cdots \\
1 & 0 & \cdots&0 &0 & 0 \cdots
\end{pmatrix}
\end{align}
with the non-zero entry in the first row being in the $6^{\text{th}}$ column, in order to encode the charge $Z_C=6$ for carbon.

The feature vector of an aggregated node represents charges of the heavy atom and its neighbouring H-atoms jointly.

As discussed in Definition \ref{proj_ops}, node feature matrices are translated as $\underline{X} = J^\downarrow X$. 
Applying $J^\downarrow$ to one-hot encoded atomic charges yields (normalized) bag-of-word embeddings on $\underline{G}$:  Individual entries of feature vectors encode how much of the total charge of the super-node is contributed by individual atom-types.
In the example of methane, the limit graph $\underline{G}$ consists of a single node with  node-weight
\begin{equation}
\mu= 6 + 1 + 1 + 1 + 1 = 10.
\end{equation}
The feature matrix 
\begin{equation}
\underline{X} = J^\downarrow X
\end{equation}
is  a single row-vector  given as
\begin{align}
\underline{X} = \left(\frac{4}{10},0,\cdots,0,\frac{6}{10},0,\cdots\right).
\end{align}

\paragraph{Results:}
\noindent

\begin{minipage}{0.56\textwidth}
	For convenience, we repeat here in Table \ref{qm7_collapsed_result_table_app} and Figure \ref{collapse_graph_corr_app} the results corresponding to the use of resolution-limited data in the form of coarse-grained molecular graphs during inference, that were already presented in the main body of the paper.
	
	\begin{table}[H]
		\small
		\centering
		\caption{MAE on QM$7$ via coarsified	molecular graphs. }
		\begingroup
		
		\setlength{\tabcolsep}{0.5pt} 
		\renewcommand{\arraystretch}{1} 
		\begin{tabular}{lr}
			\toprule
			\textbf{QM$7$} & MAE $[kcal/mol]$  \\
			\midrule
			BernNet   & $580.67$\tiny{$\pm  99.27$}  \\
			GCN & $124.53 $\tiny{$\pm 34.58$}   \\
			ChebNet & $645.14$\tiny{$\pm 34.59$}\\	
			ARMA & $248.96$\tiny{$\pm 15.56$}   \\
			\midrule	
			ResolvNet &  $\mathbf{16.23}$\tiny{$\pm 2.74$}    \\
		\end{tabular}
		\endgroup
		\label{qm7_collapsed_result_table_app}
	\end{table}
\end{minipage}\ \
\begin{minipage}{0.4\textwidth}
	\begin{figure}[H]
		\caption{Feature-vector-difference for collapsed ($\underline{F}$) and deformed ($F$) graphs. }
		\includegraphics[scale=0.40]{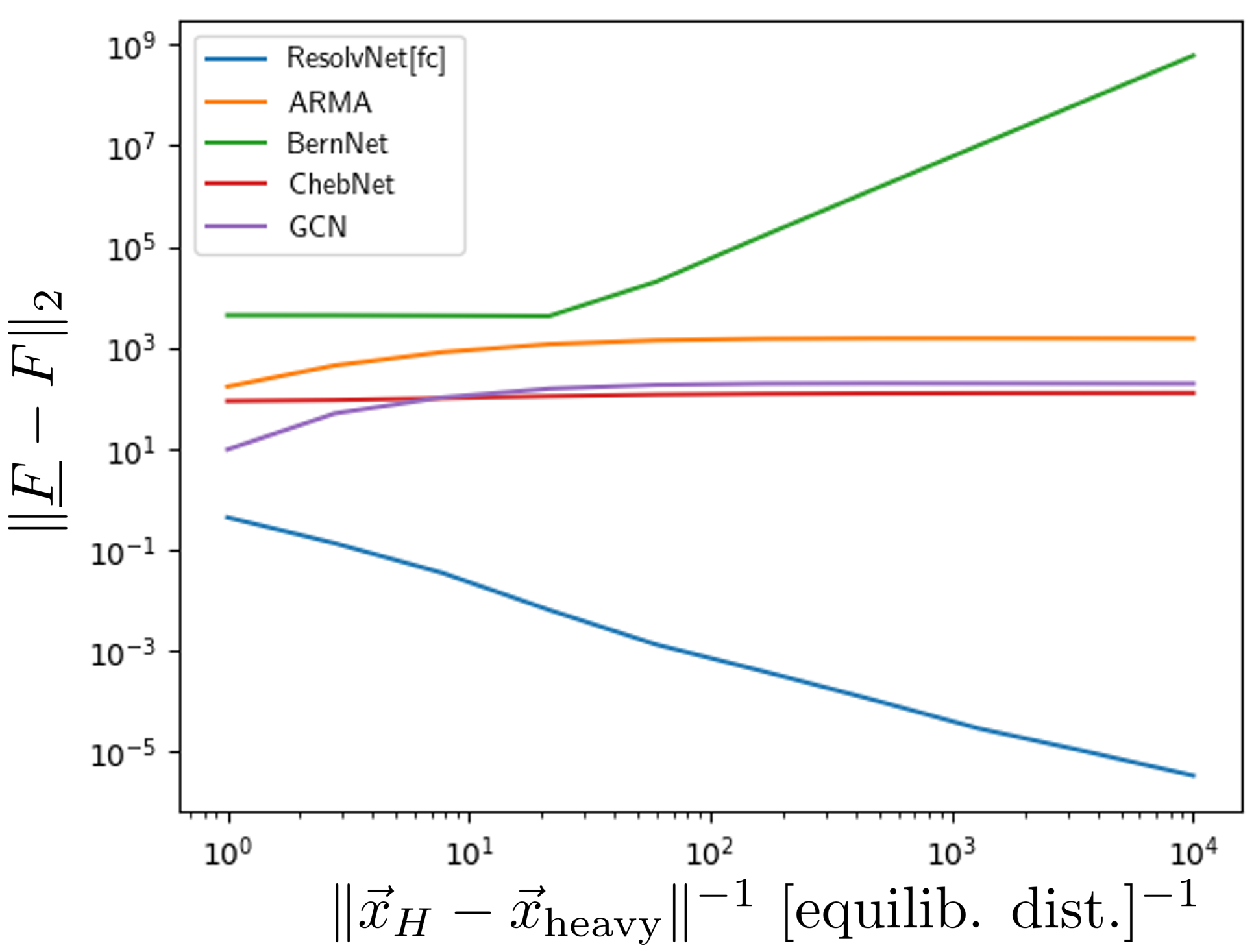}
		\noindent
		\label{collapse_graph_corr_app}
	\end{figure}
\end{minipage}\noindent

\section{Analysis of Computational Overhead}

Here we provide an analysis of the overhead of our ResolvNet method. As is evident from Tables \ref{alpha}, \ref{beta}, \ref{gamma} below, on most datasets our method is not the most memory intensive to train when compared to representative (spatial and spectral) baselines. For training times (total and per-epoch), we note that on most small to medium sized graphs, our model is not the slowest to train. On larger graphs it does take longer to train. Regarding complexity, the node update for our model is essentially $\mathcal{O}(N^2)$ (dense-dense matrix multiplication), while message passing baselines scale linearly in the number of edges.

\begin{table*}[!ht]
	\caption{Maximal Memory Consumption [GB] while training a single model of depth $2$ and  width $32$ for learning rate $\text{lr} = 0.1$, dropout $p=0.5$, weight decay $\lambda = 10^{-4}$ and early stopping patience  $t = 100$. All measurements performed on the same GPU via \texttt{torch.cuda.max\_memory\_allocated()}.  }

	\begin{scriptsize}
		\begingroup
		
		\setlength{\tabcolsep}{0.6pt} 
		\renewcommand{\arraystretch}{1} 
		
		\begin{tabular}{ccccccccc}
			\hline
			\     &  \ \ \ MS\_Acad.\ \ \ & \ \ \ Cora\ \ \ & \ \ \ Pubmed\ \ \ & \ \ \ Citeseer\ \ \ & \ \ \ Cornell\ \ \  &\ \ \ Actor\ \ \ \ \ \  & \ \ \ Squirrel\ \ \   &\ \ \ Texas\ \ \ \\ \hline
			
			ResolvNet \ \ \  &  3.47 &   0.1266  & 2.9915    & 0.0996   & 0.0070     & 0.4936    & 0.2915       & 0.0175  \\ 
			GAT  &  1.49  & 0.1559   &  0.6486    & 0.1105 &     0.0228      & 0.3666    & 2.1107        & 0.0219  \\ 
			ChebNet     &  10.19  & 0.4741 & 0.4848  & 0.3389    & 0.0249    & 0.4830    & 6.3569   & 0.0241  \\  \hline
		\end{tabular}
		\endgroup
		
	\end{scriptsize}
	\vskip -0.1in
	\label{alpha}
\end{table*}

\ \\ \ \\

\begin{table*}[!ht]
	\caption{Training Time [s] for training a single model of depth $2$ and  width $32$ for learning rate $\text{lr} = 0.1$, dropout $p=0.5$, weight decay $\lambda = 10^{-4}$ and early stopping patience  $t = 100$. All measurements performed on the same GPU. }

	\begin{scriptsize}
		\begingroup
		
		\setlength{\tabcolsep}{0.6pt} 
		\renewcommand{\arraystretch}{1} 
		
		\begin{tabular}{ccccccccc}
			\hline
			\    &   \ \ \ MS\_Acad.\ \ \ & \ \ \ Cora\ \ \ & \ \ \ Pubmed\ \ \ & \ \ \ Citeseer\ \ \ & \ \ \ Cornell\ \ \  &\ \ \ Actor\ \ \ \ \ \  & \ \ \ Squirrel\ \ \   &\ \ \ Texas\ \ \ \\ \hline
			ResolvNet \ \ \  & 474.409    &3.671     & 34.140     & 1.387     & 1.745    & 9.623     & 4.874         & 0.875     \\ 
			GAT  &  34.388     & 2.194     & 5.741      & 0.891     & 2.123      & 1.610     & 23.060        & 1.375      \\ 
			ChebNet     &   87.567   &6.818     & 3.221  & 2.833   & 2.713    & 1.488     & 14.383        & 4.511    \\ \hline
		\end{tabular}
		\endgroup
		
	\end{scriptsize}
	\vskip -0.1in
	\label{beta}
\end{table*}

\ \\ \ \\

\begin{table*}[!ht]
	\caption{Average Training Time per Epoch [ms] for training a single model of depth $2$ and  width $32$ for learning rate $\text{lr} = 0.1$, dropout $p=0.5$, weight decay $\lambda = 10^{-4}$ and early stopping patience  $t = 100$. All measurements performed on the same GPU.  }

	\begin{scriptsize}
		\begingroup
		
		\setlength{\tabcolsep}{0.6pt} 
		\renewcommand{\arraystretch}{1} 
		
		\begin{tabular}{ccccccccc}
			\hline
			\     &   \ \ \ MS\_Acad.\ \ \ & \ \ \ Cora\ \ \ & \ \ \ Pubmed\ \ \ & \ \ \ Citeseer\ \ \ & \ \ \ Cornell\ \ \  &\ \ \ Actor\ \ \ \ \ \  & \ \ \ Squirrel\ \ \  &\ \ \ Texas\ \ \ \\ \hline
			ResolvNet \ \ \  &  1359.34   &13.16    & 161.80     & 11.01    & 2.58    &32.51     & 41.30       & 2.54 \\ 
			GAT  & 60.01     & 8.22   & 29.59      & 7.24      & 3.93      & 15.05      & 62.49        & 4.07   \\
			ChebNet    &  202.23 &  12.11     & 14.31     & 10.61    & 3.89     & 13.28     & 126.17        & 3.83     \\ \hline
		\end{tabular}
		\endgroup
		
	\end{scriptsize}
	\vskip -0.1in
	\label{gamma}
\end{table*}

\end{document}